%% file: dsvgd.tex
\documentclass{article} 
\usepackage{iclr2021_conference,times}
\input{math_commands.tex}

\usepackage{hyperref}
\usepackage{url}
\usepackage{graphics}
\usepackage{graphicx}
\usepackage{amsthm}
\usepackage{bbm}
\usepackage{amsmath,amssymb,amsfonts}
\usepackage[nolist]{acronym}
\usepackage{subfigure}
\usepackage{enumitem,kantlipsum}
\usepackage{wrapfig}
\usepackage{algorithmic}
\usepackage{booktabs}
\usepackage{footnote}
\usepackage{pifont}
\usepackage{amsthm}
\usepackage{thmtools, thm-restate}

\usepackage{multirow}
\usepackage{siunitx}
\usepackage{adjustbox}
\usepackage[normalem]{ulem}
\usepackage[ruled]{algorithm2e}

\SetCommentSty{mycommfont}
\makesavenoteenv{tabular}
\makesavenoteenv{table}

\input{acronymes.tex}

\title{Federated Generalized Bayesian Learning \\ via  Distributed Stein Variational Gradient Descent}

\author{Rahif Kassab \& Osvaldo Simeone\\
King's Communications, Learning and Information Processing Lab (KCLIP) \\ Department of Engineering, King's College London\\
\texttt{\{rahif.kassab,osvaldo.simeone\}@kcl.ac.uk}
}

\usepackage{tikz}
\newcommand*\circled[1]{\tikz[baseline=(char.base)]{
            \node[shape=circle,draw,inner sep=1.5pt] (char) {#1};}}

\iclrfinalcopy 
\begin{document}

\maketitle
\begin{abstract}
This paper introduces Distributed Stein Variational Gradient Descent (DSVGD), a non-parametric generalized Bayesian inference framework for federated learning. DSVGD maintains a number of non-random and interacting particles at a central server to represent the current iterate of the model global posterior. The particles are iteratively downloaded and updated by one of the agents with the end goal of minimizing the global free energy. By varying the number of particles, DSVGD enables a flexible trade-off between per-iteration communication load and number of communication rounds. DSVGD is shown to compare favorably to benchmark frequentist and Bayesian federated learning strategies in terms of accuracy and scalability with respect to the number of agents, while also providing well-calibrated, and hence trustworthy, predictions.
\end{abstract}
\section{Introduction}
\label{sec:intro}
Federated learning refers to the collaborative training of a machine learning model across agents with distinct data sets, and it applies at different scales, from industrial data silos to mobile devices \citep{advances_openproblems_FL_kairouz}. While some common challenges exist, such as the general statistical heterogeneity -- ``non-iidnes'' -- of the distributed data sets, each setting also brings its own distinct problems. In this paper, we are specifically interested in a small-scale federated learning setting consisting of mobile or embedded devices, each having a limited data set and running a small-sized model due to their constrained memory. As an example, consider the deployment of health monitors based on data from smart-watch ECG data. In this context, we argue that it is essential to tackle the following challenges, which are largely not addressed by existing solutions:\\
\noindent $\bullet$ \emph{Trustworthiness}: In applications such as personal health assistants, the learning agents' recommendations need to be reliable and trustworthy, e.g., to decide when to contact a doctor in case of a possible emergency;\\
\noindent $\bullet$ \emph{Number of communication rounds}: When models are small, the payload per communication round may not be the main contributor to the overall latency of the training process. In contrast, accommodating many communication rounds requiring arbitrating channel access among multiple devices may yield slow wall-clock time convergence \citep{FL_comm_delay}. 
\par
Most existing federated learning algorithms, such as \ac{FedAvg} \citep{comm_efficient_learning_dec_data}, are based on frequentist principles, relying on the identification of a single model parameter vector. Frequentist learning is known to be unable to capture epistemic uncertainty, yielding overconfident decisions \citep{NN_calibration}. Furthermore, the focus of most existing works is on reducing the load per-communication round via compression, rather than decreasing the number of rounds by providing more informative updates at each round \citep{advances_openproblems_FL_kairouz}. This paper introduces a trustworthy solution that is able to reduce the number of communication rounds via a non-parametric variational inference-based implementation of federated Bayesian learning.\par
Federated Bayesian learning has the general aim of computing the global posterior distribution in the model parameter space. Existing decentralized, or federated, Bayesian learning protocols are either based on \textbf{\ac{VI}} \citep{angelino_patterns_scalable_BI, EPVI, streaming_variational_bayes_tamara, variational_multi_task} or \textbf{\ac{MC} sampling} \citep{dsgld, embarassingly_parallel_MCMC, parallel_MCMC_hierarchical}.
State-of-the-art methods in either category include \textbf{\ac{PVI}}, which has been recently introduced as a unifying distributed \ac{VI} framework that relies on the optimization over parametric posteriors; and \textbf{\ac{DSGLD}}, which is an \ac{MC} sampling technique that maintains a number of Markov chains updated via local Stochastic Gradient Descent (SGD) with the addition of Gaussian noise \citep{dsgld, sgld}. The performance of VI-based protocols is generally limited by the bias entailed by the variational approximation, while MC sampling is slow and suffers from the difficulty of assessing convergence \citep{angelino_patterns_scalable_BI}.


\par

\par
\textbf{\ac{SVGD}} has been introduced in \citep{SVGD} as a non-parametric Bayesian framework that approximates a target posterior distribution via non-random and interacting particles. \ac{SVGD} inherits the flexibility of non-parametric Bayesian inference methods, while improving the convergence speed of  \ac{MC} sampling \citep{SVGD}. By controlling the number of particles, \ac{SVGD} can provide flexible performance in terms of bias, convergence speed, and per-iteration complexity. This paper introduces a novel non-parametric distributed learning algorithm, termed \textbf{\ac{DSVGD}}, that transfers the mentioned benefits of \ac{SVGD} to federated learning.

\begin{figure}[t]
    \centering
    \subfigure[]{\includegraphics[height=1.6in]{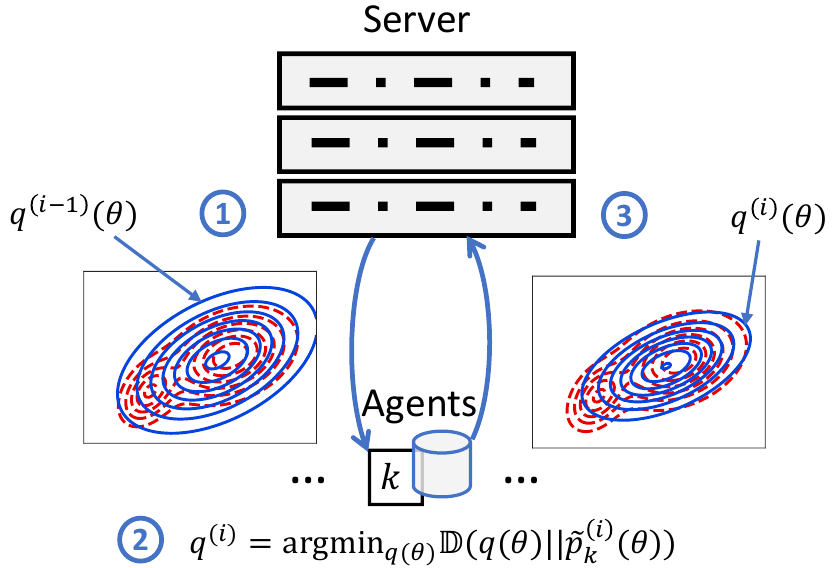}\label{fig:system_model_param}} 
    \subfigure[]{\includegraphics[height=1.75in]{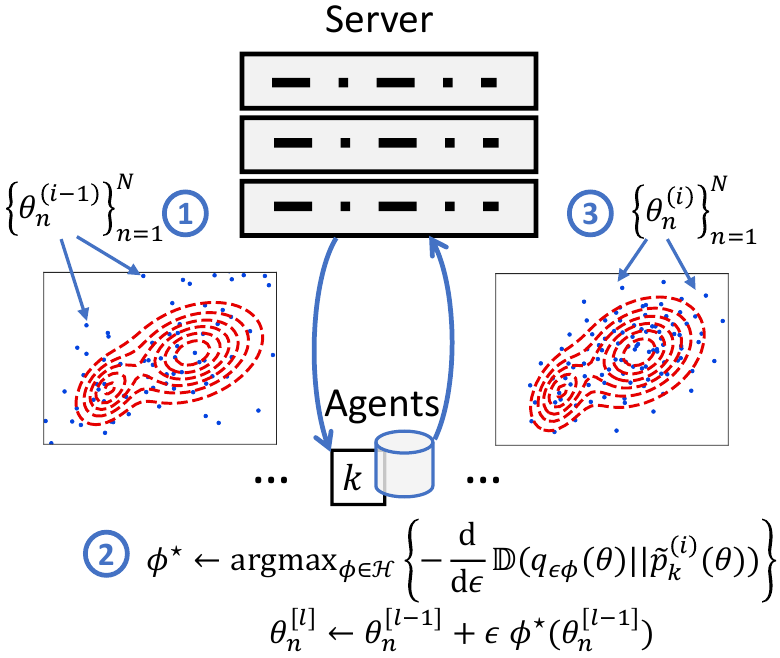} \label{fig:system_model_part}} 

    \caption{Federated learning across $K$ agents equipped with local datasets and assisted by a central server: \textcolor{black}{(a) in \ac{DVI} agents exchange the current model posterior $q^{(i)}(\theta)$ with the server,} while (b) in DSVGD agents exchange particles $\{\theta_n \}_{n=1}^{N}$ providing a non-parametric estimate of the posterior.}

    \label{fig:system_model}
\end{figure}

As illustrated in Fig. \ref{fig:system_model}, \ac{DSVGD} targets a generalized Bayesian learning formulation, with arbitrary loss functions \citep{generalized_VI}; and maintains a number of non-random and interacting particles at a central server to represent the current iterate of the global posterior. At each iteration, the particles are downloaded and updated by one of the agents by minimizing a local free energy functional before being uploaded to the server. \ac{DSVGD} is shown to enable (\emph{i}) a trade-off between per-iteration communication load and number of communication rounds by varying the number of particles; while (\emph{ii}) being able to make trustworthy decisions through Bayesian inference.
\section{System Set-up}
\label{sec:system_set_up}
We consider the federated learning set-up in Fig. \ref{fig:system_model}, where each agent $k=1, \ldots, K$ has a distinct local dataset with associated training loss $L_k(\theta)$ for model parameter $\theta$. The agents communicate through a central node with the goal of computing the global posterior distribution $q(\theta)$ over the shared model parameter $\theta \in \mathbbm{R}^d$ for some prior distribution $p_0(\theta)$ \citep{angelino_patterns_scalable_BI}. Specifically, following the generalized Bayesian learning framework \citep{generalized_VI}, the agents aim at obtaining the distribution $q(\theta)$ that minimizes the \textbf{global free energy} 
\begin{equation}
\min_{q(\theta)} \bigg\{F(q(\theta)) = \sum_{k=1}^{K} \mathbbm{E}_{\theta \sim q(\theta)} [L_k(\theta)] + \alpha \mathbbm{D}(q(\theta)||p_0(\theta)) \bigg\}, \label{eq:main_minimization}
\end{equation}
where $\alpha > 0$ is a temperature parameter. The (generalized, or Gibbs) \textbf{global posterior} $q_{opt}(\theta)$ solving problem (\ref{eq:main_minimization}) must strike a balance between minimizing the sum loss function (first term in $F(q)$) and the model complexity defined by the divergence from a reference prior (second term in $F(q)$). {\color{black} It is given as
\begin{equation}
 q_{opt}(\theta) = \frac{1}{Z} \cdot \Tilde{q}_{opt}(\theta), \text{ with}\ \ \Tilde{q}_{opt}(\theta) = p_0 (\theta) \exp\Big(-\frac{1}{\alpha} \sum_{k=1}^{K} L_k (\theta)\Big), \label{eq:exact_posterior}
\end{equation}}where we denoted as {$Z$ the normalization constant}. It is useful to note that the global free energy can also be written as the scaled KL $F(q(\theta)) = \alpha \mathbbm{D}(q(\theta)||\Tilde{q}_{opt}(\theta)).$

\par The main challenge in computing the optimal posterior $q_{opt}(\theta)$ in a distributed manner is that each agent $k$ is only aware of its local loss $L_k (\theta)$. By exchanging information through the server, the $K$ agents wish to obtain an estimate of the global posterior (\ref{eq:exact_posterior}) without disclosing their local datasets neither to the server nor to the other agents. In this paper, we introduce a novel non-parametric distributed generalized Bayesian learning framework that addresses this challenge by integrating Distributed VI (DVI) and \ac{SVGD} \citep{SVGD}.
\textcolor{black}{\section{Distributed Variational Inference}
\label{sec:distributed_VI}
In this section, we describe a general Expectation Propagation (EP)-based  framework \citep{expectation_propagation_way_life_paratitioned_data}, which we term as \ac{DVI}, that aims at computing the global posterior in a federated fashion \citep{PVI, variational_multi_task}.
\ac{DVI} starts from the observation that the posterior (\ref{eq:exact_posterior}) factorizes as the product 
\begin{equation}
    q(\theta) = p_{0}(\theta) \prod_{k=1}^{K} t_k (\theta), \label{eq:factorized_parametric}
\end{equation}
where the term $t_k(\cdot)$ is given by the scaled local likelihood $\exp(\alpha^{-1} L_k(\theta))/Z^{1/K}$. Since the normalization constant $Z$ depends on all data sets, the true scaled local likelihood $t_k(\cdot)$ cannot be directly computed at agent $k$. The idea of DVI is to iteratively update  \textbf{approximate likelihood} factors $t_k(\theta)$ for $k=1,...,K$ by means of local optimization steps at the agents and communication through the server, with the aim of minimizing the global free energy (\ref{eq:main_minimization}) over distribution (\ref{eq:factorized_parametric}).} \par
\textcolor{black}{
We give here the standard implementation of DVI in which a single agent is schedule at each time, although parallel implementations are possible and discussed below. Accordingly, at each communication round $i=1,2,...$, the server maintains the current iterate $q^{(i-1)}(\theta)$ of the global posterior, and schedules an agent $k \in \{1,2,\ldots,K\}$, which proceeds as follows:
\begin{enumerate}[wide, labelwidth=!, labelindent=0pt, font=\bfseries]
    \item Agent $k$ downloads the current global variational posterior distribution $q^{(i-1)}(\theta)$ from the server (see Fig. \ref{fig:system_model_param}, step \circled{1});
    \item Agent $k$ updates the global posterior by minimizing the local free energy $F_{k}^{(i)}(q(\theta))$ (see Fig. \ref{fig:system_model_param}, step \circled{2})
    \begin{equation}
        q^{(i)}(\theta) = \underset{q(\theta)}{\mathrm{argmin}}\  \bigg\{ F_{k}^{(i)}(q(\theta))=\mathbbm{E}_{\theta \sim q(\theta)} [ L_k (\theta)] + \alpha \mathbbm{D} (q(\theta)||\hat{p}^{(i)}_{k}(\theta)) \bigg\},\  \label{eq:min_local_energy_cavity}
    \end{equation}
where we have defined the (unnormalized) \textbf{cavity distribution} $\hat{p}_{k}^{(i)} (\theta)$ as
\begin{equation}
    \hat{p}_{k}^{(i)} (\theta) = \frac{q^{(i-1)}(\theta)}{t_{k}^{(i-1)}(\theta)}. \label{eq:cavity_dist}
\end{equation}
The cavity distribution $\hat{p}_{k}^{(i)}(\theta)$, which removes the contribution of the current approximate likelihood of agent $k$ from the current global posterior iterate, serves as a prior for the update in (\ref{eq:min_local_energy_cavity}).
In a manner similar to (\ref{eq:exact_posterior}), the local free energy is minimized by the \textbf{tilted distribution} $p_{k}^{(i)}  (\theta) \propto \Tilde{p}_{k}^{(i)} (\theta)$ with
\begin{equation}
    \Tilde{p}_{k}^{(i)} (\theta) = \hat{p}_{k}^{(i)} (\theta) \exp\bigg(-\frac{1}{\alpha} L_k(\theta)\bigg); \label{eq:tilted_dist}
\end{equation}
\item Agent $k$ sends the updated posterior $q^{(i)}(\cdot)=p_k^{(i)}(\cdot)$ to the server (see Fig. \ref{fig:system_model_param}, step \circled{3}), and updates its approximate likelihood accordingly as \vspace{-0.2cm}
\begin{equation}
    t_{k}^{(i)}(\theta) = \frac{q^{(i)}(\theta)}{q^{(i-1)}(\theta)} t_{k}^{(i-1)} (\theta);\vspace{-0.2cm} \label{eq:t_k}
\end{equation}
\end{enumerate}}
\textcolor{black}{Finally, non-scheduled agents $k^{\prime} \neq k$ set $t_{k^\prime}^{(i)}(\theta) = t^{(i-1)}_{k^{\prime}} (\theta)$, and the server sets the next iterate as $q^{(i)}(\theta)$.}
\textcolor{black}{We have the following key property of DVI.
\begin{restatable}{theorem}{stationarypoints}
\label{th:stationary_points}
The global posterior $q_{opt}(\theta)$ in (\ref{eq:exact_posterior}) is the unique fixed point of the DVI algorithm. 
\end{restatable}
The fixed-point property in Theorem \ref{th:stationary_points} can be verified directly by setting $q^{(i-1)}(\theta)=q_{opt}(\theta)$ and $t_{k}^{(i-1)}(\theta)=\exp(\alpha^{-1} L_k(\theta))/Z^{1/K}$ and by observing that this leads to the fixed point condition $q^{(i)}(\theta)=q^{(i-1)}(\theta)=q_{opt}(\theta)$. The proof is provided in Sec. \ref{app:theoretical_analysis}. Importantly, this property is not tied to the sequential implementation detailed above, and it applies also if multiple devices are scheduled in parallel, as long as one sets the next iterate as $ q^{(i)}(\theta) = p_0(\theta) \prod_{k \in \mathcal{K}^{(i)}} t_{k}^{(i)}(\theta) \prod_{k^\prime \not\in \mathcal{K}^{(i)}} t_{k^\prime}^{(i)}(\theta)$, where $\mathcal{K}^{(i)}$ denotes the set of scheduled agents at communication round $i$ and we have $t_{k^\prime}^{(i)} (\theta) = t_{k^\prime}^{(i-1)} (\theta) $ and $t_{k}^{(i)} (\theta) $ updated following (\ref{eq:t_k}).
}
\textcolor{black}{\section{Preliminaries}
\label{sec:background}
In this section, we briefly review PVI, which serves as an important benchmark, and SVGD, on which we build the proposed Bayesian federated learning solution.
\subsection{Partitioned Variational Inference}
The exact minimization of the local free energy function (\ref{eq:min_local_energy_cavity}) assumed by DVI is often not tractable. To address this problem, in its most typical form, \ac{PVI} constrains the local free energy minimization (\ref{eq:min_local_energy_cavity}) to the space of parametric distributions that factorize as $q(\theta|\eta) = p_{0}(\theta| \eta_0) \prod_{k=1}^{K} t_k (\theta|\eta_k),$
where prior $p_0(\cdot|\eta_0)=\mathrm{ExpFam}(\cdot|\eta_0)$ and approximate likelihood $t_k(\cdot | \eta_k)=\mathrm{ExpFam}(\cdot|\eta_k)$ are selected from the same exponential-family distribution, with natural parameters $\eta_0$ and $\eta_k$, respectively. \ac{PVI} follows the same steps as \ac{DVI} with the caveat that the local free energy (\ref{eq:min_local_energy_cavity}) for agent $k$ is minimized over the natural parameter $\eta$. This can be done efficiently, albeit approximately, using for e.g., natural gradient descent \citep{natural_gradient_amari}.} 

\textcolor{black}{
The bias imposed by the parametrization in PVI  significantly affects the quality of the approximation of the obtained posterior $q(\theta)$ with respect to the true global posterior $q_{opt}(\theta)$ in the presence of model misspecification. In this case, the fixed point property in Theorem 1 no longer applies.}
\subsection{Stein Variational Gradient Descent (SVGD)}
\ac{SVGD} tackles the minimization of the (scaled) free energy functional $\mathbbm{D}(q(\theta) || \Tilde{p}(\theta))$, for an unnormalized target distribution $\Tilde{p}(\theta)$, over a non-parametric generalized posterior $q(\theta)$ defined over the model parameters $\theta \in \mathbbm{R}^{d}$. The posterior $q(\theta)$ is represented by a set of particles $\{\theta_n \}_{n=1}^{N}$, with $\theta_n \in \mathbbm{R}^d$. In practice, an approximation of $q(\theta)$ can be obtained from the particles $\{ \theta_n\}_{n=1}^{N}$ through a \ac{KDE} as $q(\theta) = N^{-1} \sum_{n=1}^{N} \mathrm{K}(\theta, \theta_n)$ for some kernel function $\mathrm{K}(\cdot, \cdot)$ \citep{bishop_book_pattern_recognition}. The particles are iteratively updated through a series of transformations that are optimized to minimize the free energy. The transformations are restricted to lie within the unit ball of a \ac{RKHS} $\mathcal{H}^d=\mathcal{H} \times \ldots \times \mathcal{H}$. It is shown by \citet{SVGD} that  
this optimization yields the SVGD update \vspace{-0.2cm}
\begin{equation}
    \theta^{[l]}_{n} \xleftarrow[]{} \theta_{n}^{[l-1]} + \frac{\epsilon}{N} \sum_{j=1}^{N} [\mathrm{k}(\theta_{j}^{[l-1]}, \theta_{n}^{[l-1]}) \nabla_{\theta_j} \log \Tilde{p}(\theta_{j}^{[l-1]}) + \nabla_{\theta_j} \mathrm{k}(\theta_{j}^{[l-1]}, \theta_{n}^{[l-1]})] \label{eq:particles_gradient}
\end{equation}
for $n=1, \ldots, N$, where $\mathrm{k}(\cdot,\cdot)$ is the positive definite kernel associated with \ac{RKHS} $\mathcal{H}$. The first term in the update (\ref{eq:particles_gradient}) drives the particles towards the regions of the target distribution $\Tilde{p}(\theta)$ with high probability, while the second term drives the particles away from each other, encouraging exploration in the model parameter space. It is known that, in the asymptotic limit of a large number $N$ of particles, the empirical distribution encoded by the particles $\{\theta_{n}^{[l]}\}_{n=1}^{N}$ converges to the normalized target distribution $p(\theta) \propto \tilde{p}(\theta)$ \citep{svgd_gradient_flow}.
\section{Distributed Stein Variational Gradient Descent}
\label{sec:DSVGD}
In this section, we introduce \ac{DSVGD}, a novel distributed algorithm that tackles the generalized Bayesian inference problem (\ref{eq:main_minimization}) via DVI over a non-parametric particle-based representation of the global posterior. As illustrated in Fig. \ref{fig:system_model_part}, \ac{DSVGD} is based on the iterative optimization of local free energy functionals (\ref{eq:min_local_energy_cavity}) via \ac{SVGD} (see Sec. \ref{sec:background}), and on the exchange of particles between the central server and agents. \textcolor{black}{Given the flexibility of the non-parametric form of the posterior, DSVGD doesn't suffer from the bias caused by the parametrization assumed by PVI. As a result, in the limit of a sufficiently large number of particles, DSVGD benefits from the fixed point property of DVI stated in Theorem 1, recovering the true global posterior as a fixed point of its iterations. Furthermore, as we will discuss, DSVGD enables devices to exchange more informative messages regarding the current iterate of the posterior by increasing the number of particles. This can in turn reduce the number of communication rounds and the overall communication load to convergence, at the cost of a larger per-round load. In this regard, we note that, in practice, a small number of particles is sufficient to obtain state-of-the-art performance \citep{SVGD}, as verified in Sec. \ref{sec:experiments}.}\par

In order to facilitate the presentation, we first introduce a simpler version of \ac{DSVGD} that has the practical drawback of requiring each agent to store a number of particles that increases linearly with the number of iterations in which the agent is scheduled. Then, we present a more practical algorithm, for which the memory requirements do not scale with the number of iterations as each agent must only memorize a set of $N$ local particles across different iterations. Algorithmic table for \ac{U-DSVGD} in addition to discussions on complexity and convergence, can be found respectively in Sec. \ref{app:udsvgd} and Sec. \ref{app:complexity_convergence} in the supplementary materials. \textcolor{black}{A direct extension of \ac{DSVGD}, termed Parallel-DSVGD (P-DSVGD), where multiple agents are scheduled per round can be found in Sec. \ref{sec:P_DSVGD} of the Appendix}.
\subsection{U-DSVGD}
\label{sec:u-dsvgd}
In this section, we present a simplified \ac{DSVGD} variant, which we refer to as \ac{U-DSVGD}. We follow the standard implementation of DVI with a single agent $k$ scheduled at each communication round $i=1, 2, \ldots$, although, as discussed, parallel implementations are also possible. Let us define as $\mathcal{I}_{k}^{(i)} \subseteq \{1,\ldots,i \}$ the subset of rounds at which agent $k$ is scheduled prior, and including, iteration $i$. At the beginning of each round $i$, the server maintains the iterate of the current global particles $\{\theta^{(i-1)}_{n}\}_{n=1}^{N}$, while each agent $k$ keeps a local buffer of particles $\{\theta_{n}^{(j-1)}, \theta_{n}^{(j)} \}_{n=1}^{N}$ for all previous rounds $j \in \mathcal{I}_{k}^{(i-1)}$ at which agent $k$ was scheduled. The growing memory requirements at the agents will be dealt with by the final version of \ac{DSVGD} to be introduced in Sec. \ref{sec:dsvgd}. Furthermore, as illustrated in Fig. \ref{fig:system_model_part}, at each iteration $i$, \ac{U-DSVGD} schedules an agent $k \in \{1,2, \ldots, K \}$ and carries out the following steps. \vspace{-0.1cm}
\begin{enumerate}[wide, labelwidth=!, labelindent=0pt, font=\bfseries]
    \item Agent $k$ downloads the current global particles $\{\theta^{(i-1)}_{n}\}_{n=1}^{N}$ from the server (see Fig. \ref{fig:system_model_part}, step \circled{1}) and includes them in the local buffer.
    \item Agent $k$ updates each downloaded particle as 
    \begin{equation}
        \theta_{n}^{[l]} \xleftarrow[]{} \theta_{n}^{[l-1]} +  \epsilon\phi (\theta_{n}^{[l-1]}),\ \text{for}\ l=1,\ldots,L, \label{eq:initial_update_udsvgd}
\end{equation}
\begin{algorithm}[t!]
\small
\LinesNumbered
{\color{black}
\KwIn{prior $p_{0}(\theta)$, local loss functions $\{L_{k}(\theta)\}_{k=1}^{K}$, temperature $\alpha > 0$, kernels $\mathrm{K}(\cdot, \cdot)$ and $\mathrm{k}(\cdot, \cdot)$ }
\KwOut{global approximate posterior $q(\theta) = N^{-1} \sum_{n=1}^{N} \mathrm{K}(\theta, \theta_n)$}
\vspace{0.1cm}
\hrule
\vspace{0.1cm}
{\bf initialize} $q^{(0)} (\theta) = p_0(\theta)$; $\{\theta_{n}^{(0)} \}_{n=1}^{N} \overset{\text{i.i.d}}{\sim} p_0 (\theta)$; $ \{\theta_{k, n}^{(0)} = \theta_{n}^{(0)} \}_{n=1}^{N}$ and $t_{k}^{(0)}(\theta) = 1$ for $k=1, \ldots, K$ \\
\For{{\em $i=1,\ldots, I$} }{
\textcolor{red}{Server} schedules an Agent $k$\\
\textcolor{blue}{Agent} $k$ downloads current global particles $\{\theta_{n}^{(i-1)} \}_{n=1}^{N}$ from \textcolor{red}{server} \\

\textcolor{blue}{Agent} $k$ obtains updated global particles $\{ \theta_{n}^{(i)}\}_{n=1}^{N}$ using (\ref{eq:particles_updates_udsvgd}), $\{\theta_{n}^{(i-1)} \}_{n=1}^{N}$ and $\{ \theta_{k,n}^{(i-1)}\}_{n=1}^{N}$ \\

\textcolor{blue}{Agent} $k$ sends the updated global particles $\{\theta_{n}^{(i)} \}_{n=1}^{N}$ to the \textcolor{red}{server}\\

\textcolor{blue}{Agent} $k$ carries distillation to obtain $\{ \theta_{k,n}^{(i)}\}_{n=1}^{N}$ encoding $t_{k}^{(i)}(\theta)$ using (\ref{eq:nu_updates_dsvgd}) and $\{ \theta_{n}^{(i)}\}_{n=1}^{N}$\\

}
\Return $q(\theta) = N^{-1}\sum_{n=1}^{N} \mathrm{K}(\theta, \theta_{n}^{(I)})$
}
\caption{\textcolor{black}{Distributed Stein Variational Gradient Descent (DSVGD)}}
\label{algo1}
\end{algorithm}
where $L$ is the number of local iterations; $[l]$ denotes the \textit{local iteration} index; we have the initialization $\theta^{[0]}_{n} = \theta^{(i-1)}_{n}$; and the function $\phi(\cdot)$ is to be optimized within the unit ball of a \ac{RKHS} $\mathcal{H}^d$. The function $\phi(\cdot)$ is specifically optimized to maximize the steepest descent decrease of a particle-based approximation of the local energy (\ref{eq:min_local_energy_cavity}). To elaborate, we denote as $q^{(i-1)}(\theta) = \sum_{n=1}^{N} \mathrm{K}(\theta, \theta_{ n}^{(i-1)})$ the \ac{KDE} of the current global posterior iterate encoded by particles $\{\theta^{(i-1)}_{ n}\}_{n=1}^{N}$. Adopting the factorization (\ref{eq:factorized_parametric}) for the global posterior (cf. (\ref{eq:t_k})), we define the current local approximate likelihood \vspace{-0.2cm}
\begin{equation}
    t_{k}^{(i-1)}(\theta) = \prod_{j \in \mathcal{I}_{k}^{(i-1)}} \frac{q^{(j)}(\theta)}{q^{(j-1)}(\theta)}= \frac{q^{(i-1)}(\theta)}{q^{(i-2)}(\theta)} t_{k}^{(i-2)}(\theta). \label{eq:t_part_udsvgd} \vspace{-0.2cm}
\end{equation}
Note that (\ref{eq:t_part_udsvgd}) can be computed using \textit{all} the particles in the buffer at agent $k$ at iteration $i$. Finally, the (unnormalized) tilted distribution $\tilde{p}_{k}^{(i)}$ (cf. (\ref{eq:tilted_dist})) is written as
\begin{equation}
    \Tilde{p}_{k}^{(i)} (\theta) = \frac{q^{(i-1)} (\theta)}{t_{k}^{(i-1)} (\theta)} \exp\bigg(-\frac{1}{\alpha} L_k(\theta)\bigg). \label{eq:tilted_udsvgd}
\end{equation}
Following \ac{SVGD}, the update (\ref{eq:initial_update_udsvgd}) is optimized to maximize the steepest descent decrease of the \ac{KL} divergence between the approximate global posterior $q_{\epsilon \phi}^{[l]}(\theta)$ encoded via particles $\{\theta^{[l]}_{n}\}_{n=1}^{N}$ and the tilted distribution $\Tilde{p}_{k}^{(i)}(\theta)$ in (\ref{eq:tilted_udsvgd}) (see Fig. \ref{fig:system_model_part}, step \circled{2}), i.e.,\vspace{-0.2cm}
\begin{equation}
    \phi^\star(\cdot) \xleftarrow[]{} \argmax_{\phi(\cdot) \in \mathcal{H}^d} \bigg\{ - \frac{d}{d\epsilon} \mathbbm{D}(q_{\epsilon \phi}^{[l-1]} (\theta)||\Tilde{p}_{k}^{(i)}(\theta)),\ \ \ \text{s.t.}\ \ \ ||\phi||_{\mathcal{H}^d} \leq 1  \bigg\}.
\end{equation}
Thus, recalling (\ref{eq:particles_gradient}), the particles are updated as
\begin{equation}
        \theta^{[l]}_{n} \xleftarrow[]{} \theta_{n}^{[l-1]} + \frac{\epsilon}{N} \sum_{j=1}^{N} [\mathrm{k}(\theta_{j}^{[l-1]}, \theta_{n}^{[l-1]}) \nabla_{\theta_j} \log \Tilde{p}_{k}^{(i)}(\theta_{j}^{[l-1]}) \!\!+ \!\!\nabla_{\theta_j} \mathrm{k}(\theta_{j}^{[l-1]}, \theta_{n}^{[l-1]})],\! \text{for}\ l\!\!=\!\!1,\ldots, \!\!L.\vspace{-0.4cm}
        \label{eq:particles_updates_udsvgd}
\end{equation}

\item Agent $k$ sets $\theta^{(i)}_{n} = \theta^{[L]}_{n}$ for $n=1, \ldots, N$. Particles $\{\theta_{n}^{(i)} \}_{n=1}^{N}$ are added to the buffer and sent to the server (see Fig. \ref{fig:system_model_part}, step \circled{3}) that updates the current global particles as $\{\theta_n \}_{n=1}^{N} = \{ \theta_{n}^{(i)}\}_{n=1}^{N}$.
\end{enumerate}
In order to implement the described \ac{U-DSVGD} algorithm, we need to compute the gradient in (\ref{eq:particles_updates_udsvgd}) at agent $k$. First, by (\ref{eq:tilted_udsvgd}), we have
\begin{equation}
    \nabla_{\theta} \log \Tilde{p}_{k}^{(i)} (\theta) = \nabla_{\theta} \log q^{(i-1)}(\theta) - \nabla_{\theta} \log t_{k}^{(i-1)} (\theta)  - \frac{1}{\alpha} \nabla_{\theta} L_k(\theta). \label{eq:gradient_log_tilted_udsvgd}
\end{equation}
Using (\ref{eq:t_part_udsvgd}), the second gradient term can be obtained in a recursive manner using the local buffer as
\begin{equation}
    \nabla_{\theta} \log  t_{k}^{(i-1)} (\theta) = \begin{cases}
        \nabla_{\theta} \log t_{k}^{(i-2)} (\theta)\ \text{if agent $k$ not scheduled at iteration $(i-1)$}\\
    \nabla_{\theta}\log t_{k}^{(i-2)} (\theta) + \nabla_{\theta} \log q^{(i-1)} (\theta) - \nabla_{\theta} \log q^{(i-2)} (\theta) \ \text{otherwise}.
    \end{cases} \label{eq:grad_log_t_udsvgd}
\end{equation}
Finally, the gradients $\nabla_{\theta} \log q^{(j)} (\theta)$ can be directly computed from the \ac{KDE} expression of $q^{(j)}(\theta)$, with initializations $t^{(0)}(\theta) = 1$ and $q^{(0)}(\theta) = p_0(\theta)$.
\begin{figure}[t]
    \centering
    \subfigure{\includegraphics[height=0.79in]{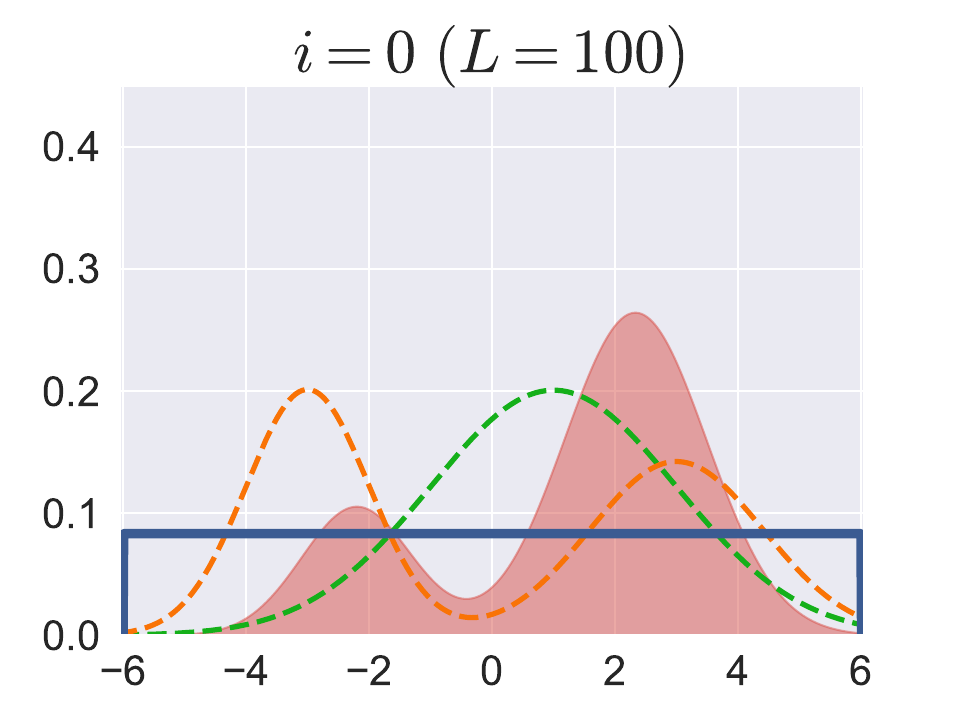}}
    \subfigure{\includegraphics[height=0.79in]{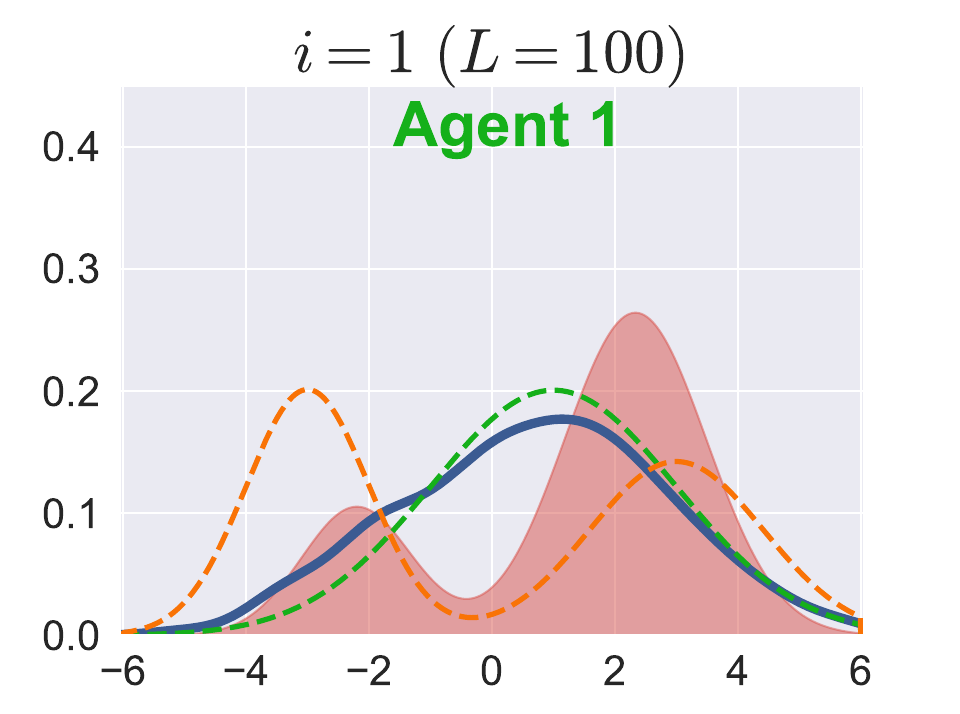}} 
    \subfigure{\includegraphics[height=0.79in]{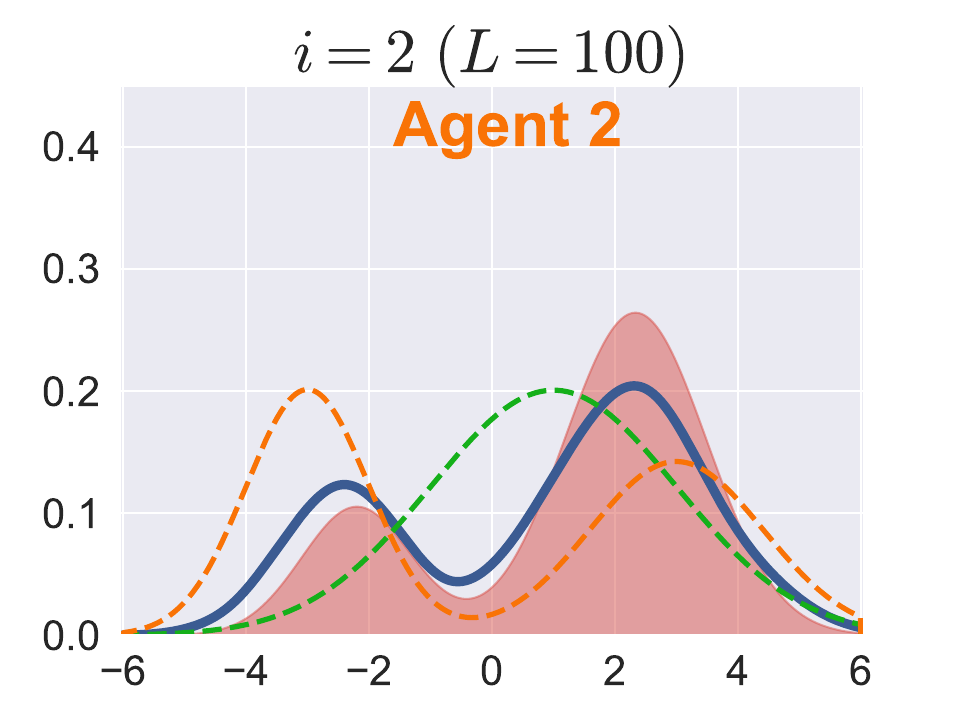}} 
    \subfigure{\includegraphics[height=0.79in]{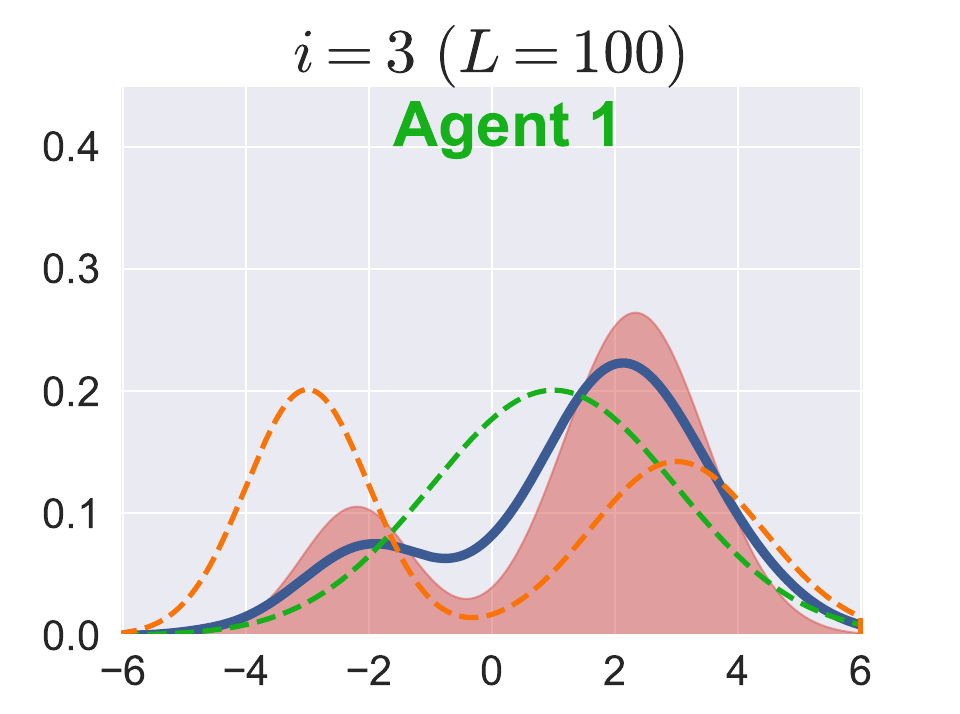}}
    \subfigure{\includegraphics[height=0.79in]{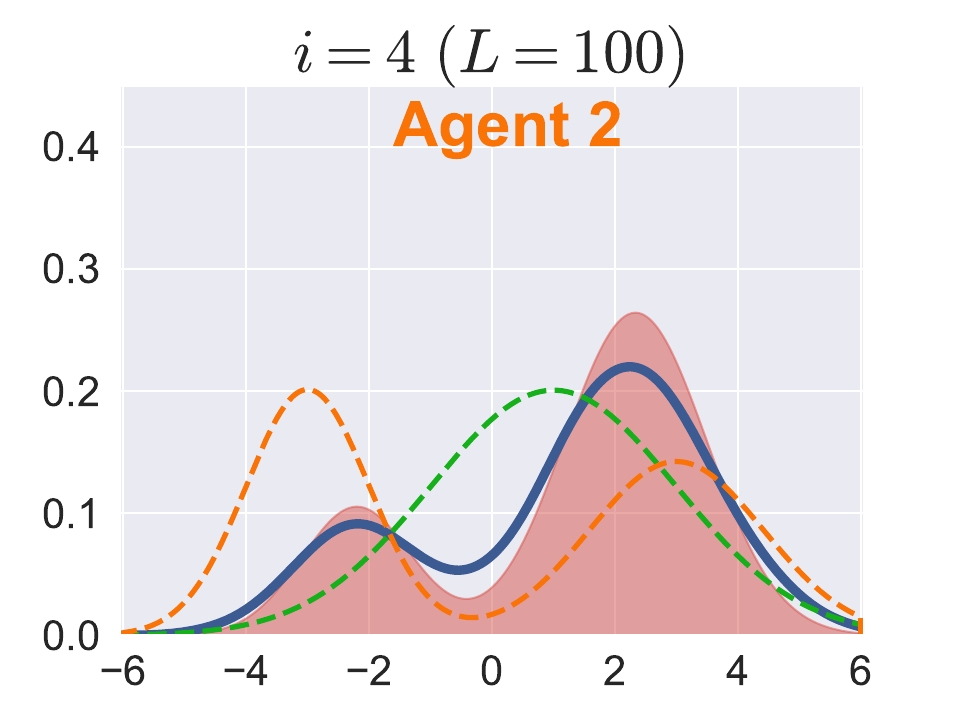}} 
    \caption{Gaussian toy example with uniform prior and $K=2$. Dashed lines represent local posteriors, the shaded area represents the true global posterior, while the solid black line is the approximate posterior obtained using a \ac{KDE} over the particles.\ac{DSVGD} schedules agent $1$ and $2$ at odd and even number of communication rounds $i$, respectively.}
    \label{fig:mixture_toy}
\end{figure}
\textcolor{black}{
The inner loop of U-DSVGD inherits the asymptotic convergence properties of SVGD in terms of local free energies, but existing results do not imply that the global free energy decreases across the iterations. This result is provided in the next theorem, whose precise formulation can be found in Sec. \ref{app:theoretical_analysis} of the Appendix. 
\begin{restatable}[Guaranteed per-iteration decrease of the global free energy.]{theorem}{globalenergydecrease}
\label{th:global_energy_decrease}
The decrease in the global free energy from local iteration $l$ to  $l+1$ during communication round $i$ for which agent $k$ is scheduled can be lower bounded as
\begin{equation}
    F(q^{[l]}(\theta)) -  F(q^{[l+1]}(\theta)) \geq  \alpha \epsilon S(q^{[l]}, p_{k}^{(i)})(1- \epsilon  \gamma ) - 2 \alpha (K-1) l_{\mathrm{max}}^{(i)} \sqrt{2 \mathbb{D}(q^{[l+1]}||q^{[l]})}, \label{eq:bound_energy}
\end{equation}
where $l_{\mathrm{max}}^{(i)}= \underset{\theta}{\mathrm{sup}}\ \underset{m \neq k}{\max}| \log ( t_{m}^{(i-1)}(\theta))\cdot\exp(\frac{1}{\alpha} L_m (\theta)) |$, $S(q,p)$ denotes the Kernalized Stein Discrepancy between distributions $q$ and $p$ \citep{kernelized_stein_discrepancy}, and $\gamma$ is a constant depending on the \ac{RKHS} kernel and the target distribution.
\end{restatable}
The first term in bound (\ref{eq:bound_energy}) quantifies the decrease in the local free energy at agent $k$, which depends on the ``distance'' between current iterate $q^{[l]}$ and the local target given by the tilted distribution $p_{k}^{(i)}(\theta)$; while the second term quantifies the effect of the update on the local free energies of other agents. In the presence of only one agent, the second terms reduce to zero, and one recovers the upper bound on the guaranteed per-iteration improvement for \ac{SVGD} derived in \citet{korba_svgd_nonasymptotic}.}
\label{sec:dsvgd}
\subsection{DSVGD}
\label{sec:dsvgd}
In this section, we describe the final version of \ac{DSVGD}, which, unlike \ac{U-DSVGD}, requires each agent $k$ to maintain only $N$ \textit{local} particles $\{ \theta^{(i)}_{k,n}\}_{n=1}^{N}$ across the communication rounds $i=1,2,\ldots$. To this end, in each round $i$, at the end of the $L$ local \ac{SVGD} updates in (\ref{eq:particles_updates_udsvgd}), \ac{DSVGD} carries out a form of model \textbf{distillation} \citep{hinton_distilling_knowledge_NN, feddistill} via \ac{SVGD}. Specifically, $L^{\prime}$ additional \ac{SVGD} steps are used to approximate the term $t_{k}^{(i)}(\theta) $ using the $N$ local particles $\{\theta_{k,n}^{(i)} \}_{n=1}^{N}$. It is noted that this approximation step is not necessarily harmful to the overall performance, since describing the factor $t_{k}^{(i)}(\theta)$ with fewer particles can have a denoising effect acting as a regularizer.
\par
\ac{DSVGD} operates as \ac{U-DSVGD} apart from the computation of the gradient in (\ref{eq:gradient_log_tilted_udsvgd}) and the management of the local particle buffers. The key idea is that, instead of using the recursion (\ref{eq:grad_log_t_udsvgd}) to compute (\ref{eq:gradient_log_tilted_udsvgd}), \ac{DSVGD} computes the gradient $\nabla_{\theta} \log t_{k}^{(i-1)}(\theta)$ from the \ac{KDE} $t_{k}^{(i-1)}(\theta) = \sum_{n=1}^{N} \mathrm{K}(\theta, \theta_{k,n}^{(i-1)})$ based on the local particles $\{ \theta_{k,n}^{(i-1)}\}_{n=1}^{N}$ in the buffer. At the end of each round $i$, the local particles $\{\theta_{k,n}^{(i-1)} \}_{n=1}^{N}$ are updated by running $L^{\prime}$ local \ac{SVGD} iterations with target given by the updated local factor $t_{k}^{(i)}(\theta) = \frac{q^{(i)}(\theta)}{q^{(i-1)}(\theta)} t_{k}^{(i-1)}(\theta)$. This amounts to the updates
\begin{equation}
        \theta_{k,n}^{[l^{\prime}]} \xleftarrow[]{} \theta_{k, n}^{[l^{\prime}-1]} + \frac{\epsilon^\prime}{N} \sum_{j=1}^{N} [k(\theta_{k, j}^{[l^{\prime}-1]}, \theta_{k, n}^{[l^{\prime}-1]}) \nabla_{\theta_j} \log t_{k}^{(i)}(\theta) + \nabla_{\theta_j} k(\theta_{k, j}^{[l^{\prime}-1]}, \theta_{k,n}^{[l^{\prime}-1]})],  \label{eq:nu_updates_dsvgd}
\end{equation}
for $l^{\prime}=1,\ldots,L^\prime$ and some learning rate $\epsilon^\prime$, where the gradient $\nabla_{\theta} \log t_{k}^{(i)}(\theta) = \nabla_{\theta} \log q^{(i)}(\theta) + \nabla_{\theta} \log t_{k}^{(i-1)}(\theta) - \nabla_{\theta} \log q^{(i-1)}(\theta)$ can be directly computed using \ac{KDE} based on the available particles $\{\theta_{n}^{(i)} \}_{n=1}^{N}$  (updated global particles), $\{\theta_{k, n}^{(i-1)} \}_{n=1}^{N}$  (local particles) and $\{\theta_{n}^{(i-1)} \}_{n=1}^{N}$ (downloaded global particles). Finally, we note that the distillation operation can be performed after sending the updated global particles to the server and thus enabling pipelining of the $L^{\prime}$ local iterations with operations at the server and other agents. \textcolor{black}{DSVGD is summarized in Algorithm \ref{algo1}.} \par
\begin{wrapfigure}[12]{R}{0.3\textwidth}
\vspace{-6mm}
    \centering
    \vspace{-1.45\intextsep}
    \subfigure{{\includegraphics[height=1.3 in, width=1.7 in]{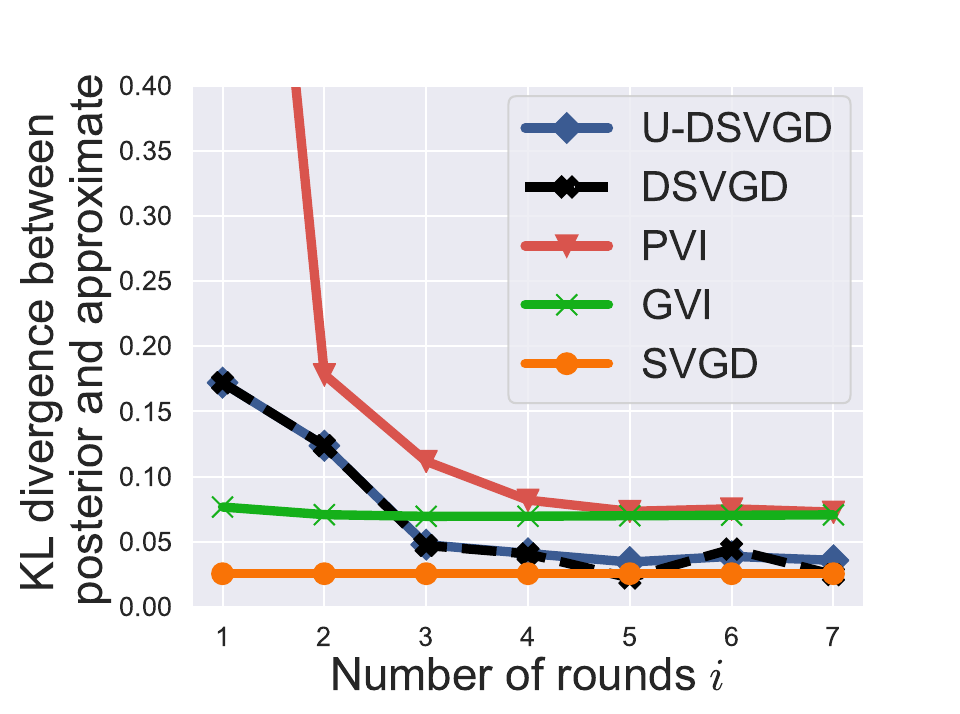}}}
    \BlankLine
    \vspace{-0.75 cm}
    \BlankLine
    \caption{\textcolor{black}{KL divergence between exact and approximate global posteriors as function of the number of rounds $i$ ($L=L^\prime=200$).}}
    \label{fig:KL_global_iter_main}
\end{wrapfigure}
\section{Related Work}
\label{sec:realted_work}
\textbf{Extensions of \ac{SVGD}.} Since its introduction, SVGD has been extended in various directions. Most related to this work is \citet{message_passing_SVGD}, which introduces a message-passing \ac{SVGD} solution for high-dimensional latent parameter spaces by leveraging conditional independence properties in the variational posterior; and  \citet{Bayesian_agnostic_Meta_learning}, which uses SVGD as the per-task base learner in a meta-learning algorithm approximating Expectation Maximization. \par
\textbf{Generalized Bayesian Inference. }Owing to its reliance on point estimates in the model parameter space, frequentist learning methods, such as \ac{FedSGD}, \ac{FedAvg} and their extensions \citep{fedpd, fedprox, fedsplit, fast_convergent_fl, wang2020tackling} are limited in their capacity to combat overfitting and quantify uncertainty \citep{NN_calibration, uncertainty_validity_BNN, neal_bayesian_nn, hands_on_BNN_tutorial, david_mackay_book}. This contrasts with the generalized Bayesian inference framework that produces distributional, rather than point, estimates by
optimizing the free energy functional, which is a theoretically principled bound on the generalization performance \citep{IT_bounds_statistics_estimation, generalized_VI}. Practical algorithms for generalized Bayesian inference can leverage computationally efficient scalable solutions based on either MC sampling or VI methods \citep{angelino_patterns_scalable_BI,properties_gibbs}. 
\begin{figure}[t]
    \centering
    \vspace{-1.3\intextsep}
    \subfigure{\includegraphics[height=1.05 in]{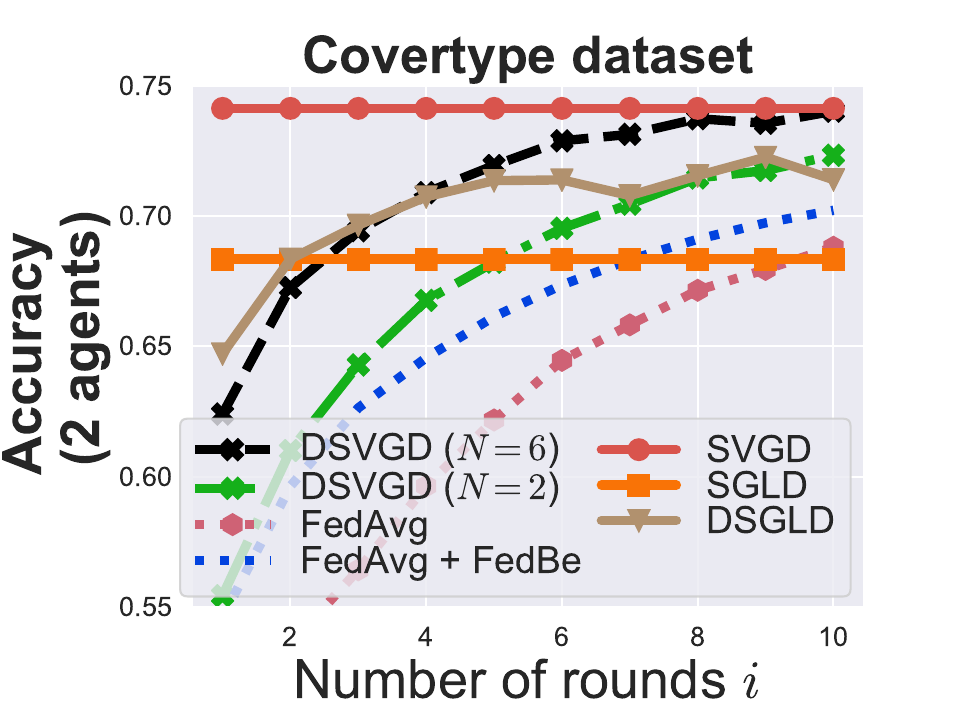}\hspace{-0.3cm}\includegraphics[height=1.05 in]{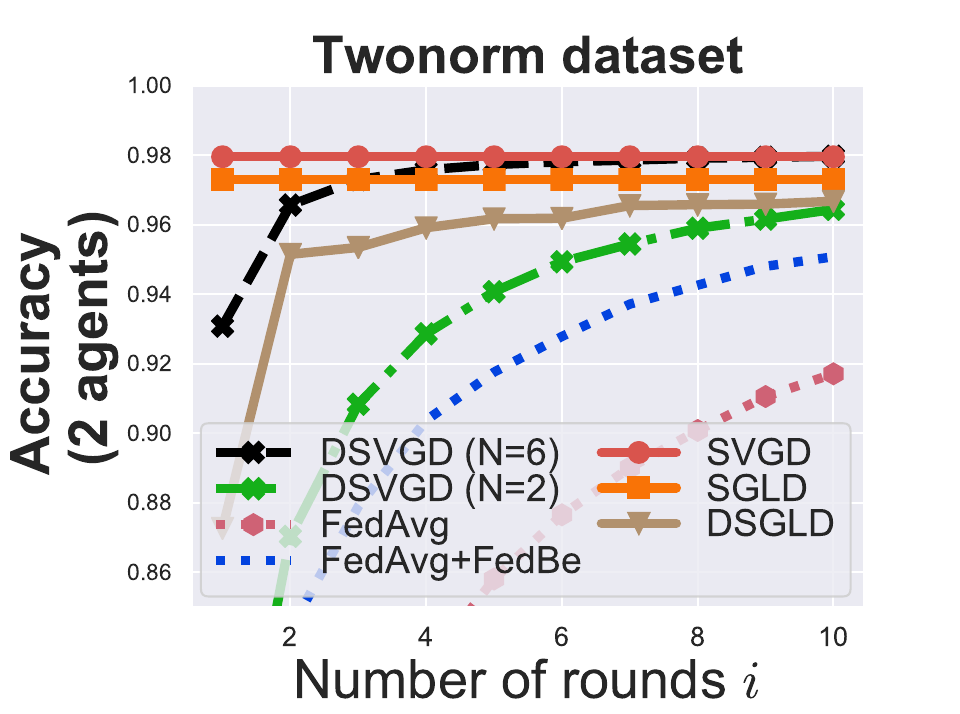}} 
    \mbox{%
  \vline height 16ex}
    \subfigure{\includegraphics[height=1.05 in]{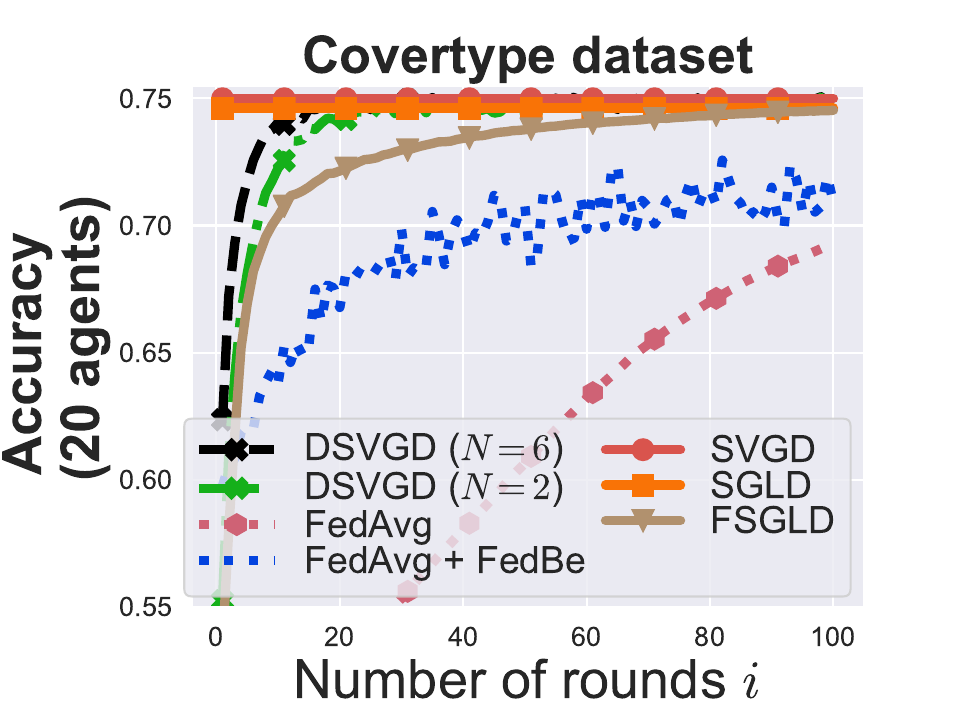}\hspace{-0.3cm}\includegraphics[height=1.05 in]{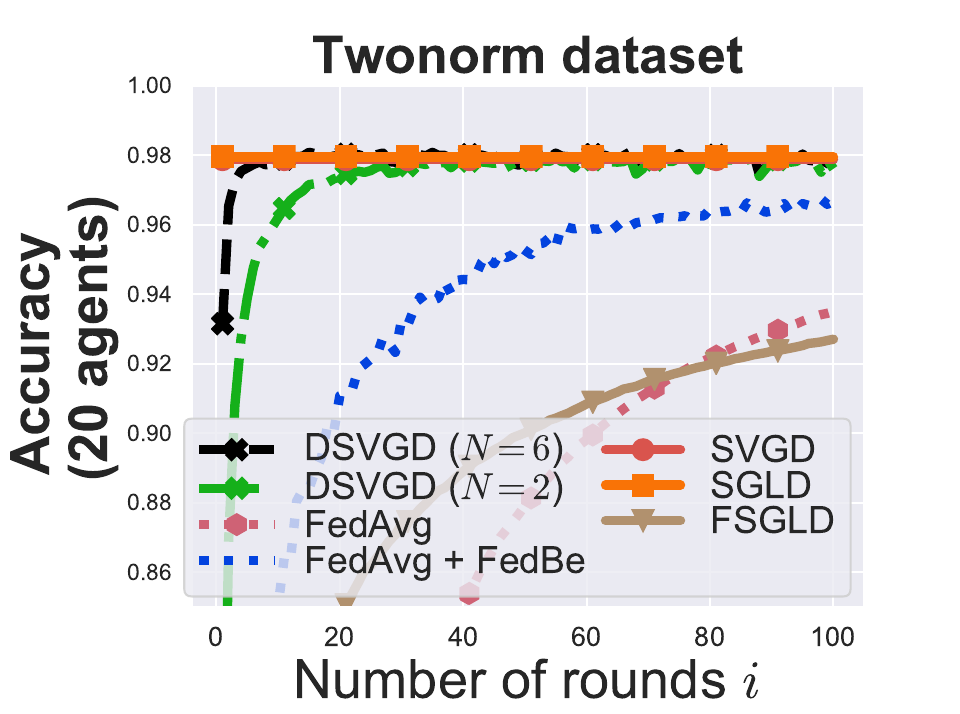}} 

    \caption{Accuracy for Bayesian logistic regression with (left) $K=2$ agents and (right) $K=20$ agents as function of the number of communication rounds $i$ ($N=6$ particles, $L=L^{\prime}=200$).}

    \label{fig:BLR_i_acc}
\end{figure}
\par 
\textbf{Distributed MC Sampling.} The design of algorithms for distributed Bayesian learning has been so far mostly focused on one-shot, or ``embarrassingly parallel'', solutions under ideal communications \citep{comm_efficient_statistical_inference}. These implement distributed \ac{MC} ``consensus'' protocols, whereby samples from the global posterior are approximately synthesized by combining particles from local posteriors \citep{bayes_and_big_data, distributed_estimation_expo}. Iterative extensions, such as Weierstrass sampling \citep{parallel_MCMC_weierstrass, global_consensus_MC}, impose consistency constraints across devices and iterations in a way similar to the Alternating Direction Method of Multipliers (ADMM) \citep{angelino_patterns_scalable_BI}. State-of-the-art results have been obtained via \ac{DSGLD} \citep{dsgld}.
\par
\textbf{Distributed VI Learning.} Considering first one-shot model fusion of local models, Bayesian methods have been used to deal with parameter invariance and weight matching \citep{Bayesian_NP_FL_NN, model_fusion_KL}. Iterative VI such as streaming variational Bias (SVB) \citep{streaming_variational_bayes_tamara} provide a \ac{VI}-based framework for the exponential family to combine local models into global ones. \ac{PVI} provides a general framework that can implement SVB, as well as online \ac{VI} \citep{PVI} and has been extended to multi-task learning in \citet{variational_federated_MTL}.
\section{Experiments} 
\label{sec:experiments}
As in \citet{SVGD}, for all our experiments with \ac{SVGD} and
\ac{DSVGD}, we use the \ac{RBF} kernel $\mathrm{k}(x, x_0) = \exp(-||x - x_0 ||^{2}_{2}/h)$. The bandwidth $h$ is adapted to the set of particles used in each update by setting $h = \mathrm{med}^2/\log n$, where $\mathrm{med}$ is the median of the pairwise distances between the particles in the current iterate. The Gaussian kernel $\mathrm{K}(\cdot, \cdot)$ used for the KDEs has a bandwidth equal to $0.55$. Unless specified otherwise, we use AdaGrad with momentum to choose the learning rates $\epsilon$ and $\epsilon^\prime$ for (U-)DSVGD. Throughout, we fix the temperature parameter $\alpha = 1$ in (\ref{eq:main_minimization}). Finally, to ensure a fair comparison with distributed schemes, we run centralized schemes for the same total number $I\times L$ of iterations across all experiments. Additional results for all experiments can be found in Appendix \ref{app:additional_experiments} in the supplementary  materials, which include also additional implementation details.
\par
\begin{wrapfigure}[12]{R}{0.3\textwidth}
    \vspace{-3mm}
    \centering
    \vspace{-1.8\intextsep}
    \subfigure{{\includegraphics[height=1.3 in, width=1.7 in]{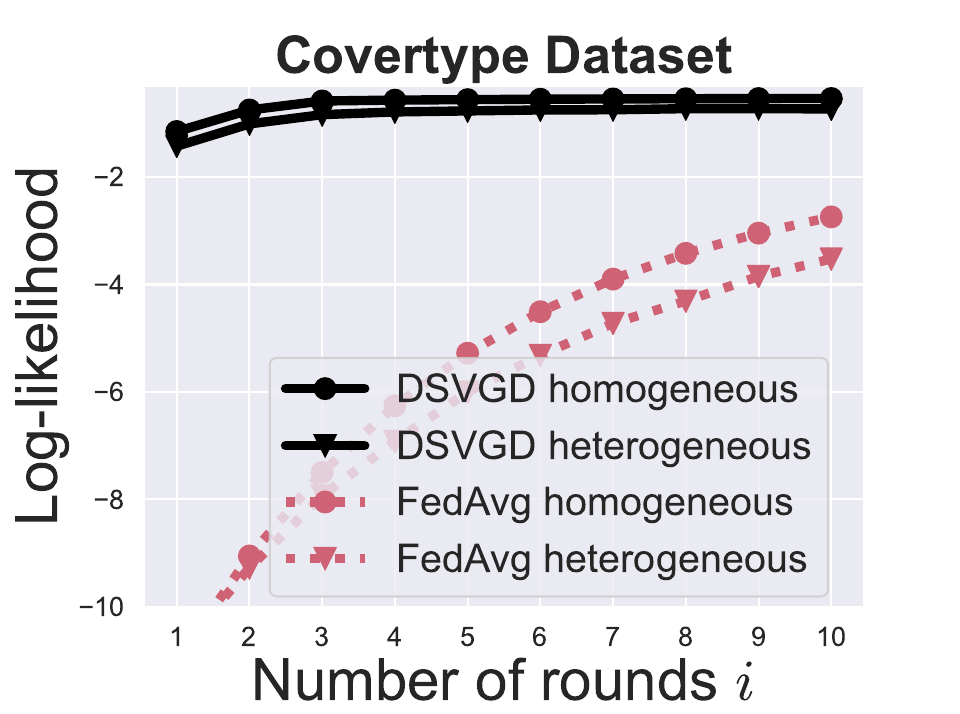}}}
    \BlankLine
    \vspace{-0.75 cm}
    \BlankLine
    \caption{\textcolor{black}{Log-likelihood for Bayesian logistic regression with non-iid data distributions ($N=6$, $L=L^{\prime}=200$).}}
    \label{fig:llh_heterogeneous}
\end{wrapfigure}
\textbf{Gaussian 1D mixture toy example. }
We start by considering a simple one-dimensional mixture model in which the local unnormalized local posteriors $p_{k}(\theta) = p_0(\theta) \exp(-\alpha^{-1} L_k (\theta))$ at each agent $k$  are defined as $p_1(\theta) = p_0 (\theta)\mathcal{N}(\theta|1, 4)$ and $p_2(\theta) = p_0 (\theta)(\mathcal{N}(\theta|-3, 1) + \mathcal{N}(\theta|3, 2))$ and the prior $p_0 (\theta)$ is uniform over $[-6, 6]$, i.e., $p_0 (\theta) = \mathcal{U}(\theta|-6,6)$. The local posteriors are shown in Fig. \ref{fig:mixture_toy} as dashed lines, along with the global posterior $q_{opt}(\theta) \propto \tilde{q}_{opt}(\theta)$ in (\ref{eq:exact_posterior}), which is represented as a shaded area. We fix the number of particles to $N=200$. The approximate posteriors obtained from the KDE over the global particles are plotted in Fig. \ref{fig:mixture_toy} as solid lines. It can be observed that at each round, the global posterior updated by DSVGD integrates the local likelihood of the scheduled agent, while still preserving information about the likelihood of the other agent from prior iterates, until (approximate) convergence to the true global posterior $q_{opt}$, which is a normalized version of $\tilde{q}_{opt}$ in (\ref{eq:exact_posterior}), is reached. \textcolor{black}{Finally, in Fig. \ref{fig:KL_global_iter_main}, we plot the KL divergence between $q(\theta)$ and $q_{opt}(\theta)$ as a function of the number of rounds. Both \ac{U-DSVGD} and \ac{DSVGD} exhibit similar behaviour, converging to \ac{SVGD} and outperforming the parametric counterparts \ac{PVI} and \ac{GVI} \citep{PVI}.}
\par
\textbf{Bayesian logistic regression. }We now consider Bayesian logistic regression for binary classification using the same setting as in \cite{NPV}. The model parameters $\theta = [\mathbf{w}, \log(\xi)]$ include the regression weights $\mathbf{w} \in \mathbbm{R}^{d}$ along with the logarithm of a precision parameter $\xi$. The prior is given as  $p_0 (\mathbf{w}, \xi) = p_0 (\mathbf{w}|\xi) p_0(\xi)$, with $p_0 (\mathbf{w}|\xi)=\mathcal{N}(\mathbf{w}| \mathbf{0}, \xi^{-1} \mathbf{I}_{d})$ and $p_0(\xi) = \mathrm{Gamma}(\xi| a, b)$ with $a=1$ and $b=0.01$. The local training loss $L_k (\theta)$ at each agent $k$ is given as $L_k(\theta) = \sum_{(\mathbf{x}_k, y_k) \in D_k} l(\mathbf{x}_k, y_k, \mathbf{w})$, where $D_k$ is the dataset at agent $k$ with covariates $\mathbf{x}_k \in \mathbbm{R}^{d}$ and label $y_k \in \{-1, 1 \}$, and the loss function $l(\mathbf{x}_k, y_k, \mathbf{w})$ is the cross-entropy.
Point decisions are taken based on the maximum of the average predictive distribution. We consider the datasets Covertype and Twonorm \citep{NPV}. 
We randomly split the training dataset into partitions of equal size among the $K$ agents. We also include \ac{FedAvg}, \ac{SGLD} and \ac{DSGLD} for comparison. We note that FedAvg is implemented here for consistency with the other schemes by scheduling a single agent at each step.
\begin{figure}[t]
    \centering
    \subfigure{\includegraphics[height=1.05 in]{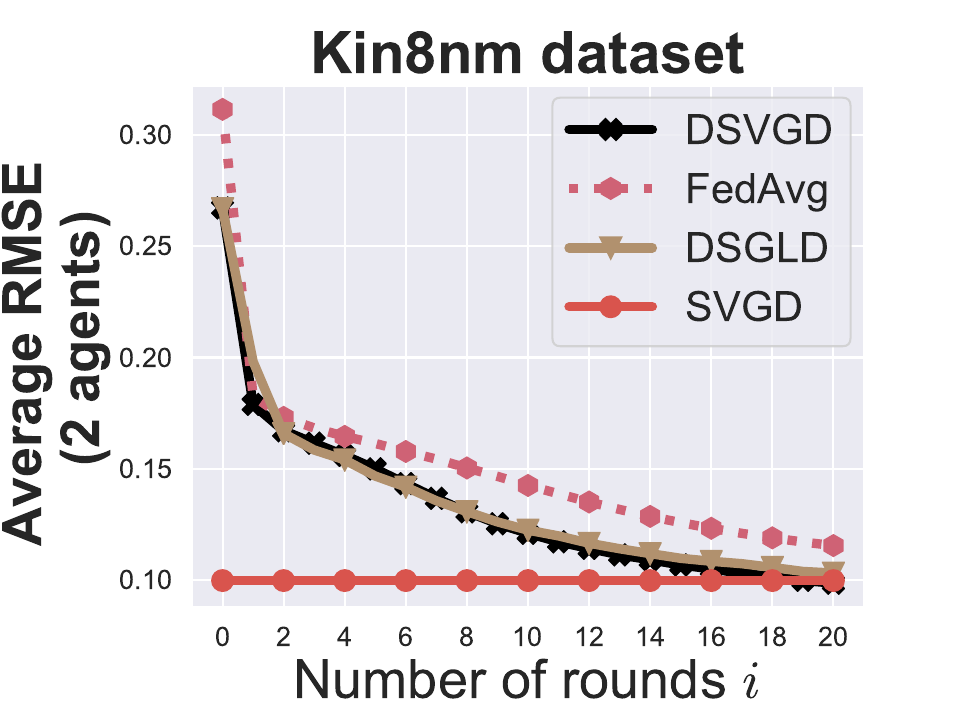}\hspace{-0.3cm}\includegraphics[height=1.05 in]{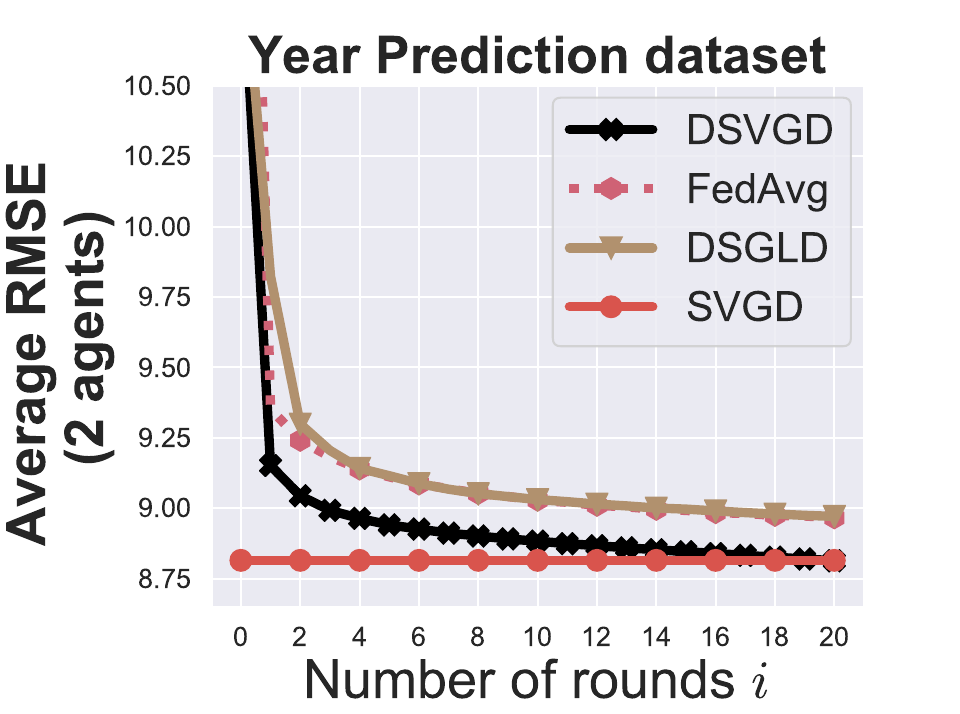}} \mbox{%
  \vline height 16ex}
    \subfigure{\includegraphics[height=1.05 in]{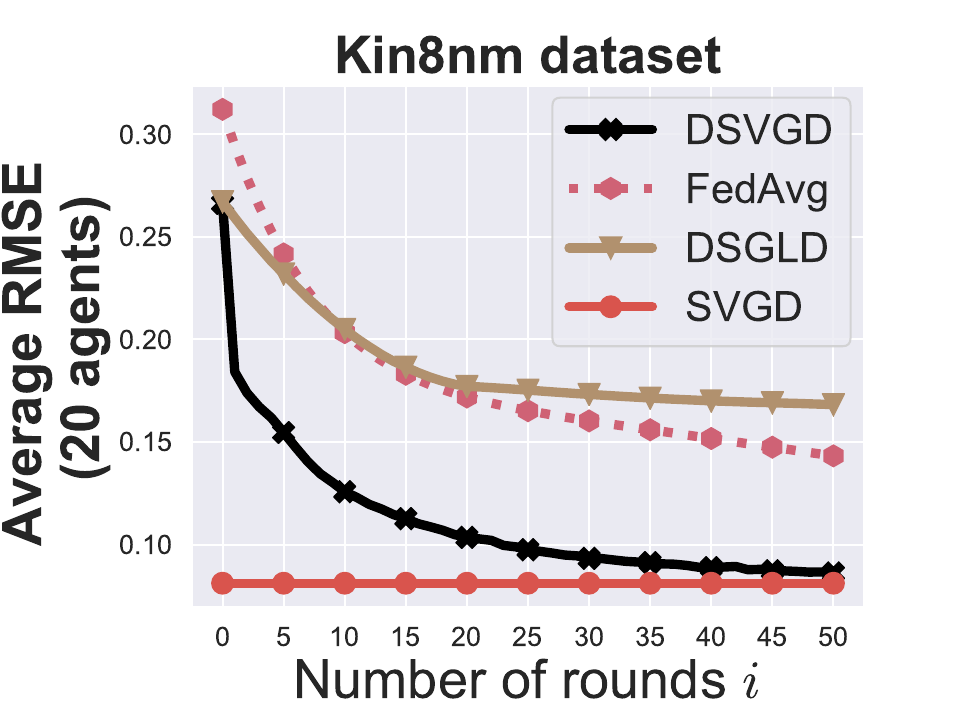}\hspace{-0.3cm}\includegraphics[height=1.05 in]{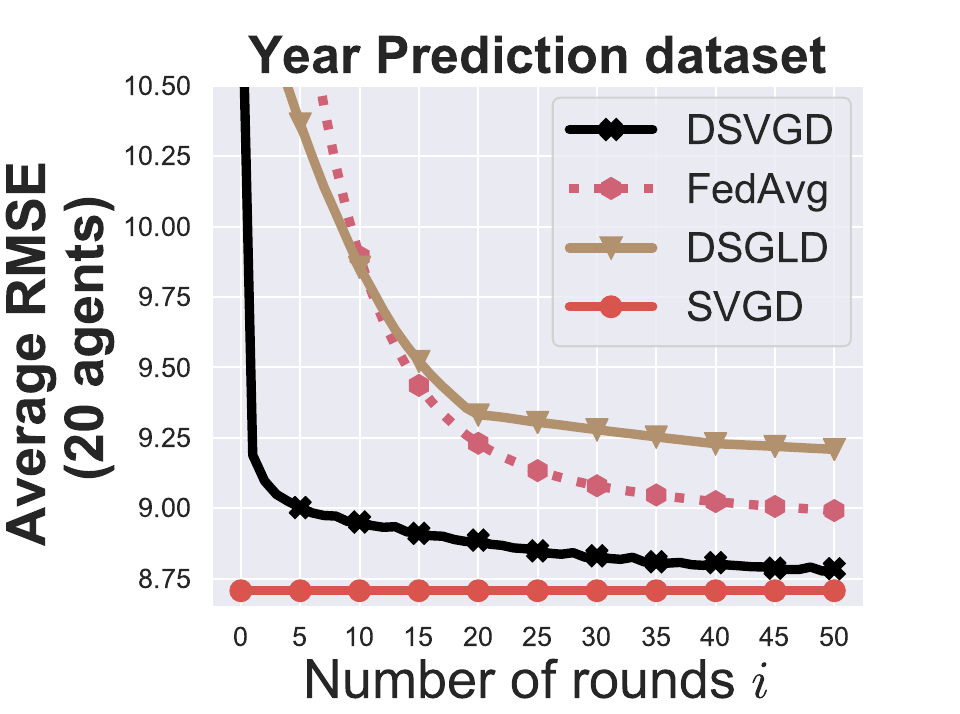}}

    \caption{Average RMSE as a function of the number of communication rounds $i$ for regression using \ac{BNN} with a single hidden layer of ReLUs with (left) $K=2$ agents and (right) $K=20$ agents ($N=20$, $L=L^\prime=200$, $100$ hidden neurons for the Year Prediction and $50$ for Kin8nm).}

    \label{fig:BNN_i_regression}
\end{figure}
\begin{figure}[t]
    \centering
    \vspace{-0.3\intextsep}
    \subfigure{{\includegraphics[height=1.05 in]{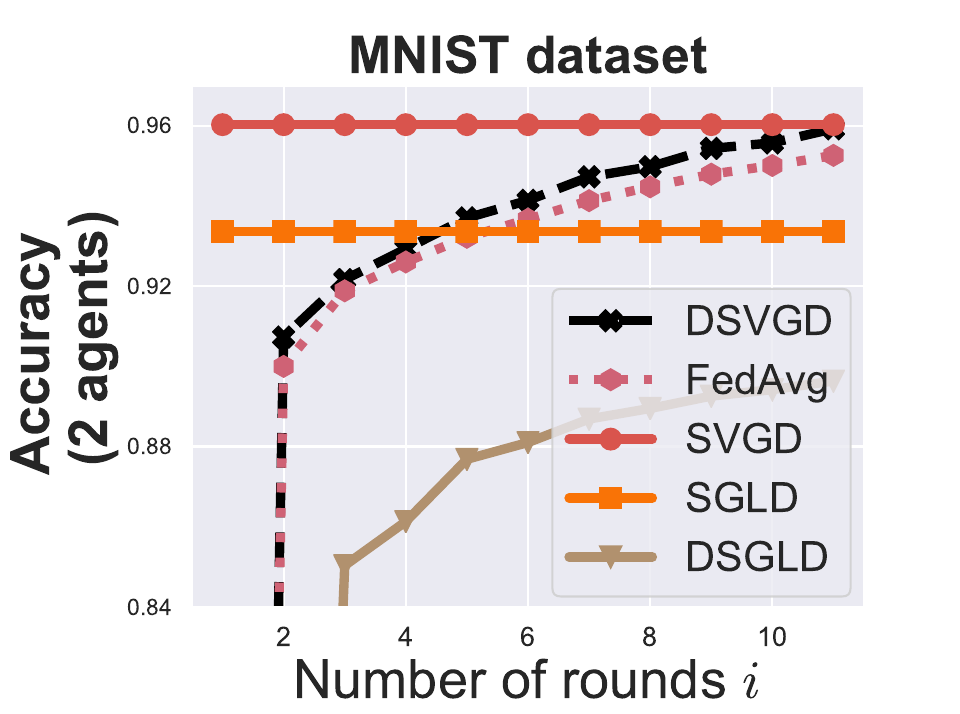}}\hspace{-0.35cm}\includegraphics[height=1.05 in]{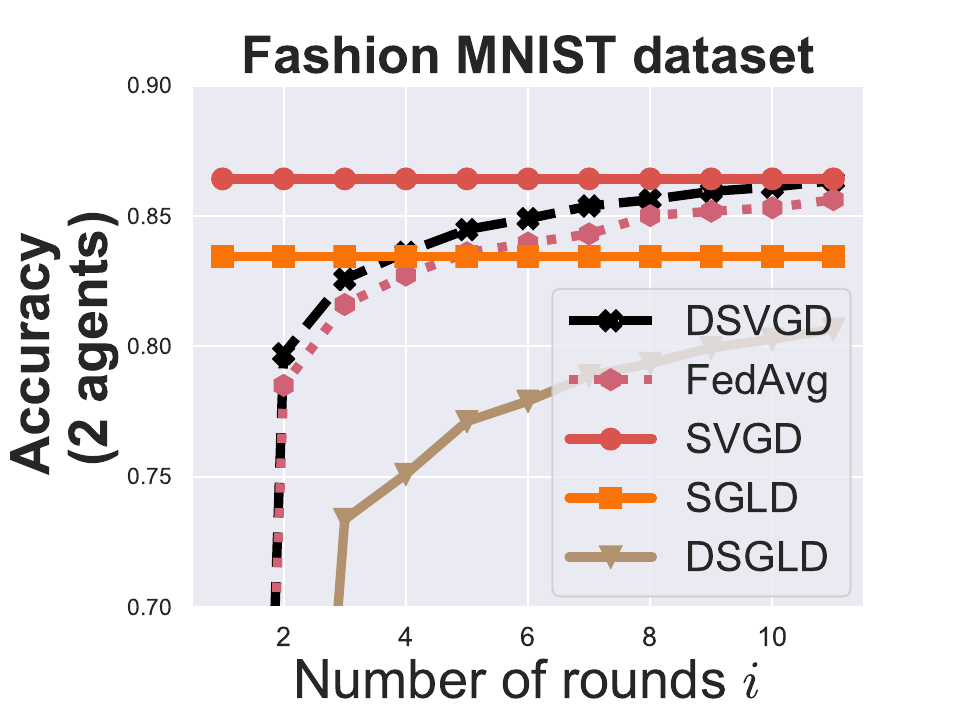}} 
    \mbox{
  \vline height 16ex}
    \subfigure{\includegraphics[height=1.05 in]{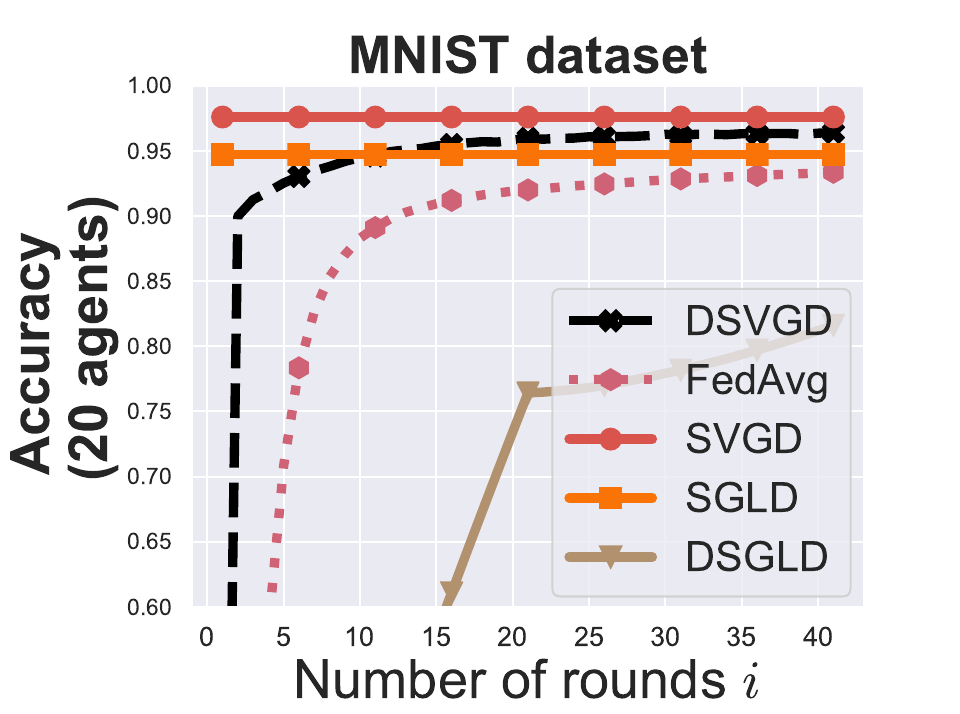}\hspace{-0.35cm}\includegraphics[height=1.05 in]{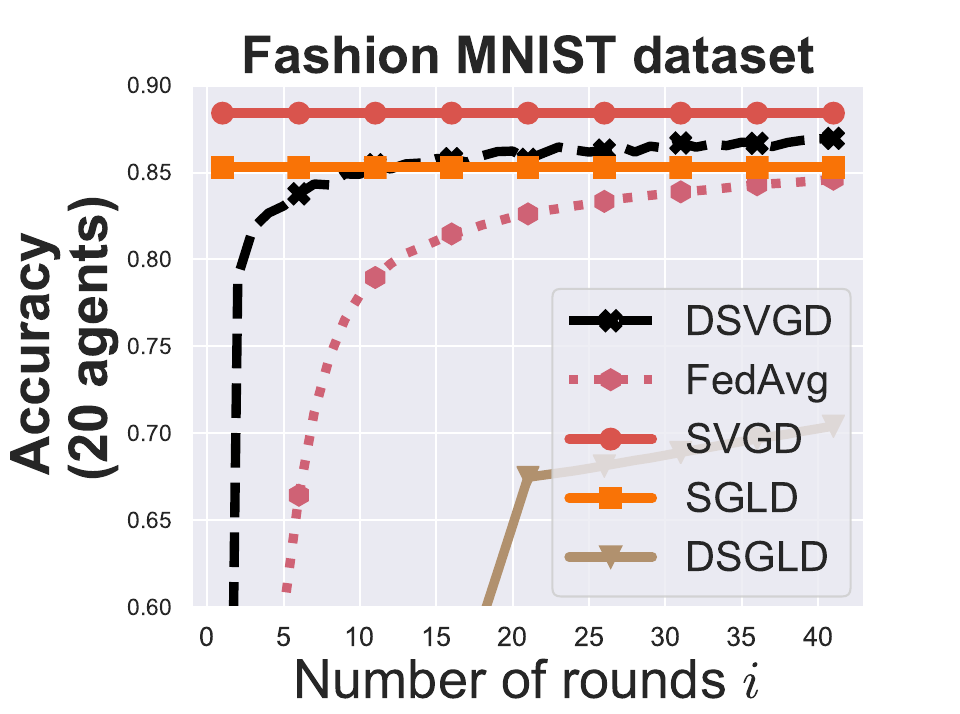}}

    \caption{Multi-label classification accuracy using \ac{BNN} with a single hidden layer of $100$ neurons as function of $i$, or number of communication rounds, using MNIST and Fashion MNIST with (left) $K=2$ agents and (right) $K=20$ agents ($N=20$, $L=L^\prime=200$).}

    \label{fig:BNN_i_classification}
\end{figure}
In Fig. \ref{fig:BLR_i_acc}, we study how the accuracy evolves as function of the number of communication rounds $i$, or number of communication rounds, across different datasets, using $N=2$ and $N=6$ particles. We observe that \ac{DSVGD} consistently outperforms the mentioned decentralized benchmarks and that, in contrast to \ac{FedAvg} and \ac{DSGLD}, its performance scales well with the number $K$ of agents. Furthermore, the number $N$ of particles is seen to control the trade-off between the communication load, which increases with $N$, and the convergence speed, which improves as $N$ grows larger. It is important to note that, in general, most benefits of the
proposed scheme appear to be obtained when the particles cover the main modes of the posterior. Since these are generally in limited number, the number of required particles is also seen to be small.
\textcolor{black}{Through reduction of the number of communication rounds, \ac{DSVGD} can also reduce the overall communication load. For example, in the third plot in Fig. \ref{fig:BLR_i_acc}, \ac{DSVGD} reaches an accuracy of $70\%$ after $5$ communication rounds with $N=6$, requiring the exchange of $30$ particles. In contrast, \ac{FedAvg} requires around $100$ rounds to obtain the same accuracy, making the total communication load much higher than that of DSVGD.} \par 
{\color{black}To capture heterogeneous datasets with non i.i.d. data, we now consider for different dataset partitions across $K=4$ agents. In the homogeneous case, labels are split equally among agents, while, in the heterogeneous case, each agent stores $40\%$ of one label and $10\%$ of the other. \ac{DSVGD} is seen in Fig. \ref{fig:llh_heterogeneous} to have a  robust performance against heterogeneity as compared to \ac{FedAvg}, whose convergence speed is severely affected. This result hinges on the fact that Bayesian learning provides a predictive distribution that is a more accurate estimate of the ground-truth posterior distribution. This is true irrespective of the level of ``non-iidness'': Bayesian learning can account in a principled away for all competing ``explanations'' provided by different devices. This is in contrast to FedAvg, whose reliance on a point estimate of the parameters yields an overconfident predictive distribution that cannot properly account for the diversity of predictions provided by different devices.}
\begin{wrapfigure}[13]{R}{0.55\textwidth}
    \centering
    \vspace{-1.8\intextsep}
    \subfigure{{\includegraphics[height=1.2 in, width=1.56 in]{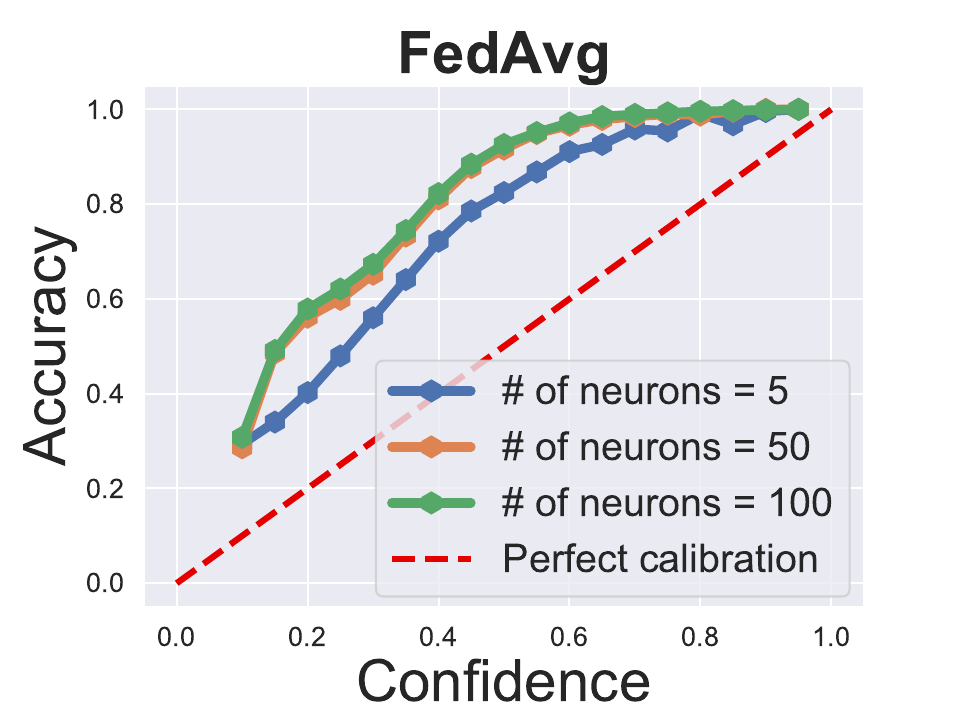}}}
    \hspace{-0.55cm}
    \subfigure{{\includegraphics[height=1.2 in, width=1.56 in]{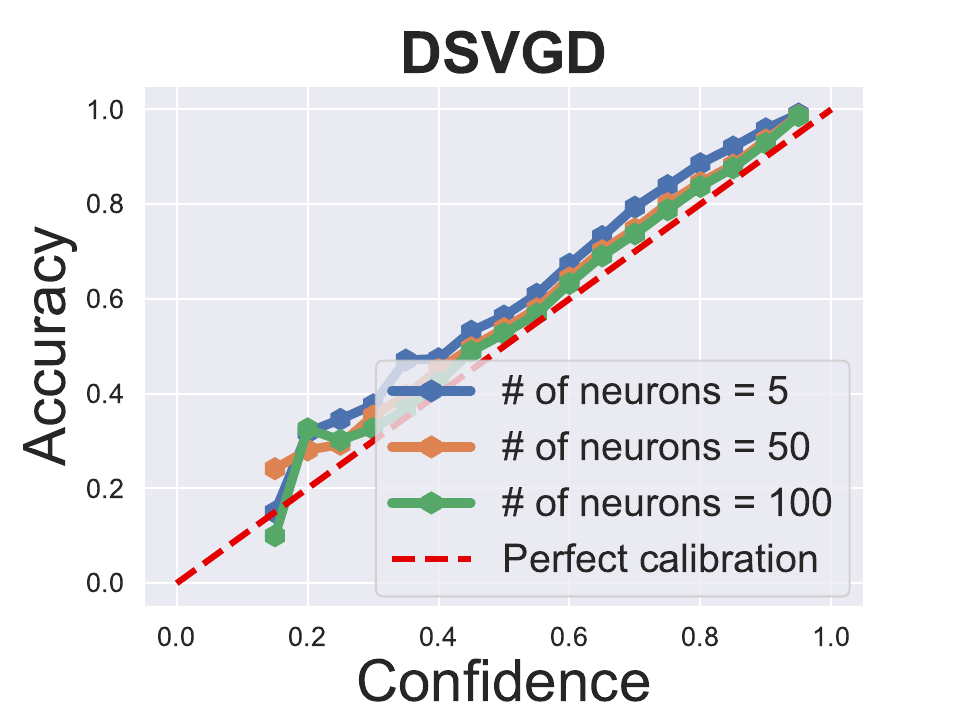}}}

    \caption{Reliability plots for classification using \ac{BNN} with variable number of hidden neurons using fashion MNIST ($N=20$, $I=10$, $L=L^{\prime}=200$, $K=20$).}
    \label{fig:rel_BNN}
\end{wrapfigure}
\par 
\textbf{Bayesian Neural Networks. }We now consider regression and multi-label classification with Bayesian Neural Networks (BNN) models. The experimental setup is the same as in \citet{PBP_lobato}, with the only exception that the prior of the weights is set to $p_0(\mathbf{w})=\mathcal{N}(\mathbf{w}|0, \lambda^{-1}\mathbf{I}_d)$ with a fixed precision $\lambda = e$. We plot the average \ac{RMSE} for $K=2$ and $K=20$ agents in Fig. \ref{fig:BNN_i_regression} for regression over the Kin8nm and Year datasets, and accuracy for multi-label classification on the MNIST and Fashion MNIST datasets in Fig. \ref{fig:BNN_i_classification}. Confirming the results for logistic regression, \ac{DSVGD} consistently outperforms the other decentralized benchmarks in terms of \ac{RMSE} and accuracy, while being more robust in terms of convergence speed to an increase in the number of agents. 
\par
\textbf{Calibration. }Reliability plots are a common visual tool used to quantify and visualize model calibration \citep{NN_calibration}. They report the average sample accuracy as function of the confidence level of the model. Perfect calibration yields an accuracy equal to the corresponding confidence (dashed line in Fig. \ref{fig:rel_BNN}). Fig. \ref{fig:rel_BNN} shows the reliability plots for \ac{FedAvg} and \ac{DSVGD} on the Fashion MNIST dataset for the \ac{BNN} setting. While increasing the number of hidden neurons negatively affects \ac{FedAvg} due to overfitting, \ac{DSVGD} enjoys excellent calibration even for large models and is hence able to make trustworthy predictions. 

\section{Conclusions}
\label{sec:conclusion}

This paper has introduced \ac{DSVGD}, a non-parametric distributed variational inference algorithm for generalized Bayesian federated learning. \ac{DSVGD} enables a flexible trade between per-iteration communication load and number of communication rounds, while being able to make trustworthy decisions via Bayesian inference.
\clearpage
\bibliography{Biblio.bib}
\bibliographystyle{iclr2021_conference}
\newpage
\appendix
\section{Complementary Materials}
\label{app:comlementary_materials}
\subsection{Algorithmic Tables}
\label{app:udsvgd}

\begin{algorithm}
\footnotesize
\LinesNumbered
\KwIn{prior $p_{0}(\theta)$, local loss function $\{L_{k}(\theta)\}_{k=1}^{K}$, temperature $\alpha>0$}
\KwOut{global posterior $q(\theta|\eta)$}
\vspace{0.1cm}
\hrule
\vspace{0.1cm}
{\bf initialize} $t_{k}^{(0)}(\theta) = 1$ for $k=1, \ldots, K$; $q^{(0)} (\theta) = p_0(\theta)$ \\
\For{{\em $i=1, \ldots, I$} }{ 
At scheduled agent $k$, download current global parameters $\eta^{(i-1)}$ from server\\
Agent $k$ solves local free energy problem in (\ref{eq:min_local_energy_cavity}) to obtain new global parameters $\eta^{(i)}$\\
Agent $k$ sends $\eta^{(i)}$ to the server and \textbf{server sets} $\eta \xleftarrow{} \eta^{(i)}$\\
Agent $k$ updates new approximate likelihood: $    t_k(\theta|\eta_{k}^{(i)}) = \frac{q(\theta|\eta^{(i)})}{q(\theta|\eta^{(i-1)})} t_k (\theta | \eta_{k}^{(i-1)})$
}
\Return $q(\theta) = q(\theta|\eta^{(I)})$
\caption{Partitioned Variational Inference (PVI) \citep{PVI}}
\label{algo_pvi}
\end{algorithm}

\begin{algorithm}
\footnotesize
\LinesNumbered
\KwIn{target distribution $\tilde{p}(\theta)$, initial particles $\{\theta_{n}^{(0)}\}_{n=1}^{N}\sim p_0 (\theta)$, kernel $\mathrm{k}(\cdot, \cdot)$, learning rate $\epsilon$}
\KwOut{particles $\{\theta_{n}\}_{n=1}^{N}$ that approximates the target normalized distribution}
\vspace{0.1cm}
\hrule
\vspace{0.1cm}
\For{{\em $i=1, \ldots, L$} }{ 
\For{{\em $n=1, \ldots, N$} }{ 
$\theta^{(i)}_{n} \xleftarrow[]{} \theta_{n}^{(i-1)} + \frac{\epsilon}{N} \sum_{j=1}^{N} [\mathrm{k}(\theta_{j}^{(i-1)}, \theta_{n}^{(i-1)}) \nabla_{\theta_j} \log \Tilde{p}(\theta_{j}^{(i-1)}) + \nabla_{\theta_j} \mathrm{k}(\theta_{j}^{(i-1)}, \theta_{n}^{(i-1)})].$
}
}
\Return $q(\theta) = \sum_{n=1}^{N} \mathrm{K}(\theta, \theta_{n}^{(L)})$
\caption{Stein Variational Gradient Descent (SVGD) \citep{SVGD}}
\label{algo_svgd}
\end{algorithm}

\begin{algorithm}
\footnotesize
\LinesNumbered
\KwIn{prior $p_{0}(\theta)$, local loss function $\{L_{k}(\theta)\}_{k=1}^{K}$, temperature $\alpha>0$, learning rate $\epsilon>0$, kernels $\mathrm{K}(\cdot, \cdot)$ and $\mathrm{k}(\cdot, \cdot)$ }
\KwOut{global posterior $q(\theta) = \sum_{n=1}^{N}\mathrm{K}(\theta, \theta_n)$}
\vspace{0.1cm}
\hrule
\vspace{0.1cm}
{\bf initialize} $t_{k}^{(0)}(\theta) = 1$ for $k=1, \ldots, K$; $q^{(0)} (\theta) = p_0(\theta)$; $\{\theta_{n}^{(0)} \}_{n=1}^{N} \overset{i.i.d}{\sim}
p_0 (\theta)$  \\
\For{{\em $i=1, \ldots, I$} }{\tcp{New communication round: server schedules an agent $k$} 
At scheduled agent $k$, download and memorize in local buffer current global particles $\{\theta_{n}^{(i-1)} \}_{n=1}^{N}$\\
Agent $k$ sets $\{ \theta_{n}^{[0]} = \theta_{n}^{(i-1)}\}_{n=1}^{N}$\\
\For{{\em $l=1,\ldots,L$}}{\tcp{Local iterations: agent $k$ minimizes local free energy}
Compute $\nabla_{\theta_n}^{[l]} = \nabla_{\theta} \log q^{(i-1)} (\theta_{n}^{[l-1]}) - \nabla_{\theta} \log t_{k}^{(i-1)} (\theta_{n}^{[l-1]}) - \frac{1}{\alpha} \nabla_{\theta} L_k (\theta_{n}^{[l-1]})$ with KDE $q^{(i-1)}(\theta) = \sum_{n=1}^{N} \mathrm{K}(\theta, \theta_{n}^{(i-1)})$ and $\nabla_{\theta} \log t_{k}^{(i-1)} (\theta)$ computed using (\ref{eq:grad_log_t_udsvgd})

\For{{\em particle $n=1, ..., N$ }}{
$\Delta \theta_n \xleftarrow[]{} \frac{1}{N} \sum_{j=1}^{N} \Big[\mathrm{k}(\theta_{j}^{[l-1]}, \theta_{n}^{[l-1]}) \nabla_{\theta_j}^{[l]} + \nabla_{\theta_j} \mathrm{k}(\theta_{j}^{[l-1]}, \theta_{n}^{[l-1]})\Big]$ \\
$\theta_{n}^{[l]} \xleftarrow[]{} \theta_{n}^{[l-1]} + \epsilon \Delta \theta_n$
}
}
Agent $k$ sets updated global particles $\{ \theta_{n}^{(i)} = \theta_{n}^{[L]} \}_{n=1}^{N}$ and memorize them in the local buffer\\
Agent $k$ sends particles $\{\theta_{n}^{(i)} \}_{n=1}^{N}$ to the server and \textbf{server sets} $\{ \theta_n = \theta_{n}^{(i)} \}_{n=1}^{N}$
}
\Return $q(\theta) = \sum_{n=1}^{N} \mathrm{K}(\theta, \theta_{n}^{(I)})$
\caption{Unconstrained-Distributed Stein Variational Gradient Descent (U-DSVGD)}
\label{algo2}
\end{algorithm}

\begin{algorithm}[t]
{\color{black}
\small
\LinesNumbered
\KwIn{prior $p_{0}(\theta)$, local loss functions $\{L_{k}(\theta)\}_{k=1}^{K}$, temperature $\alpha > 0$, kernels $\mathrm{K}(\cdot, \cdot)$ and $\mathrm{k}(\cdot, \cdot)$ }
\KwOut{global approximate posterior $q(\theta) = N^{-1} \sum_{n=1}^{N} \mathrm{K}(\theta, \phi_n)$}
\vspace{0.1cm}
\hrule
\vspace{0.1cm}
{\bf initialize} $q^{(0)} (\theta) = p_0(\theta)$; $\{\theta_{n}^{(0)} \}_{n=1}^{N} \overset{\text{i.i.d}}{\sim} p_0 (\theta)$; $ \{ \phi_{n}^{(0)}= \theta_{k, n}^{(0)} = \theta_{n}^{(0)} \}_{n=1}^{N}$ and $t_{k}^{(0)}(\theta) = 1$ for $k=1, \ldots, K$ \\
\For{{\em $i=1,\ldots, I$} }{
\textcolor{red}{Server} schedules a set $\mathcal{K}^{(i)}$ of agents in parallel \\
\textcolor{blue}{Agents} downloads current server particles $\{\phi_{n}^{(i-1)} \}_{n=1}^{N}$ from \textcolor{red}{server} \\

\textcolor{blue}{Agents} obtains updated global particles $\{ \theta_{n}^{(i)}\}_{n=1}^{N}$ using (\ref{eq:particles_updates_udsvgd}), $\{\theta_{n}^{(i-1)} = \phi_{n}^{(i-1)} \}_{n=1}^{N}$ and $\{ \theta_{k,n}^{(i-1)}\}_{n=1}^{N}$ \\

\textcolor{blue}{Agents} carries distillation to obtain $\{ \theta_{k,n}^{(i)}\}_{n=1}^{N}$ encoding $t_{k}^{(i)}(\theta)$ using (\ref{eq:nu_updates_dsvgd}) and $\{ \theta_{n}^{(i)}\}_{n=1}^{N}$\\

\textcolor{blue}{Agents} sends the obtained local particles $\{\theta_{k,n}^{(i)} \}_{n=1}^{N}$ for $k \in \mathcal{K}^{(i)}$ to the \textcolor{red}{server}\\

\textcolor{red}{Server} obtains $\{\phi_{n}^{(i)} \}_{n=1}^{N}$ using (\ref{eq:particles_updates_server}), $\{\phi_{n}^{(i-1)} \}_{n=1}^{N}$ and $\{\theta_{k,n}^{(i)} \}_{n=1}^{N}$ for $k \in \mathcal{K}^{(i)}$

}
\Return $q(\theta) = N^{-1}\sum_{n=1}^{N} \mathrm{K}(\theta, \phi_{n}^{(I)})$
}
\caption{\textcolor{black}{Parallel-Distributed Stein Variational Gradient Descent (P-DSVGD)}}
\label{algo:P_DSVGD}
\end{algorithm}

\clearpage
\subsection{A Relationship Between PVI and U-DSVGD}
\label{app:udsvgd_to_pvi}
We show here that \ac{PVI} with a Gaussian variational posterior $q(\theta|\eta) = \mathcal{N}(\theta|\lambda^2 \eta, \lambda^2 \mathrm{\mathbf{I}}_d)$ of fixed covariance $\lambda^2 \mathrm{\mathbf{I}}_d$ and mean $\lambda^2 \eta$ parametrized by natural parameter $\eta$ can be recovered as a special case of \ac{U-DSVGD}. To elaborate, consider \ac{U-DSVGD} with one particle $\theta_1$ (i.e., $N=1$), an \ac{RKHS} kernel that satisfies $\nabla_{\theta}\mathrm{k}(\theta, \theta)=0$ and $\mathrm{k}(\theta, \theta)=1$ (the \ac{RBF} kernel is an example of such kernel) and an isotropic Gaussian kernel $K(\theta, \theta_{1}^{(i)}) = \mathcal{N}(\theta|\theta_{1}^{(i)}, \lambda^2 \mathbf{I}_d)$ of bandwidth $\lambda$ used for computing the \ac{KDE} of the global posterior using the particles. The \ac{U-DSVGD} particles update in (\ref{eq:particles_updates_udsvgd}) reduces to the following single particle update:
\begin{equation}
        \theta^{[l]}_{1} \xleftarrow[]{} \theta_{1}^{[l-1]} + \epsilon  \nabla_{\theta} \log \Tilde{p}_{k}^{(i)}(\theta_{1}^{[l-1]}),\ \text{for}\ l=1,\ldots, L,
\end{equation}
with tilted distribution
\begin{equation}
    \Tilde{p}_{k}^{(i)} (\theta) \propto \frac{q^{(i-1)} (\theta)}{t_{k}^{(i-1)} (\theta)} \exp\bigg(-\frac{1}{\alpha} L_k(\theta)\bigg). \label{eq:tilted_udsvgd_app}
\end{equation}
The numerator in (\ref{eq:tilted_udsvgd_app}) can be rewritten as $q^{(i-1)} (\theta)= \mathrm{K}(\theta, \theta_{1}^{(i-1)})=q(\theta|\eta^{(i-1)})$ with $\eta^{(i-1)}=\lambda^{-2}\theta_{1}^{(i-1)}$, while the denominator can be rewritten as 
\begin{equation}
    t_{k}^{(i-1)}(\theta)=\prod_{j \in \mathcal{I}_{k}^{(i-1)}} \frac{q(\theta|\eta^{(j)})}{q(\theta|\eta^{(j-1)})}=t_{k}(\theta|\eta_{k}^{(i-1)}),
\end{equation}
with $\eta_{k}^{(i-1)} = \sum_{j \in \mathcal{I}_{k}^{(i-1)}} \eta^{(j)}-\eta^{(j-1)}$. This recovers the \ac{PVI} update (\ref{eq:tilted_dist}).
\subsection{Reliability Plots}
\label{app:reliability_plots}
In this part we give some background on reliability plots and \ac{MCE}. Reliability plots are a visual tool to evaluate model calibration \citep{rel_diagram_degroot, rel_diagrams_niculescu}. Consider a model that outputs a prediction $\hat{y}(x_i)$ and a probability $\hat{p}(x_i)$ of correct detection for an input $x_i$ with true label $y_i$. We divide the test samples into bins $\{\mathcal{B}_j \}_{j=1}^{B}$, each bin $\mathcal{B}_j$ containing all indices of samples whose prediction confidence falls into the interval $(\frac{j-1}{B}, \frac{j}{B}]$ where $B$ is the total number of bins. Reliability plots evaluate the accuracy as function of the confidence which are defined respectively as
\begin{equation}
\begin{aligned}
    \mathrm{acc}(\mathcal{B}_j) & = \frac{1}{|\mathcal{B}_j|} \sum_{i \in \mathcal{B}_j} \mathbf{1}_{\{ \hat{y}(x_i) = y_i \}} \\
    \mathrm{and}\ \mathrm{conf}(\mathcal{B}_j) & = \frac{1}{|\mathcal{B}_j|} \sum_{i \in \mathcal{B}_j} \hat{p}(x_i). \nonumber
    \end{aligned}
\end{equation}
Perfect calibration means that the accuracy is equal to the confidence across all bins. For example, given $100$ predictions, each with confidence approximately $0.7$, one should expect that around $70\%$ of these predictions be correctly classified.
\\ To compute $\hat{p}(x)$, we need the predictive probability $p(y_t|\mathbf{x}_t)$ for all samples $t \in [1;T]$. This can be obtained by marginalizing the data likelihood with respect to the weights vector $\mathbf{w}$. This marginalization is generally intractable but can be approximated for both Bayesian logistic regression and Bayesian Neural Networks as detailed in Sec. \ref{app:confidence_calibration_BLR} and Sec. \ref{app:confidence_calibration_BNN}. 
\par 
While reliability plots are a useful tool to visually represent the calibration of a model, it is often desirable to have a single scalar measure of miscalibration. In this paper, we use the \ac{MCE} that measures the worst case deviation of the model calibration from perfect calibration \citep{NN_calibration}. Mathematically, the \ac{MCE} is defined as 
\begin{equation}
    \mathrm{MCE} = \underset{j \in \{1,\ldots,B \}}{\max} |\mathrm{acc}(\mathcal{B}_j) - \mathrm{conf}(\mathcal{B}_j)|. \label{eq:MCE}
\end{equation}
Additional numerical results using both reliability plots and \ac{MCE} can be found in Sec. \ref{app:MCE}.
\subsubsection{Predictive Distribution for Bayesian Logistic Regression with SVGD and DSVGD}
\label{app:confidence_calibration_BLR}
In this section, we show how the predictive distribution for the Bayesian logistic regression experiment can be obtained when using \ac{DSVGD} or \ac{SVGD}. The predictive distribution provides the confidence values to be used in the calibration experiment. Given a \ac{KDE} of the posterior $q(\mathbf{w})=\sum_{n=1}^{N}\mathrm{k}(\mathbf{w}, \mathbf{w}_n)$ with $N$ particles $\{\mathbf{w}_n \}_{n=1}^{N}$ the predictive probability for Bayesian logistic regression can be estimated as 
\begin{equation}
    p(y_t = 1| \mathbf{x}_t) \approx \int p(y_t = 1| \mathbf{x}_t, \mathbf{w}) q(\mathbf{w}) \mathrm{d}\mathbf{w} = \sum_{n=1}^{N} \frac{1}{N (2\lambda^2\pi)^{d/2}} \int \frac{\exp (\frac{-1}{2\lambda^2} || \mathbf{w}_n - \mathbf{w} ||^2)}{1 + \exp(-\mathbf{w}\mathbf{x}_{t}^{T})} \mathrm{d}\mathbf{w}.\label{eq:BLR_predictive_dist}
\end{equation}
A good approximation of (\ref{eq:BLR_predictive_dist}) can be obtained by replacing the logistic sigmoid function with the probit function \citep[Sec. 4.5]{bishop_book_pattern_recognition}, yielding
\begin{equation}
     p(y_t = 1| \mathbf{x}_t) \approx \sum_{n=1}^{N} \frac{1}{N} \frac{1}{1 + \exp(-\kappa(\sigma^2)\mu_n)},
\end{equation}
where 
\begin{equation}
\begin{aligned}
    \mu_n &= \mathbf{w}_n \mathbf{x}_{t}^{T} ,\\ \sigma^2 &= \frac{1}{\lambda^2} \mathbf{x}_t \mathbf{x}_{t}^{T} ,\\
    \mathrm{and}\ \kappa(\sigma^2) &= \bigg(1 + \sigma^2 \frac{\pi}{8}\bigg)^{-1/2}.
\end{aligned}
\end{equation}
\subsubsection{Predictive Distribution for Bayesian Neural Networks with SVGD and DSVGD}
\label{app:confidence_calibration_BNN}
In a manner similar to (\ref{eq:BLR_predictive_dist}), the predictive distribution for \ac{BNN} can be estimated as
\begin{equation}
    p(y_t=1|\mathbf{x}_t) \approx \sum_{n=1}^{N} \frac{1}{N(2\lambda^2\pi)^{d/2}} \int f(\mathbf{x}_t, \mathbf{w}) \exp\bigg(\frac{-|| \mathbf{w}_n - \mathbf{w} ||^2}{2\lambda^2} \bigg) \mathrm{d}\mathbf{w}, \label{eq:predictive_BNN}
\end{equation}
where $f(\mathbf{x}_t, \mathbf{w})$ is the sigmoid output of the \ac{BNN} with weights $\mathbf{w}$. Using the first order Taylor approximation of the network output around the $n$-th particle \citep[Sec. 5.7.1]{bishop_book_pattern_recognition}
\begin{equation}
    f(\mathbf{x}_t, \mathbf{w}) \approx f(\mathbf{x}_t, \mathbf{w}_n) + \nabla^{\mathsf{T}}_{\mathbf{w}} f(\mathbf{x}_t, \mathbf{w}) (\mathbf{w}-\mathbf{w}_n),
\end{equation}
the predictive distribution can now be rewritten as 
\begin{equation}
\begin{aligned}
        p(y_t=1|\mathbf{x}_t) &\approx \sum_{n=1}^{N} \frac{1}{N(2\lambda^2\pi)^{d/2}} \int [f(\mathbf{x}_t, \mathbf{w}_n) + \nabla^{\mathsf{T}}_{\mathbf{w}} f(\mathbf{x}_t, \mathbf{w}) (\mathbf{w}-\mathbf{w}_n)] \exp\bigg(\frac{-|| \mathbf{w}_n - \mathbf{w} ||^2}{2\lambda^2}\bigg) \mathrm{d}\mathbf{w}\\
        &= \sum_{n=1}^{N} \frac{1}{N} f(\mathbf{x}_t, \mathbf{w}_n) + \sum_{n=1}^{N} \frac{1}{N}\big(\nabla^{\mathsf{T}}_{\mathbf{w}} f(\mathbf{x}_t, \mathbf{w})\mathbf{w}_n - \nabla^{\mathsf{T}}_{\mathbf{w}} f(\mathbf{x}_t, \mathbf{w})\mathbf{w}_n \big) \\
        &= \sum_{n=1}^{N} \frac{1}{N} f(\mathbf{x}_t, \mathbf{w}_n),
\end{aligned}
\end{equation}
where we have used the fact that $\int \mathcal{N}(\mathbf{w}|\mathbf{w}_n, \lambda^2\mathbf{I}_d) \mathrm{d}\mathbf{w}=1$ and $\int \mathbf{w} \mathcal{N}(\mathbf{w}|\mathbf{w}_n, \lambda^2\mathbf{I}_d) \mathrm{d}\mathbf{w}=\mathbf{w}_n$.
\subsection{Space-Time Complexity, Communication Load and Convergence}
\label{app:complexity_convergence}
This section offers a brief discussion on the complexity, communication load and convergence of \ac{DSVGD}. \par
\textbf{Space Complexity. }\ac{DSVGD} inherits the space complexity of \ac{SVGD}. In particular, \ac{DSVGD} requires the computation of the kernel matrix $\mathrm{k}(\cdot, \cdot)$ between all particles at each local iteration, which can then be deleted before the next iteration. This requires $\mathcal{O}(N^2)$ space complexity. As pointed out by \citet{SVGD} and noticed in our experiments, for sufficiently small problems of practical interest for mobile embedded applications, few particles are enough to obtain state-of-the art performance. Furthermore, $N$ particles of dimension $d$ need to be saved in the local buffer, requiring $\mathcal{O}(Nd)$ space. Given that $N$ is generally much lower than the number of data samples, saving the particles in the local buffer shouldn't be problematic.
\par
\textbf{Time complexity.} When scheduled, an agent has to perform $\mathcal{O}(\max(L, L^{\prime})N^2)$ operations with $\mathcal{O}(LN^2)$ operations for the first loop (lines $\mathbf{5}$-$\mathbf{11}$) and $\mathcal{O}(L^\prime N^2)$ operations for the second loop (lines $\mathbf{15}$-$\mathbf{21}$) in Algorithm \ref{algo1}. Furthermore, the $L'$ distillation iterations in the second loop can be performed by the scheduled agent after it has sent its global particles to the central server. This enables the pipelining of the second loop with the operations at the server and at other agents, which can potentially reduce the wall-clock time per communication round.
\par
\textbf{Communication load. }Using DSVGD, the communication load between a scheduled agent and the central server is of the order $\mathcal{O}(Nd)$ since $N$ particles of dimensions $d$ need to be exchanged at each communication round. In contrast, the communication load of \ac{PVI} depends on the selected parametrization. For instance, one can use \ac{PVI} with a fully factorized Gaussian approximate posterior, which requires only $2d$ parameters to be shared with the server, namely mean and variance of each of the $d$ parameters at the price of having lower accuracy.
\par
\textbf{Convergence. }The two local \ac{SVGD} loops produce a set of global and local particles, respectively, that are convergent to their respective targets as the number $N$ of particles increases \citep{SVGD_proof}. Furthermore, as discussed, a fixed point of the set of local free energy minimization problems is guaranteed to be a local optimum for the global free energy problem (see Property $3$ in \citet{PVI}). This property hence carries over to \ac{DSVGD} in the limit of large number of particles. However, convergence to a fixed point is an open question for \ac{PVI}, and consequently also for \ac{DSVGD}.
{\color{black}
\subsection{Parallel-DSVGD }
\label{sec:P_DSVGD}
In this section, we present a direct extension of DSVGD in which multiple agents can be scheduled in parallel during the same communication round. In Parallel-DSVGD (P-DSVGD), each agent in the set $\mathcal{K}^{(i)}$ of scheduled agents at round $i$ applies the same steps as in DSVGD except that it shares the local particles $\{\theta_{k,n}^{(i)}\}_{n=1}^{N}$ with the server instead of the global ones. Then, the server distills the received local particles into a set of $N$ server-side particles $\{\phi_{n}^{(i)} \}_{n=1}^{N}$ using SVGD to obtain the next iterate of the global posterior.
\par
As discussed in Sec. \ref{sec:distributed_VI}, a parallel implementation requires the $i$-th iterate of the global posterior to be obtained as
\begin{align}
    q^{(i)}(\theta) = p_0(\theta) \prod_{k \in \mathcal{K}^{(i)}} t_{k}^{(i)}(\theta) \prod_{k^\prime \not\in \mathcal{K}^{(i)}} t_{k^\prime}^{(i)}(\theta), \label{eq:approx_global_parallel}
\end{align}
where $t_{k^\prime}^{(i)}(\theta) = t_{k^\prime}^{(i-1)}(\theta)$ for $k^\prime \not\in \mathcal{K}^{(i)}$. To replicate this same behaviour while preserving the non-parametric property of \ac{DSVGD}, in P-DSVGD, each agent $k \in \mathcal{K}^{(i)}$ shares its local particles $\{\theta_{k,n}^{(i)} \}_{n=1}^{N}$  representing the approximate likelihood where $t_{k}^{(i)}(\theta) = N^{-1}\sum_{n}^{N}\mathrm{K}(\theta, \theta_{k,n}^{(i)})$. Then, to approximate $q^{(i)}(\theta)$ in (\ref{eq:approx_global_parallel}), using SVGD, the server carries out $L_s$ \ac{SVGD} updates as
\begin{equation}
        \phi^{[l]}_{n} \xleftarrow[]{} \phi_{n}^{[l-1]} + \frac{\epsilon}{N} \sum_{j=1}^{N} [\mathrm{k}(\phi_{j}^{[l-1]}, \phi_{n}^{[l-1]}) \nabla_{\theta_j} \log  q^{(i)}(\phi_{j}^{[l-1]}) \!\!+ \!\!\nabla_{\phi_j} \mathrm{k}(\phi_{j}^{[l-1]}, \phi_{n}^{[l-1]})],\! \text{for}\ l\!\!=\!\!1,\ldots, \! L_s.
        \label{eq:particles_updates_server}
\end{equation}
For the $(i+1)$-th communication round, scheduled agents $\mathcal{K}^{(i+1)}$ download particles $\{\phi_{n}^{(i+1)} \}_{n=1}^{N} = \{\phi_{n}^{[L_s]} \}_{n=1}^{N}$ that are treated in a similar fashion as in \ac{DSVGD}. The full algorithmic table for \ac{P-DSVGD} is provided in Algorithm \ref{algo:P_DSVGD}. Numerical results for P-DSVGD are provided in Sec. \ref{app:distributed_BLR} of the Appendix.}\par 

\include{my_theorems}
\clearpage
\section{Additional Experiments}
\label{app:additional_experiments}
An overview of the benchmarks considered in the experiments is provided in Table \ref{table:benchmarks}.
\begin{table}[h]
\caption{Overview of benchmarks used in the experiments.}
\centering
  \scalebox{0.7}{\begin{tabular}{c  c  c c}
    \\
    \toprule
    \multirow{1}{*}{Algorithm}
        & {Non-parametric} & {Decentralized} & {Inference} \\
      \hline
     Stein Variational Gradient Descent (SVGD) \citep{SVGD} & \textbf{Yes} & No & VI\\
    Stochastic Gradient Langevin Dynamics (SGLD) \citep{sgld} & \textbf{Yes} & No & MC \\
    Distributed Stochastic Gradient Langevin Dynamics (DSGLD) \citep{dsgld} & \textbf{Yes} & \textbf{Yes} & MC\\
    Particle Mirror Descent (PMD) \citep{PMD_bo_dai} & \textbf{Yes} & No & VI\\
    Partitioned Variational Inference (PVI) \citep{PVI} & No & \textbf{Yes} & VI\\
    Global Variational Inference (GVI) \citep{GVI_sato} & No & No & VI\\
    Non-Parametric Variational Infernce (NPV) \citep{NPV} & No & No & VI\\
    Federated Averaging (FedAvg) \citep{comm_efficient_learning_dec_data} & No & \textbf{Yes} & Freq.\\
    Federated Stochastic Gradient Descent (FedSGD) \citep{comm_efficient_learning_dec_data} & No & \textbf{Yes} & Freq.\\
    Federated Bayesian Model Ensemble (FedBe) \citep{feddistill} & No & \textbf{Yes} & Freq.\\
    \textbf{Distributed Stein Variational Gradient Descent} (\textbf{DSVGD}) (ours) & \textbf{Yes} & \textbf{Yes} & VI\\
    
    \bottomrule
  \end{tabular}}
  \label{table:benchmarks}
\end{table}
\subsection{1-D Mixture of Gaussians Toy Example}
\begin{figure}[h]
    \centering
    \subfigure[]{\includegraphics[height=1.6in]{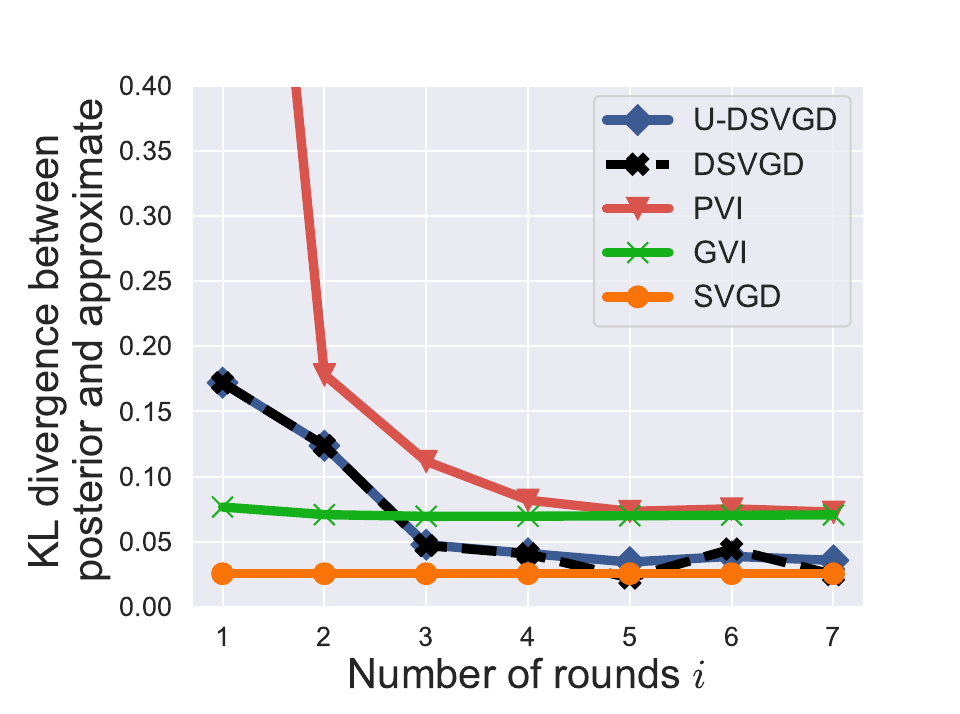} \label{fig:all_global_iter_number}}
    \subfigure[]{\includegraphics[height=1.6in]{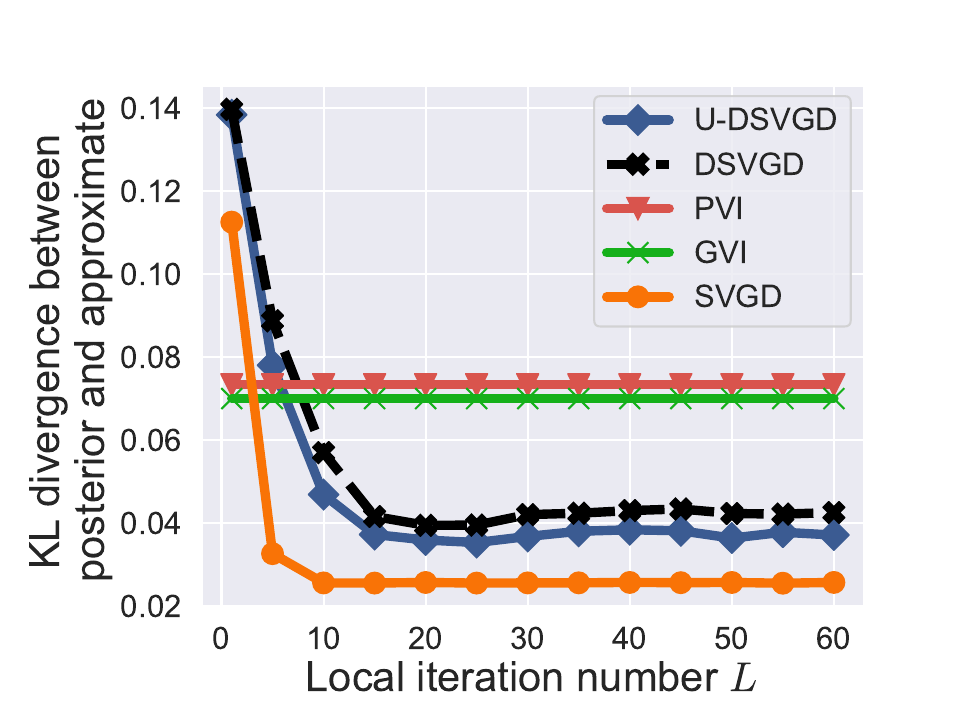}\label{fig:far_lh_global_local}}
    \caption{KL divergence between exact and approximate global posteriors (a) as function of the number of communication rounds $i$ for $L=L^\prime=200$; and (b) as function of the local iterations number $L$ for $I = 5$.}
\end{figure}
This section is complementary to the $1$-D mixture of Gaussians experiment in Sec. \ref{sec:experiments} of the main text. We compare \ac{DSVGD} with \ac{PVI} and the counterpart centralized schemes. In Fig. \ref{fig:all_global_iter_number}, we plot the KL divergence between the global posterior $q_{opt}(\theta)$ and its current approximation $q(\theta)$ as a function of the number of communication rounds $i$, which corresponds to the number of communication rounds for decentralized schemes. We use $N=200$ particles for \ac{U-DSVGD} and \ac{DSVGD} with $L=L^\prime=200$ local iterations. The number of \ac{SVGD} iterations is fixed to $800$. A Gaussian prior $p_0(\theta) = \mathcal{N}(\theta|0,1)$ is assumed in lieu of the uniform prior considered in Fig. \ref{fig:mixture_toy} to facilitate the implementation of \ac{PVI} and conventional centralized \ac{GVI} which was done following \citet[Property 4]{PVI}. More specifically, we use Gaussian approximate likelihoods, i.e., $t_k(\theta|\eta)=\mathcal{N}(\theta| \frac{-\eta_1}{2\eta_2}, \frac{-1}{2\eta_2})$ with natural parameters $\eta_1$ and $\eta_2 < 0$. We observe that \ac{DSVGD} has similar convergence speed as PVI, while having a superior performance thanks to the reduced bias of non-parametric models. Furthermore, \ac{DSVGD} exhibits the same performance as \ac{U-DSVGD} with the advantage of having memory requirements that do not scale with the number of iterations. Finally, both \ac{U-DSVGD} and \ac{DSVGD} converge to the performance of (centralized) \ac{SVGD} as the number of rounds increases.
\par
In Fig. \ref{fig:far_lh_global_local}, we plot the same KL divergence as function of the number of local iterations $L$. We use $I=5$ rounds for the decentralized schemes. It is observed that non-parametric schemes-namely \ac{SVGD} and (U-)DSVGD-require a sufficiently large number of local iterations in order to outperform the parametric strategies \ac{PVI} and \ac{GVI}.
\subsection{2-D Mixture of Gaussians Toy Example}
\label{app:2d_mixture}
\begin{figure}[h]
\centering
\subfigure[][]{%
\label{fig:contour_mixture-a}%
\includegraphics[height=1.4in]{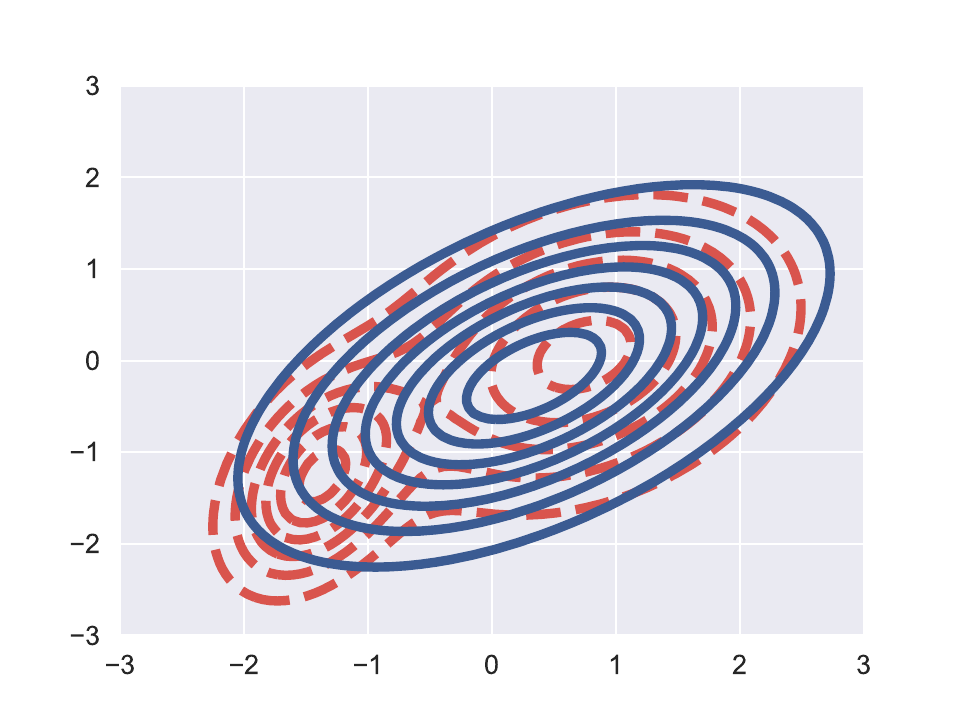} }%
\subfigure[][]{%
\label{fig:contour_mixture-b}%
\includegraphics[height=1.4in]{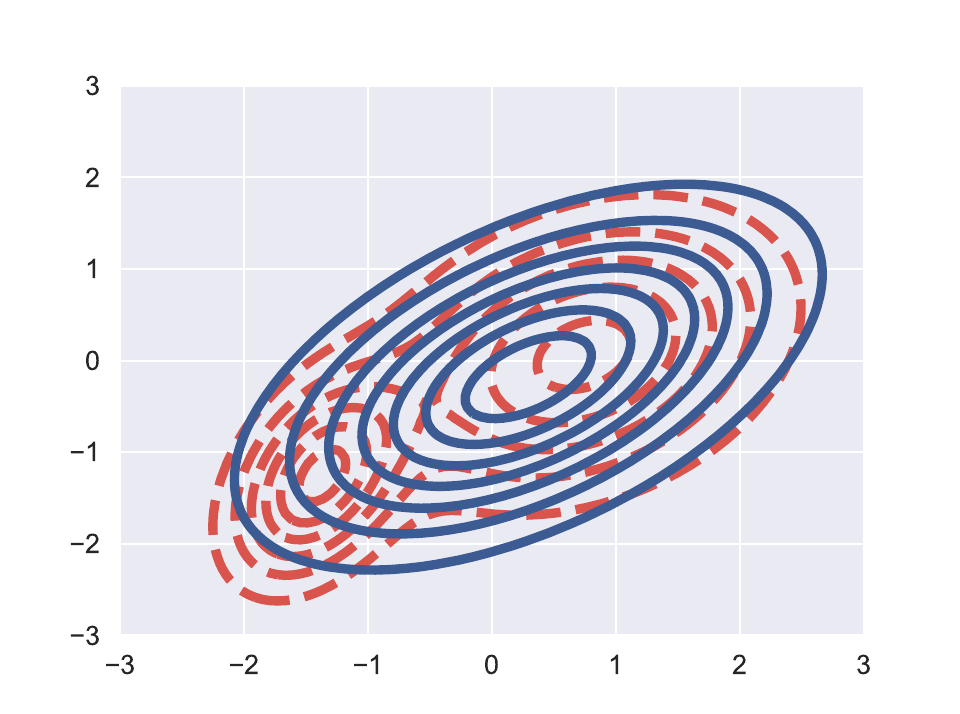}} \\
\vspace{-0.3cm}
\subfigure[][]{%
\label{fig:contour_mixture-c}%
\includegraphics[height=1.4in]{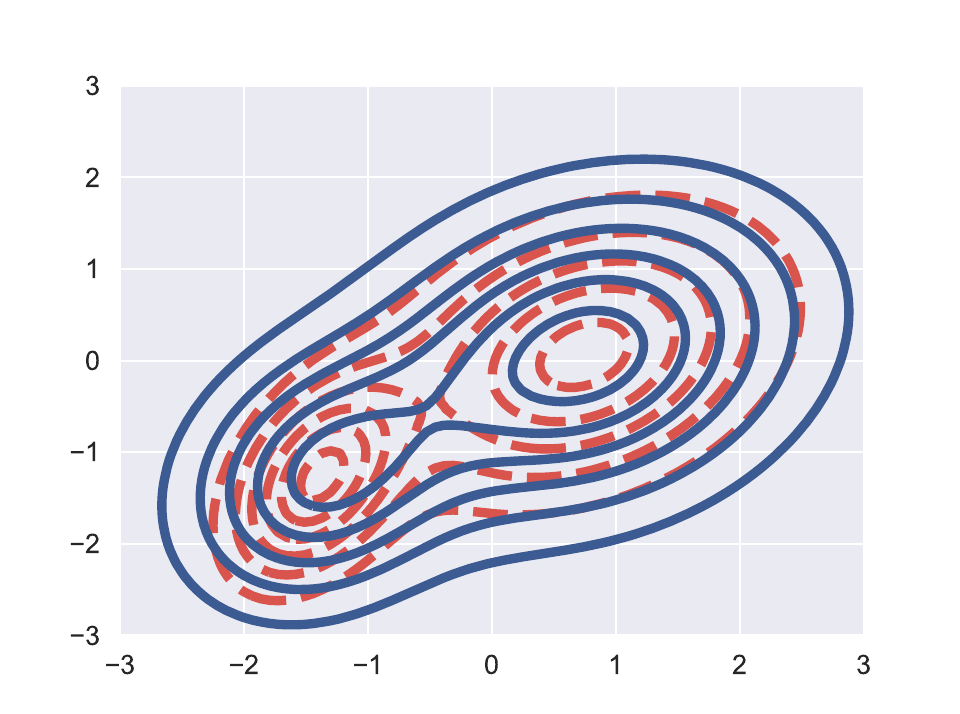}}%
\subfigure[][]{%
\label{fig:contour_mixture-d}%
\includegraphics[height=1.42in]{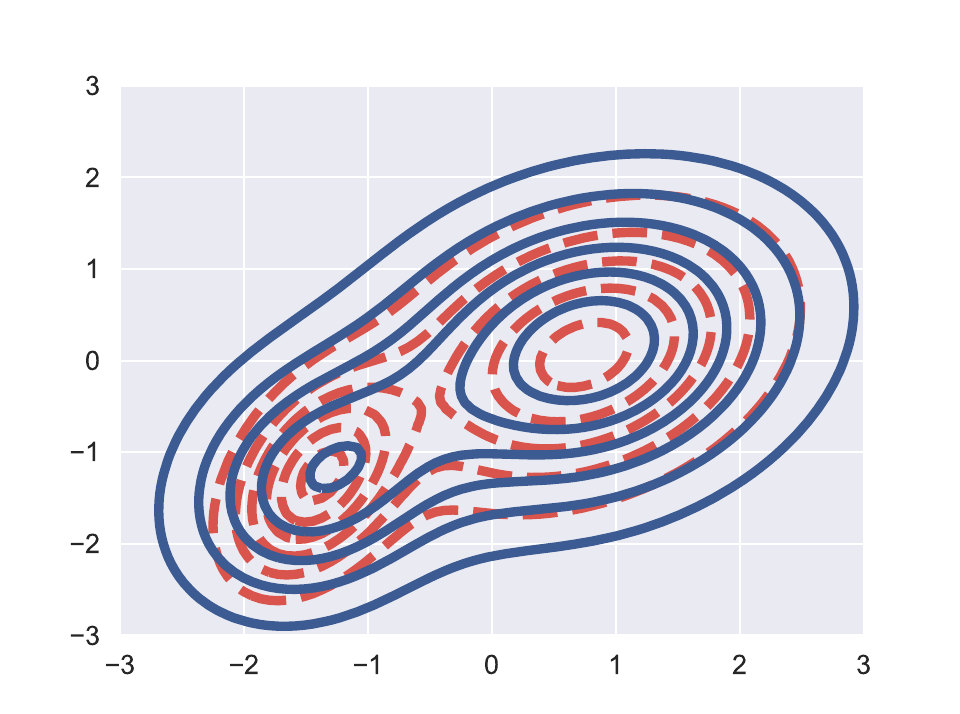}}%
\caption[Performance comparison of ]{Performance comparison of 
\subref{fig:contour_mixture-a} GVI,
\subref{fig:contour_mixture-b} PVI,
\subref{fig:contour_mixture-c} SVGD and 
\subref{fig:contour_mixture-d} DSVGD for a multivariate Gaussian mixture model. Solid contour lines correspond to the approximate posterior while dashed contour lines to the exact posterior ($N=200$, $I=5$, $L=200$ and $b=0.1$).}%
\label{fig:contour_mixture_approx_exact_posteruior}%
\end{figure}
We now consider the following 2-D mixture of Gaussians model: $p_1(\theta) = \mathcal{N}(\boldsymbol{{\mu}}_0, \mathbf{\Sigma}_0) (\mathcal{N}(\boldsymbol{{\mu}}_1, \mathbf{\Sigma}_1) + \mathcal{N}(\boldsymbol{{\mu}}_2, \mathbf{\Sigma}_2)) $ and $p_2(\theta) = \mathcal{N}(\boldsymbol{{\mu}}_0, \mathbf{\Sigma}_0) \mathcal{N}(\boldsymbol{{\mu}}_3, \mathbf{\Sigma}_3)$ where 
\begin{equation}
\begin{aligned}
& \boldsymbol{\mu}_0 = [0, 0]\ ;\ 
\mathbf{\Sigma}_0 = \begin{bmatrix}
4 & 2\\
2 & 4
\end{bmatrix} \\
& \boldsymbol{\mu}_1 = [-1.71, -1.801]\ ;\ 
\mathbf{\Sigma}_1 = \begin{bmatrix}
0.226 & 0.1652\\
0.1652 & 0.6779
\end{bmatrix} \\
& \boldsymbol{\mu}_2 = [1, 0]\ ;\ 
\mathbf{\Sigma}_2 = \begin{bmatrix}
2 & 0.5\\
0.5 & 2
\end{bmatrix}\\
& \boldsymbol{\mu}_3 = [1, 0]\ ;\ 
\mathbf{\Sigma}_3 = \begin{bmatrix}
3 & 0.5\\
0.5 & 3
\end{bmatrix}.
\end{aligned} \nonumber
\end{equation}
We plot in Fig. \ref{fig:contour_mixture_approx_exact_posteruior} the approximate posterior $q(\theta)$ (black solid contour lines) and the exact posterior $q_{opt}(\theta)$ (red dashed contour lines) for \ac{PVI}, \ac{GVI}, \ac{SVGD} and \ac{DSVGD}.
We see that, as in the 1-D case and in contrast to parametric methods PVI and GVI, non-parametric methods SVGD and DSVGD are able to capture the different modes of the posterior, obtaining lower values for the KL divergence between the approximate and exact posterior.
\subsection{Bayesian Logistic Regression}
\label{app:distributed_BLR}
\begin{figure}[h]
    \centering
    \subfigure{\includegraphics[height=1.5 in]{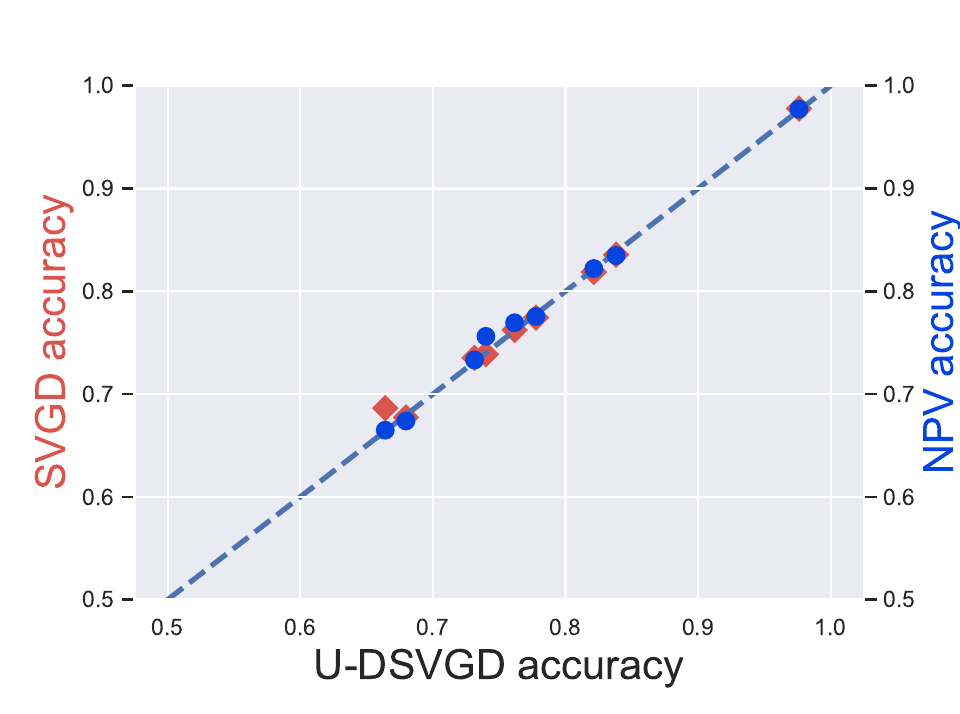}} 
    \subfigure{\includegraphics[height=1.5 in]{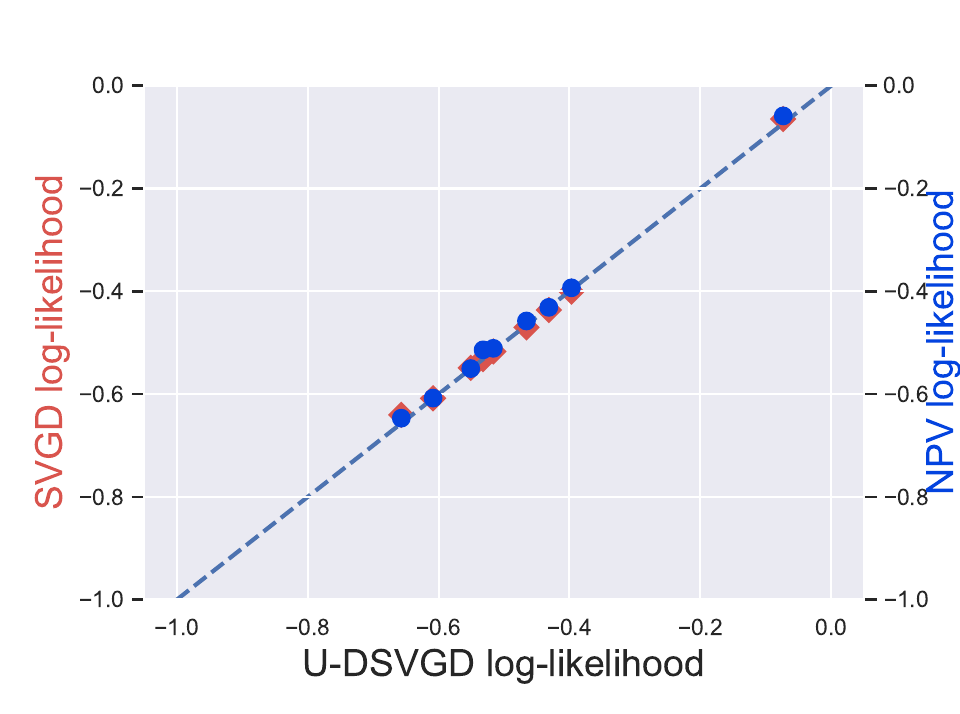}}\\
    \BlankLine
    \vspace{-0.7cm}
    \BlankLine
    \subfigure{\includegraphics[height=1.5 in]{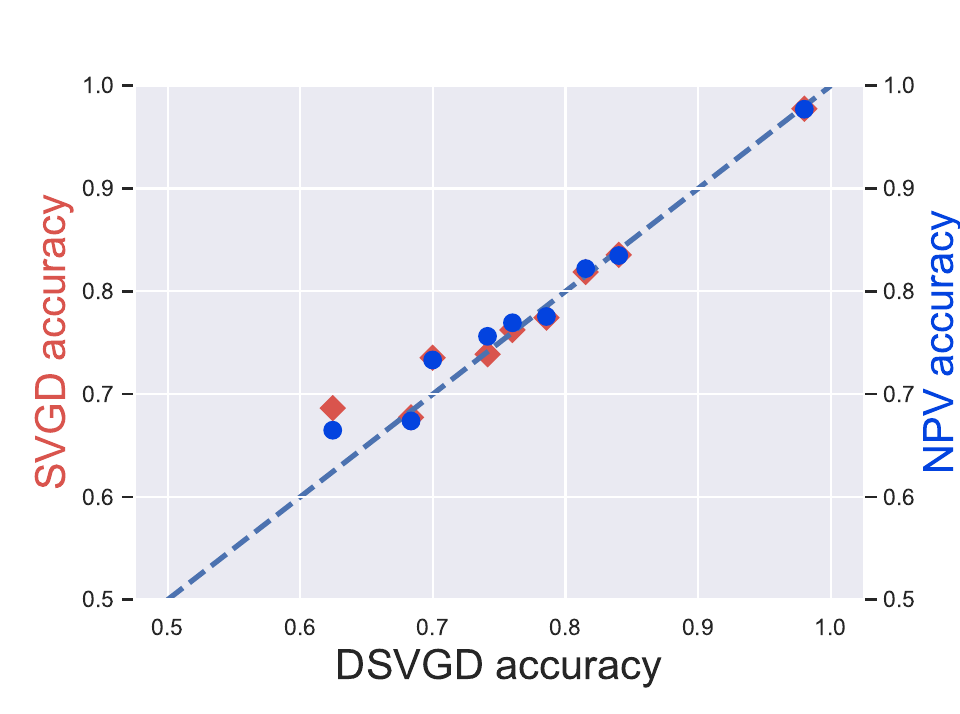}} 
    \subfigure{\includegraphics[height=1.5 in]{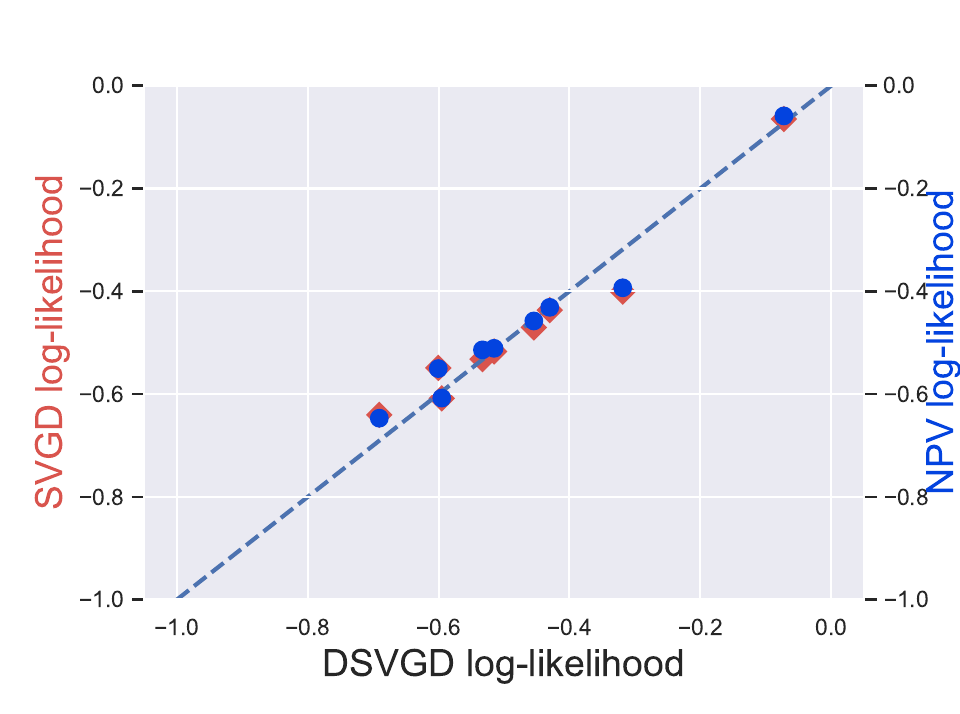}} 
    \caption{Binary classification with Bayesian logistic regression results using the setting in \citet{NPV}: accuracy and log-likelihood for \ac{U-DSVGD} (upper row) and \ac{DSVGD} (bottom row), along with \ac{NPV} and \ac{SVGD}, for various datasets.}
    \label{fig:accuracy_llh}
\end{figure}
This section provides additional results for the Bayesian logistic regression experiment in Sec. \ref{sec:experiments} of the main text. In Fig. \ref{fig:accuracy_llh}, we compare the performance of DSVGD (bottom row), and \ac{U-DSVGD} (top row) both with SVGD and \ac{NPV} \citep{NPV} using the model described in Sec. \ref{sec:experiments}. We use $9$ binary classification datasets summarized in Appendix \ref{app:implementation_details} as used in \citet{SVGD} and \citet{NPV}. We assumed $N=100$ particles. To ensure fairness, we used $L=800$ iterations for SVGD, while U-DSVGD and DSVGD are executed with two agents with half of the dataset split randomly at each agent. We set $I=4$ rounds and $L=L'=200$ local iterations. In Fig. \ref{fig:accuracy_llh}, we plot the accuracy and the log-likelihood of the four algorithms. We observe that both \ac{U-DSVGD} and \ac{DSVGD} perform similarly to \ac{SVGD} and \ac{NPV} over most datasets, while allowing a distributed implementation. We note that \ac{NPV} requires computation of the Hessian matrix which is relatively impractical to compute.
\par
\begin{figure}
	\centering
	\subfigure[]{\includegraphics[width=0.35\textwidth]{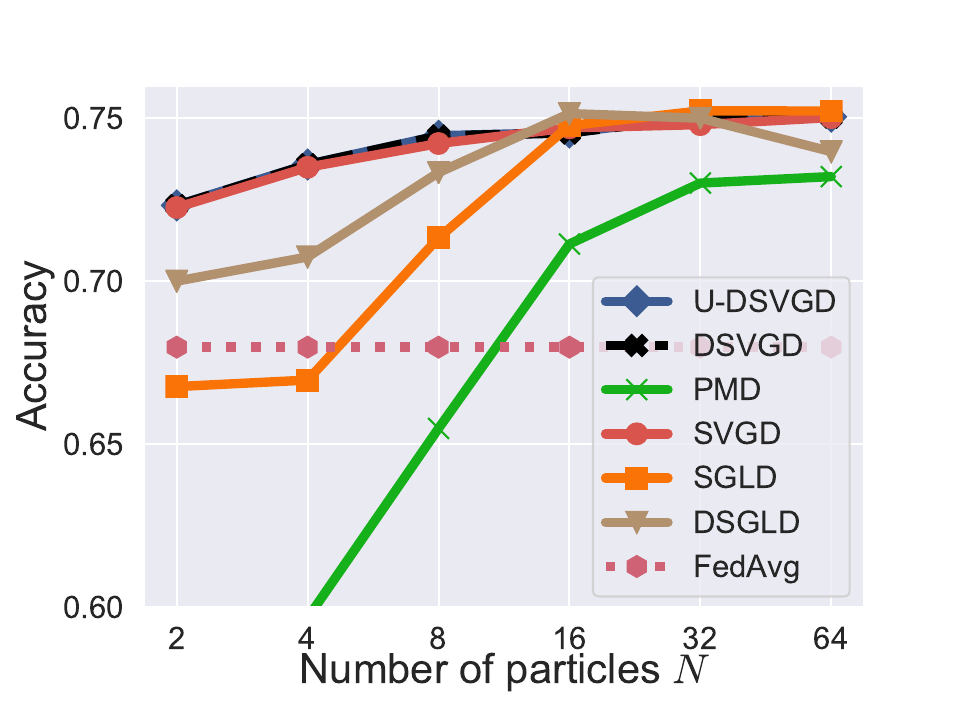}	\label{fig:accuracy_nb_particles}}
	\subfigure[]{\includegraphics[width=0.35\textwidth]{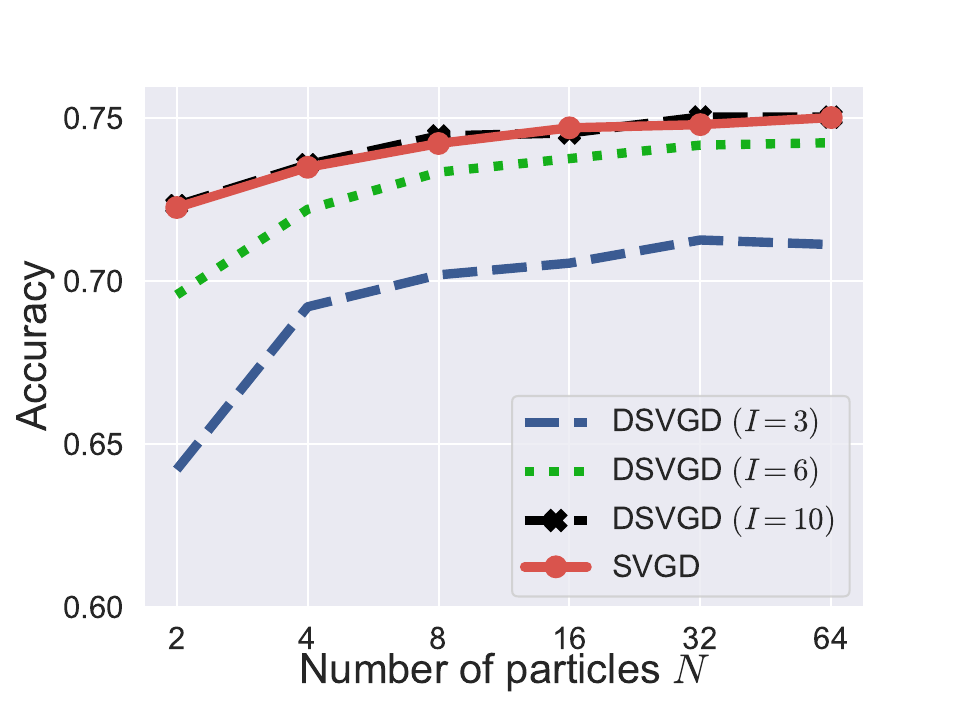}	\label{fig:accuracy_nb_particles_I}}
	\caption{Accuracy as a function of the number of particles $N$ for Bayesian logistic regression on the Covertype dataset: (a) comparison with various benchmarks summarized in Table \ref{table:benchmarks}, and (b) performance for different number of rounds $I$. $L=2000$ iterations were used for centralized schemes while $I \times L=10\times200$ total local iterations were used for decentralized schemes.}
		\label{fig:accuracy_nb_particles_all}
\end{figure}
We plot in Fig. \ref{fig:accuracy_nb_particles} the accuracy as function of the number of particles $N$. \ac{DSGLD} is executed with two agents, where $N/2$ chains per agent are ran for a trajectory of length $4$ and $500$ rounds, which we have found to work best. We found that SVGD, DSVGD and U-DSVGD exhibit the same performance, which is superior to \ac{PMD} and similar to \ac{SGLD} and \ac{DSGLD} when the number of particles increases. Fig. \ref{fig:accuracy_nb_particles_I} plots the accuracy for \ac{DSVGD} for the same setting for different number of communication rounds. We can see that, by increasing the number of particles, i.e., the communication load, one can obtain similar accuracy as for a lower number of particles but with a higher number of communication rounds. For example, $N=8$ with $I=6$ communication rounds achieves similar performance as $N=4$ with $I=10$ communication rounds.
\par
\begin{figure}[h]
    \centering
    \subfigure{\includegraphics[height=1 in]{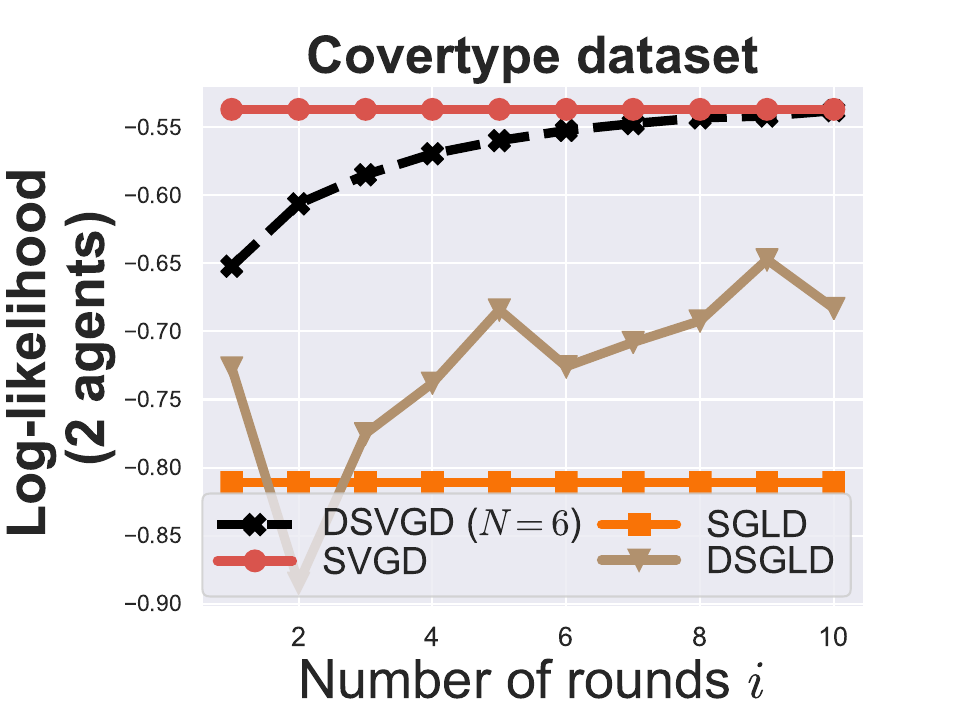}} 
    \subfigure{\includegraphics[height=1 in]{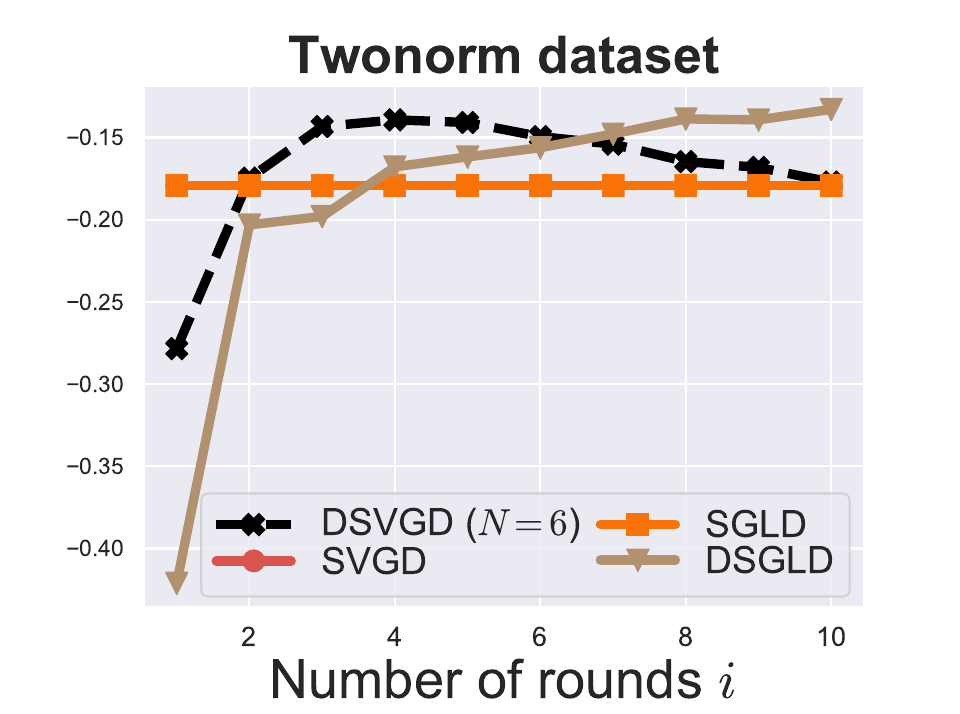}} 
    \subfigure{\includegraphics[height=1 in]{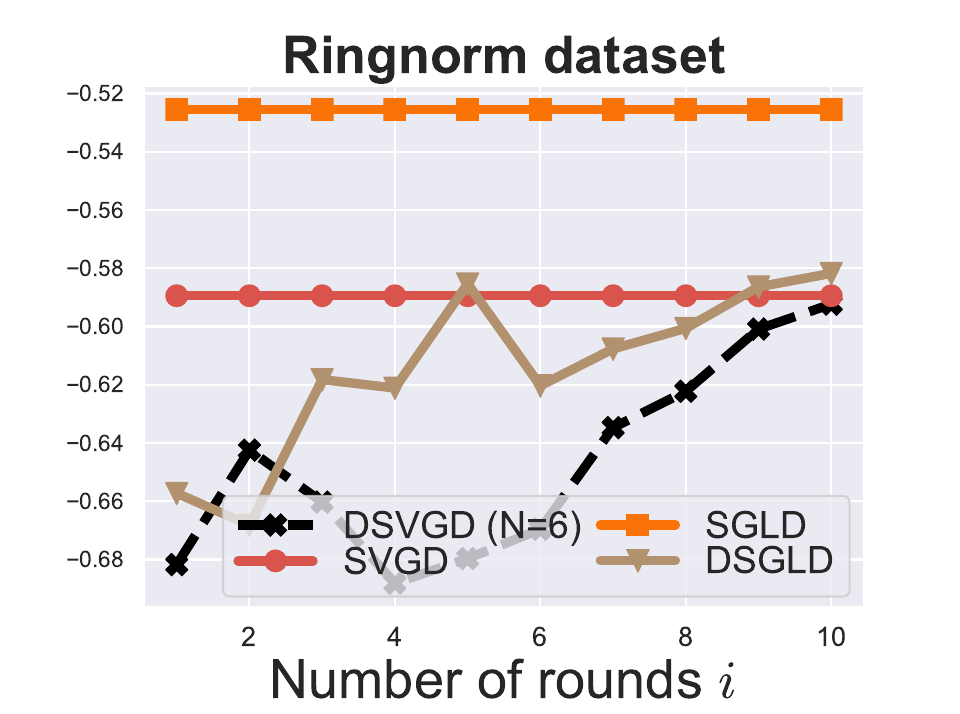}} 
    \subfigure{\includegraphics[height=1 in]{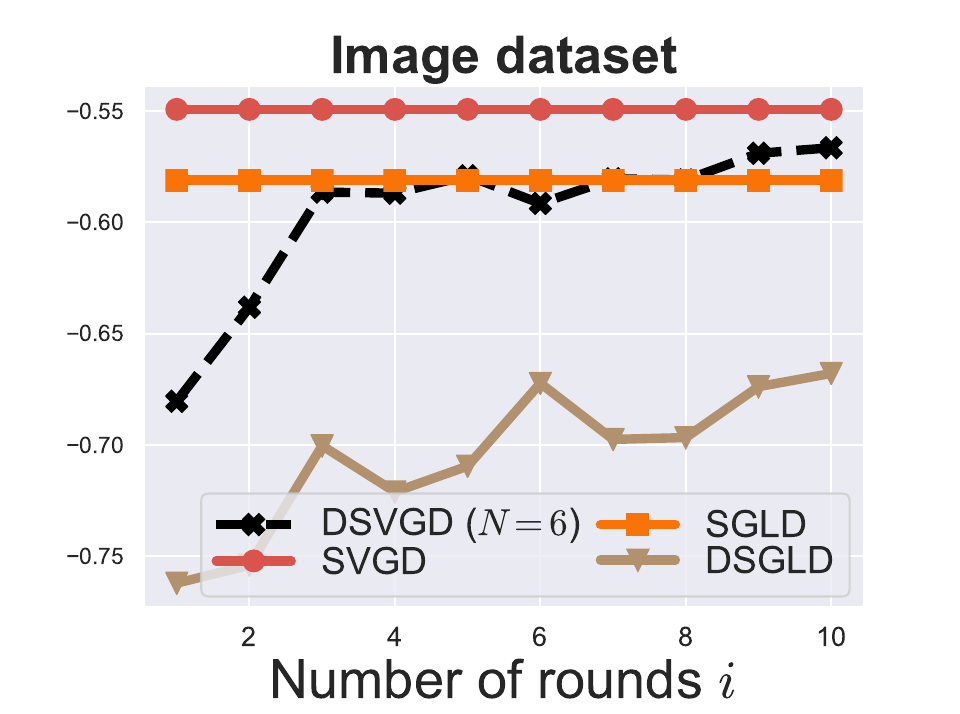}} 
    
    \subfigure{\includegraphics[height=1 in]{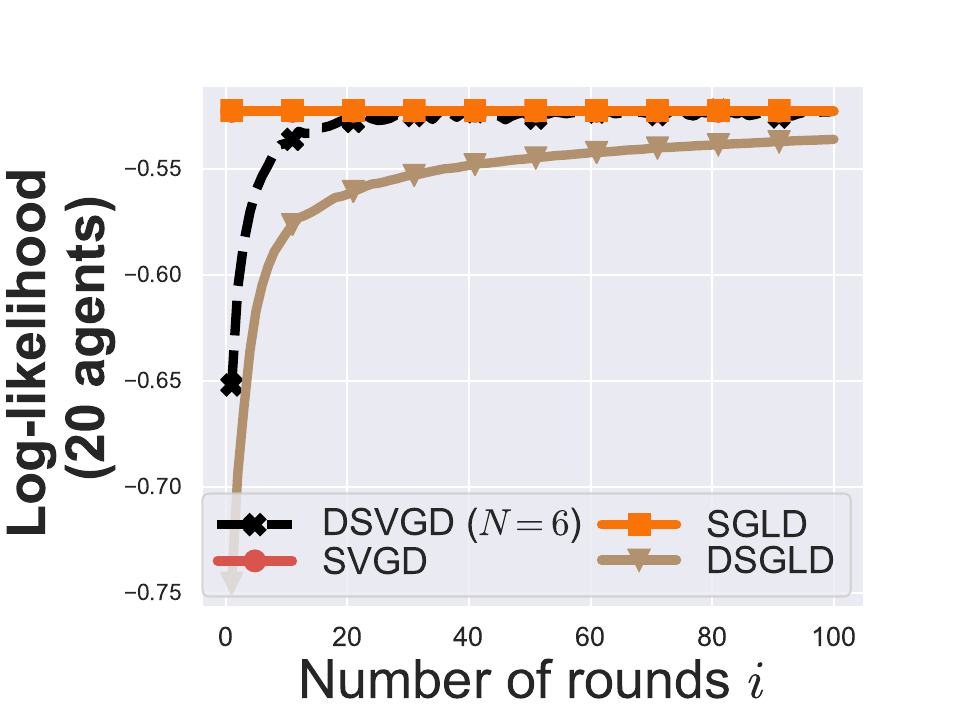}} 
    \subfigure{\includegraphics[height=1 in]{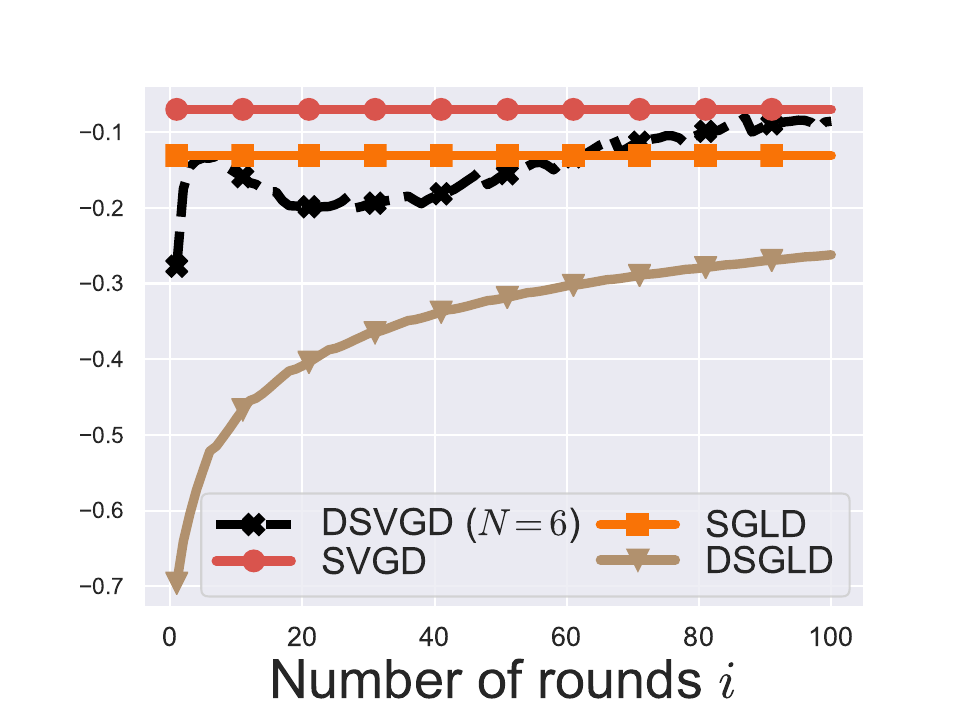}} 
    \subfigure{\includegraphics[height=1 in]{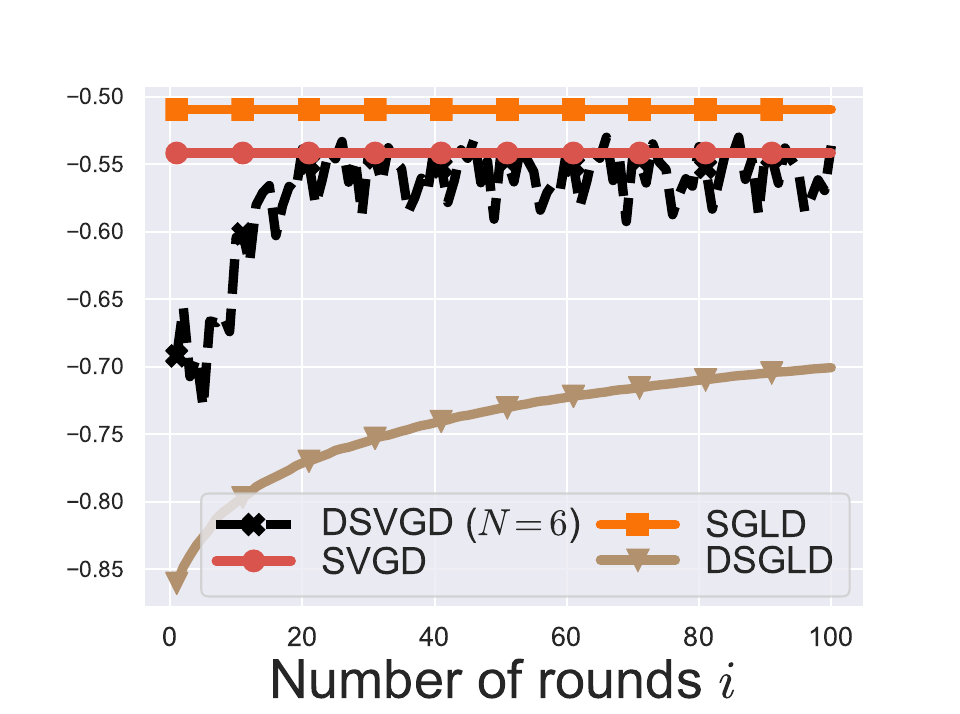}} 
    \subfigure{\includegraphics[height=1 in]{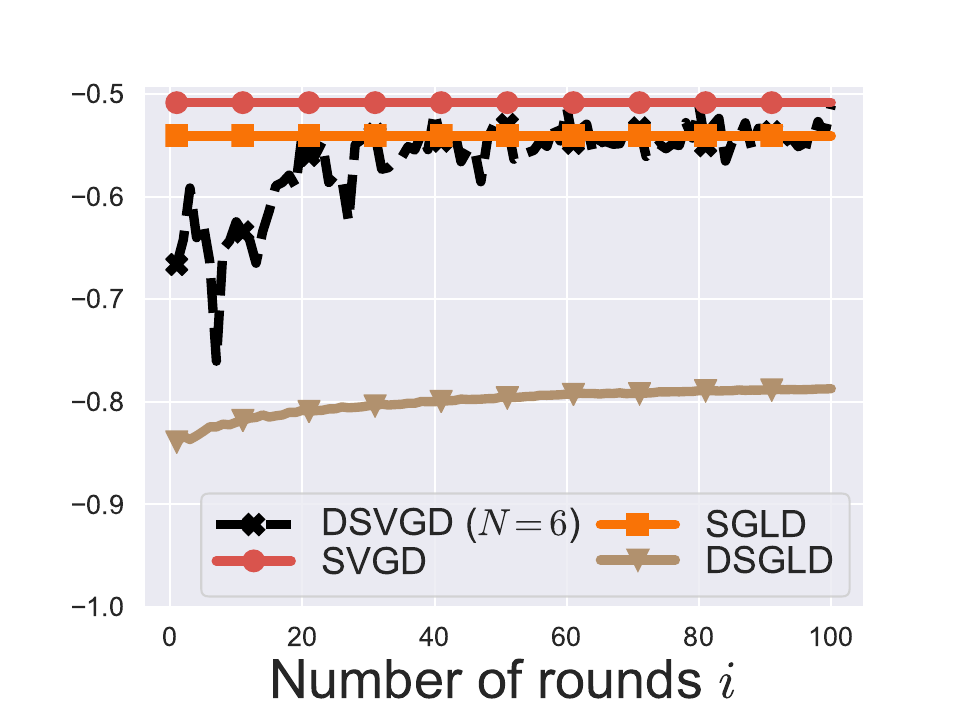}} 
    \caption{Bayesian logistic regression log-likelihood with $K=2$ and $K=20$ agents using the setting in \citet{NPV} comparing \ac{DSVGD} to distributed (\ac{DSGLD}) and centralized (\ac{SVGD} and \ac{SGLD}) schemes  as function of the number of communication rounds $i$. We use $N=6$ particles and fix $L=L^{\prime}=200$. \ac{FedAvg} has been removed as it has a log-likelihood lower than $-1$ in all cases and to allow us to focus on relevant values for \ac{DSVGD}.}
    \label{fig:BLR_i_llh}
\end{figure}

Fig. \ref{fig:BLR_i_llh} is a complementary figure for Fig. \ref{fig:BLR_i_acc} in the main text. It shows that similar conclusions based on accuracy can be made when using the log-likelihood.
\par
\begin{figure}[h]
    \centering
    \subfigure{\includegraphics[height=1 in]{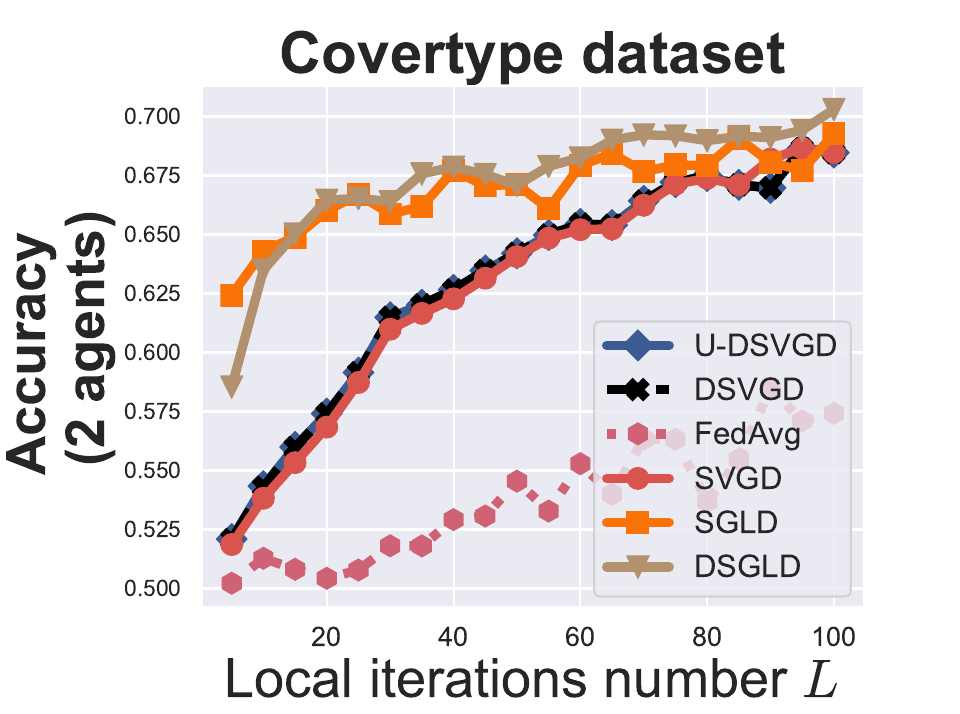}} 
    \subfigure{\includegraphics[height=1 in]{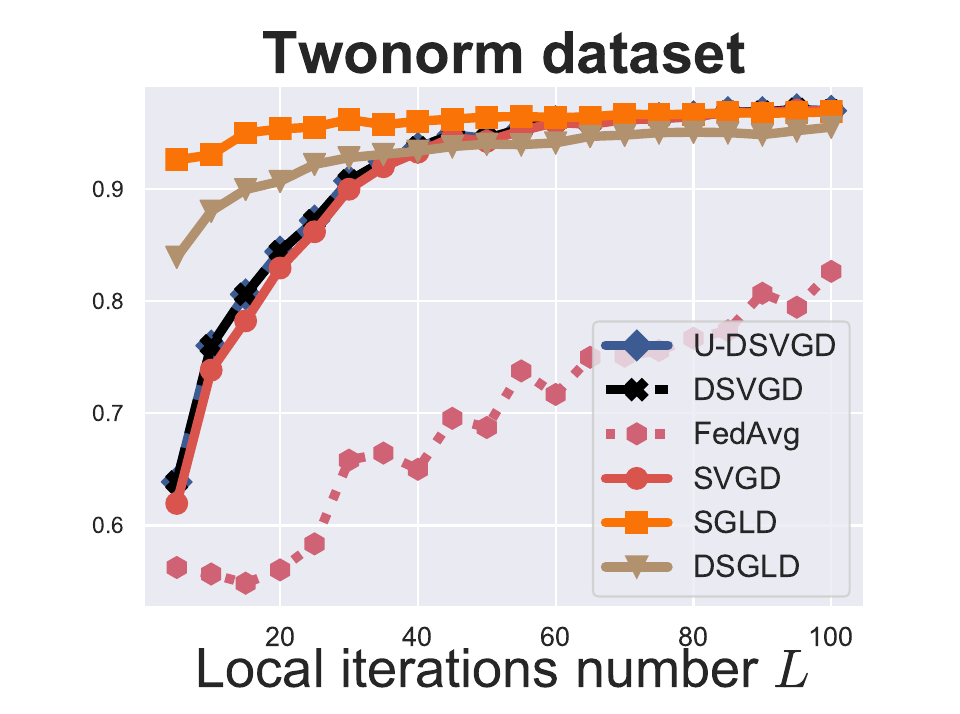}} 
    \subfigure{\includegraphics[height=1 in]{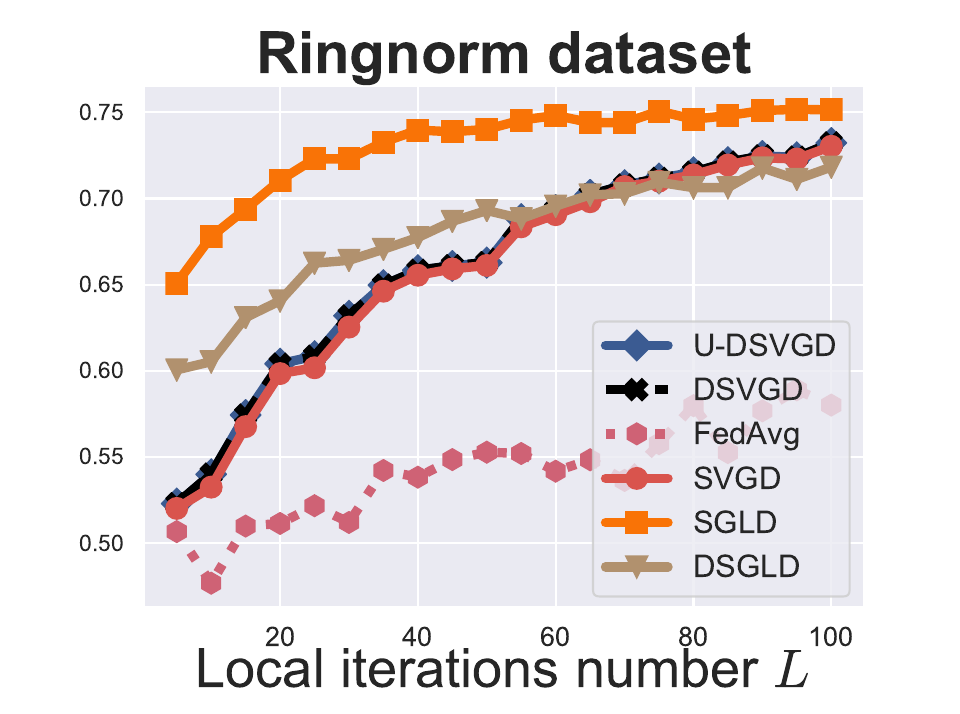}} 
    \subfigure{\includegraphics[height=1 in]{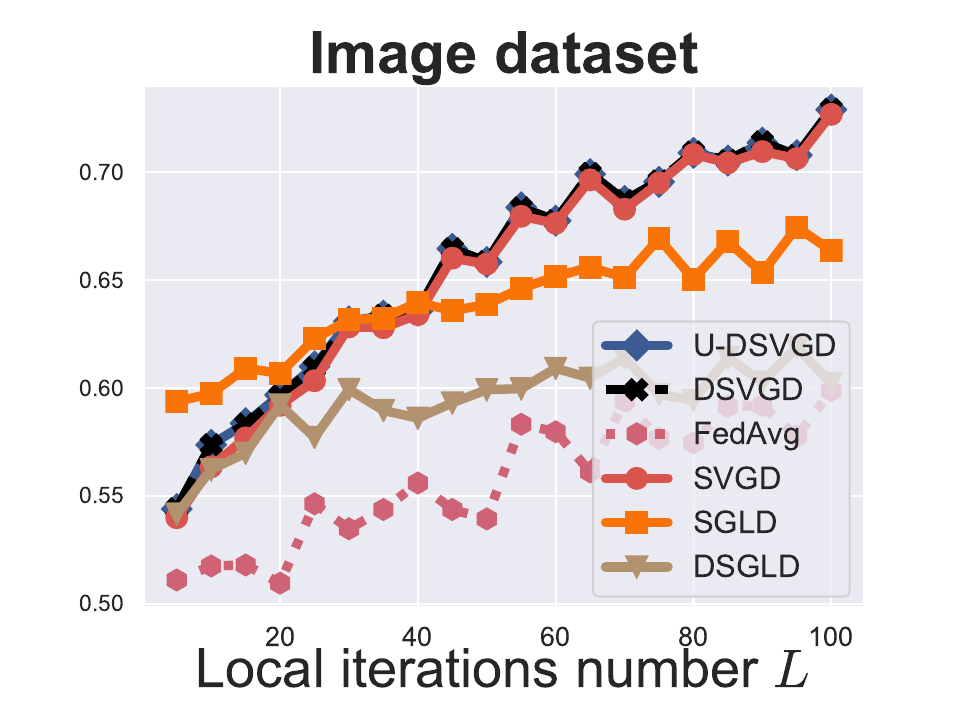}} 
    
    \subfigure{\includegraphics[height=1 in]{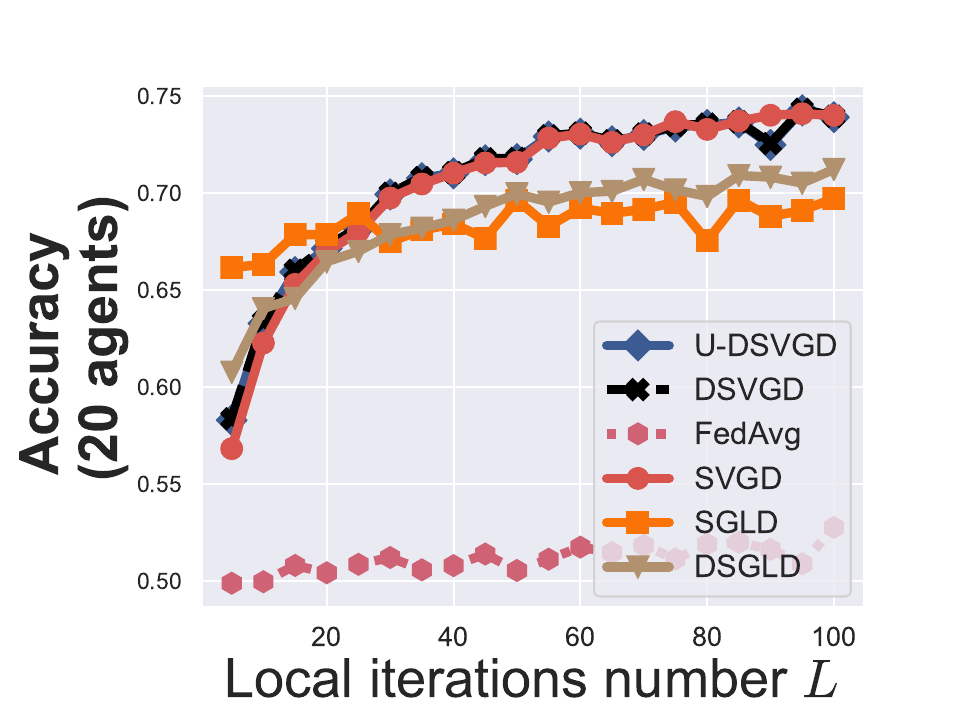}}
    \subfigure{\includegraphics[height=1 in]{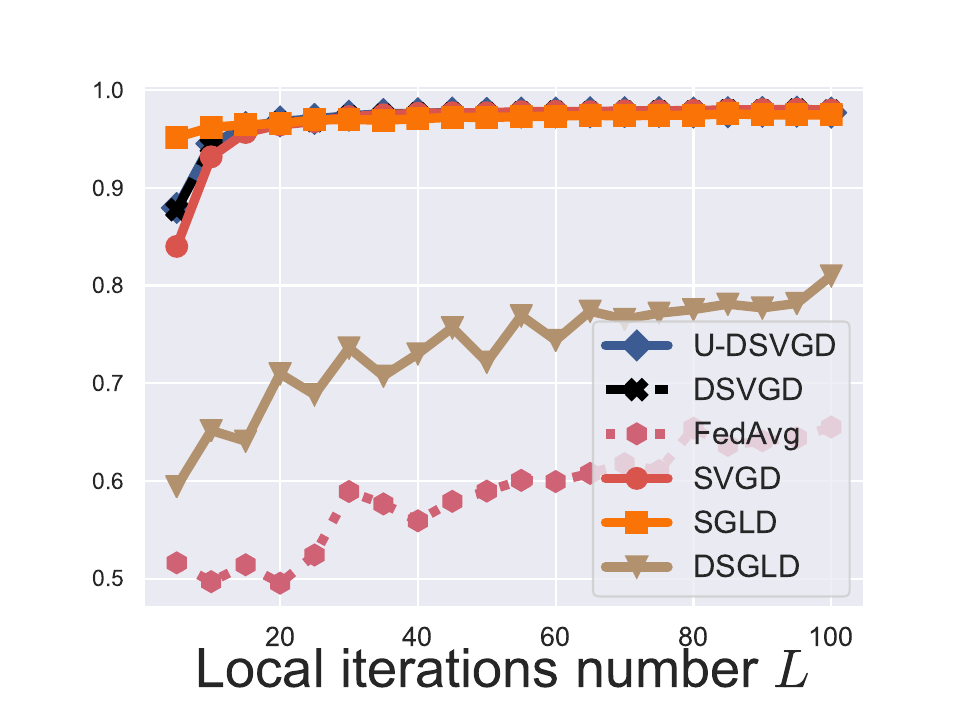}} 
    \subfigure{\includegraphics[height=1 in]{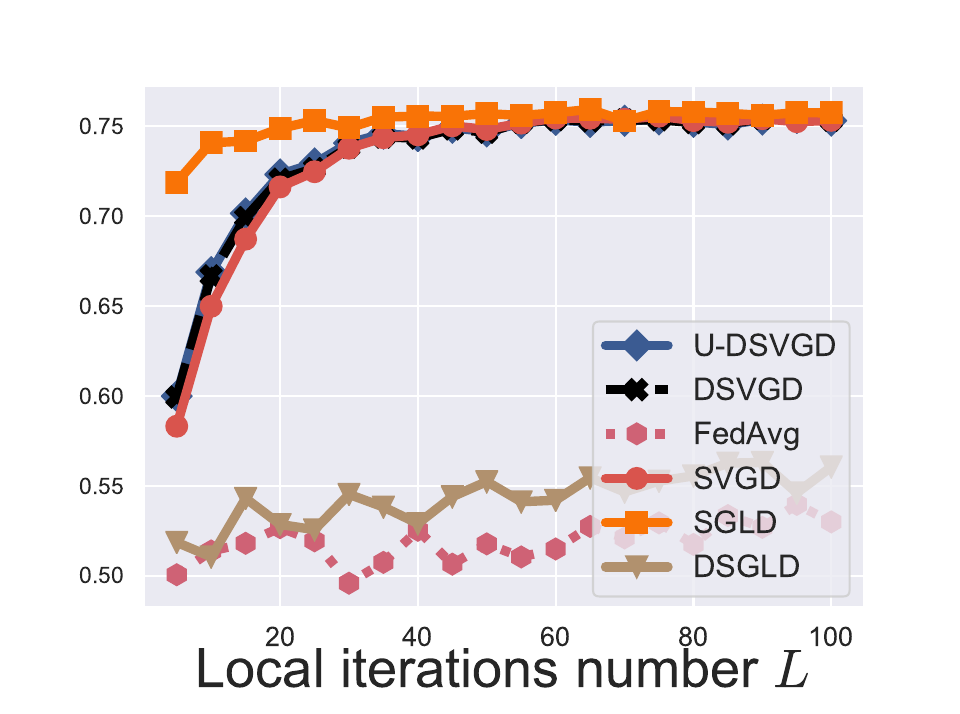}} 
    \subfigure{\includegraphics[height=1 in]{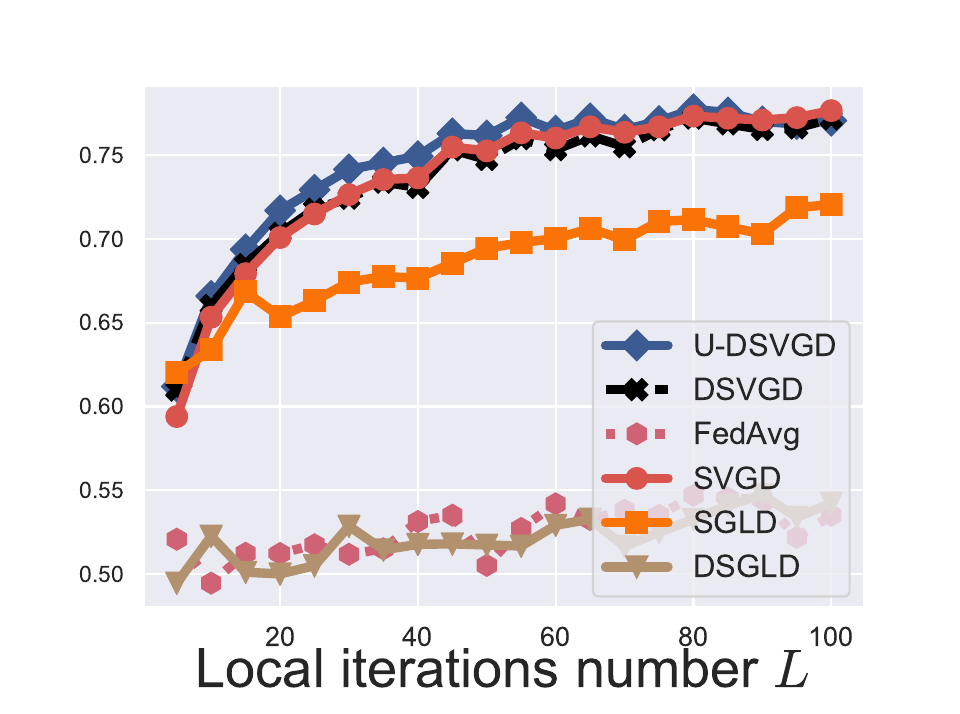}}
    
    \caption{  Bayesian logistic regression accuracy for $K=2$ (top row) and $K=20$ (bottom row) agents using the setting in \citet{NPV} comparing \ac{U-DSVGD} and \ac{DSVGD} to distributed (\ac{DSGLD}) and centralized (\ac{SVGD} and \ac{SGLD}) schemes  as function of the local iterations number $L$. We fix $N=6$ particles, $I=5$ (top row) and $I=20$ (bottom row). }
    \label{fig:BLR_L}
\end{figure}

Fig. \ref{fig:BLR_L} shows the accuracy of \ac{DSVGD} for different datasets as function of the total number $L$ of local iterations. We fix $N=6$, $I=10$, $L=L'=200$ for U-DSVGD, DSGLD and DSVGD while $L=2000$ for SVGD and SGLD.
We observe that \ac{U-DSVGD} and \ac{DSVGD} have similar performance to \ac{SVGD} and that they consistently outperform other schemes for sufficiently high $L$.
\par
\begin{figure}[h]
    \centering
    \subfigure{\includegraphics[height=1 in]{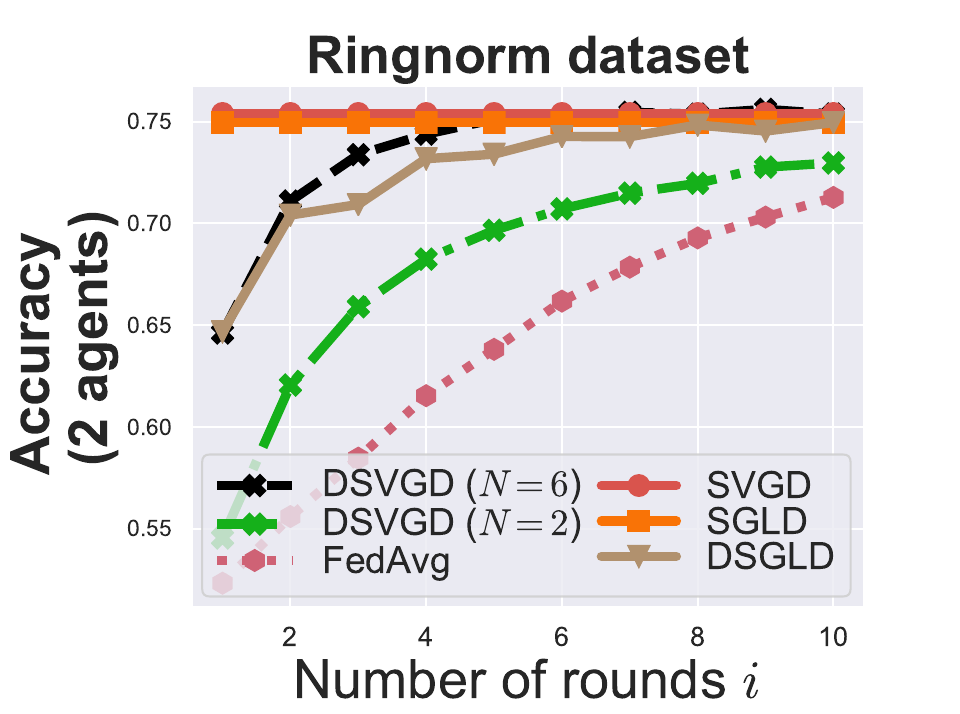}} 
    \subfigure{\includegraphics[height=1 in]{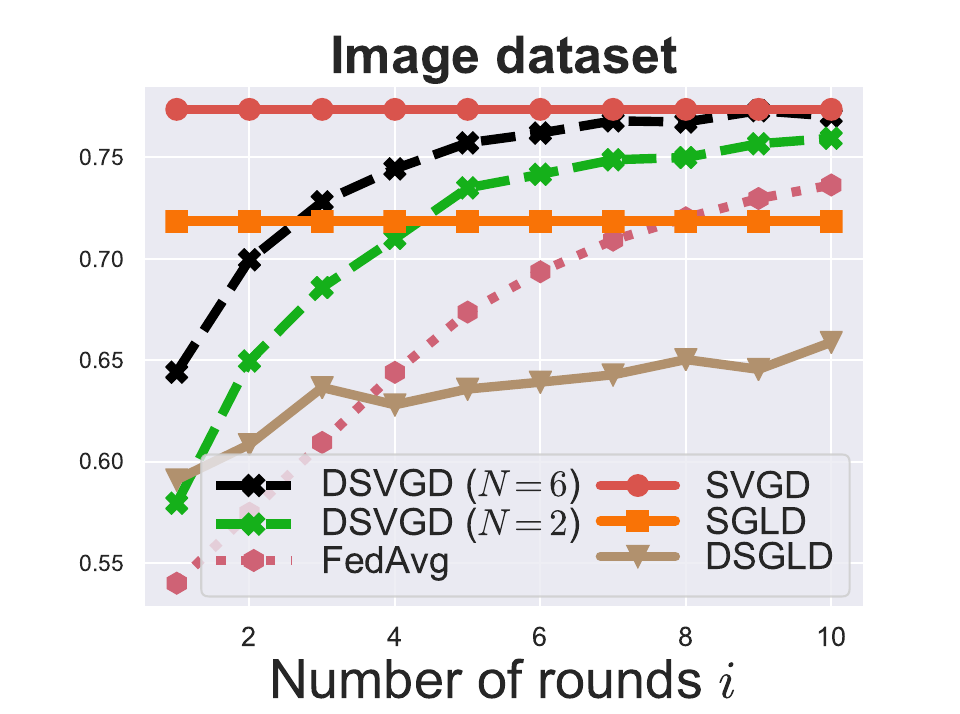}}
    \subfigure{\includegraphics[height=1 in]{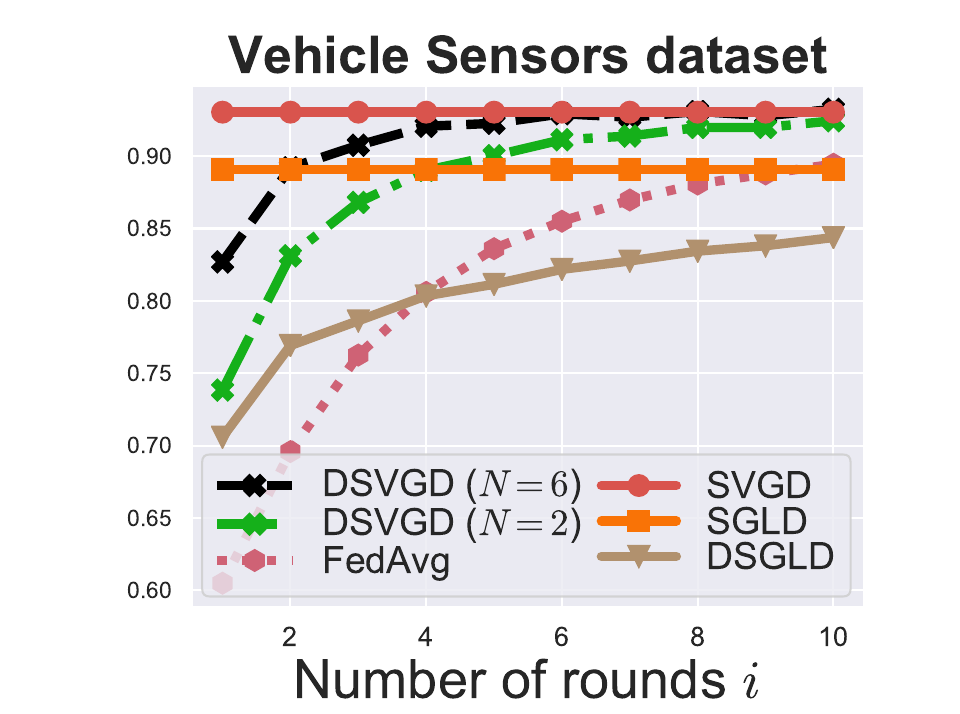}} \\ 
    \BlankLine
    \vspace{-0.755 cm}
    \BlankLine
    \subfigure{\includegraphics[height=1 in]{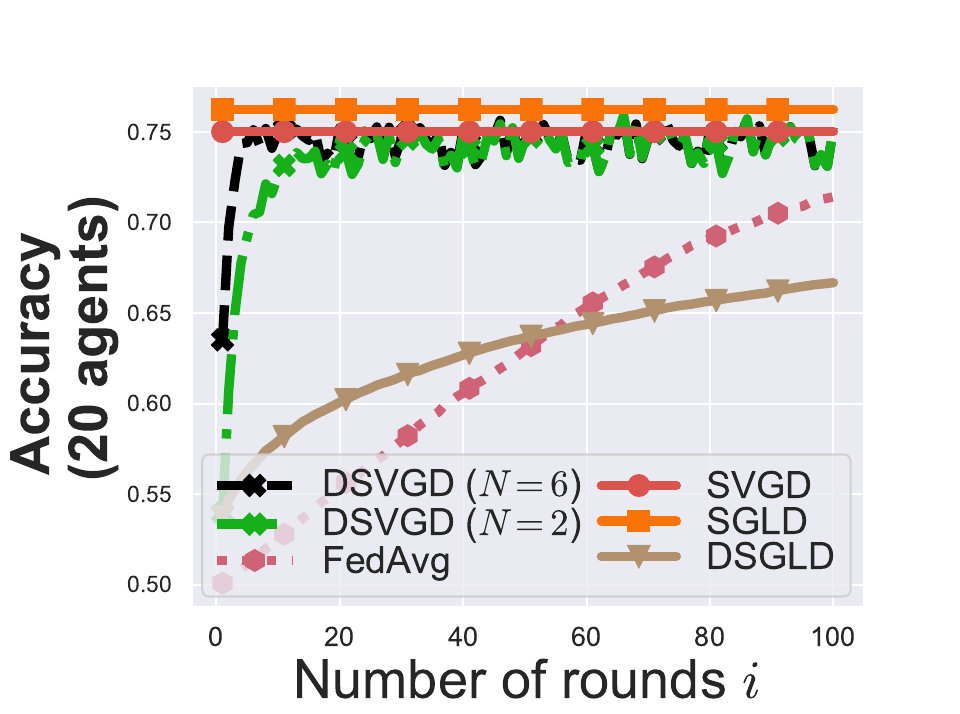}}
    \subfigure{\includegraphics[height=1 in]{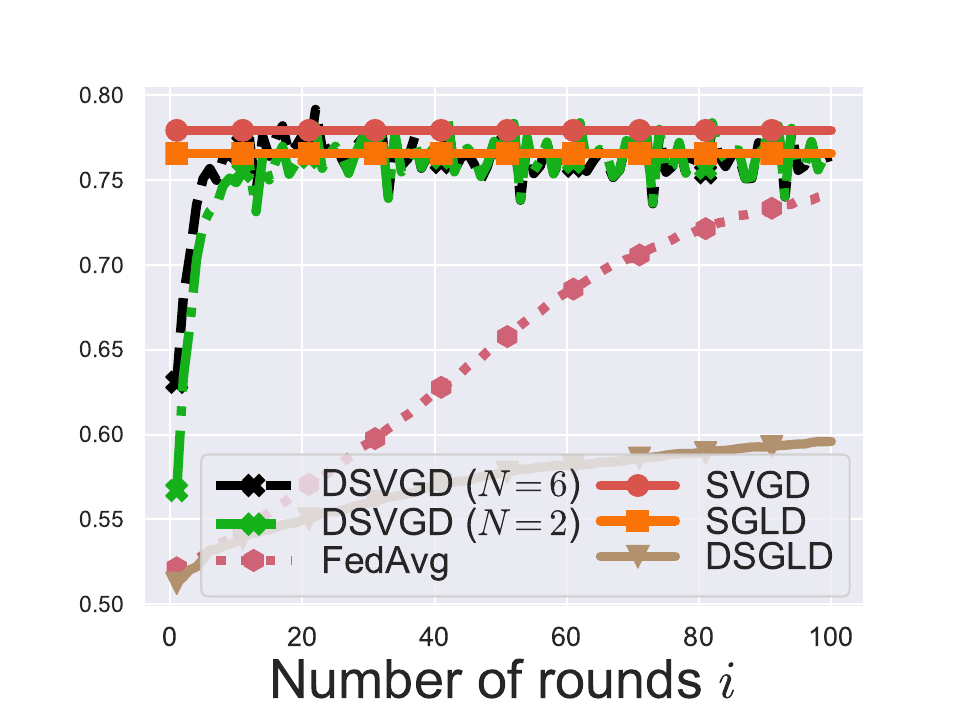}} 
    \subfigure{\includegraphics[height=1in]{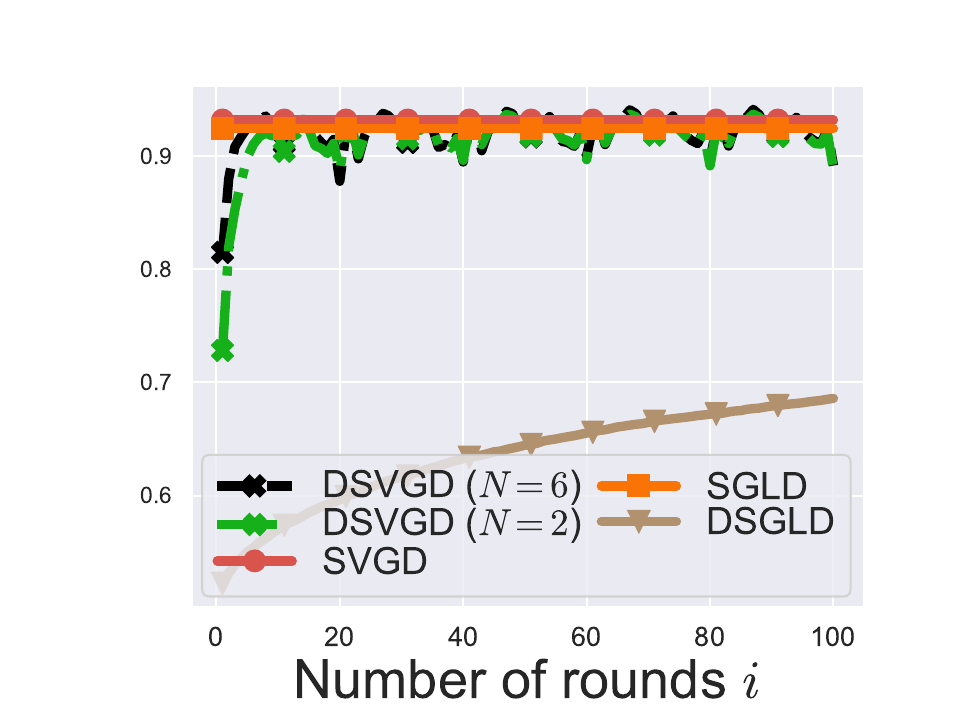}} 

    \caption{Accuracy for Bayesian logistic regression with $K=2$ (top row) and $K=20$ (bottom row) agents under the setting in \citet{NPV} as function of the number of communication rounds $i$, or number of communication rounds ($N=6$ particles, $L=L^{\prime}=200$).}
    \label{fig:BLR_i_acc_app}\vspace{-0.5cm}
\end{figure}

Fig. \ref{fig:BLR_i_acc_app} is complementary to Fig. \ref{fig:BLR_i_acc} in the main text. We note that the slightly noisy behaviour of \ac{DSVGD} with $K=20$ agents is attributed to the small local dataset sizes resulting from splitting the original small datasets.
\par
\begin{figure}[h]
    \centering
    \subfigure{{\includegraphics[height=0.95 in]{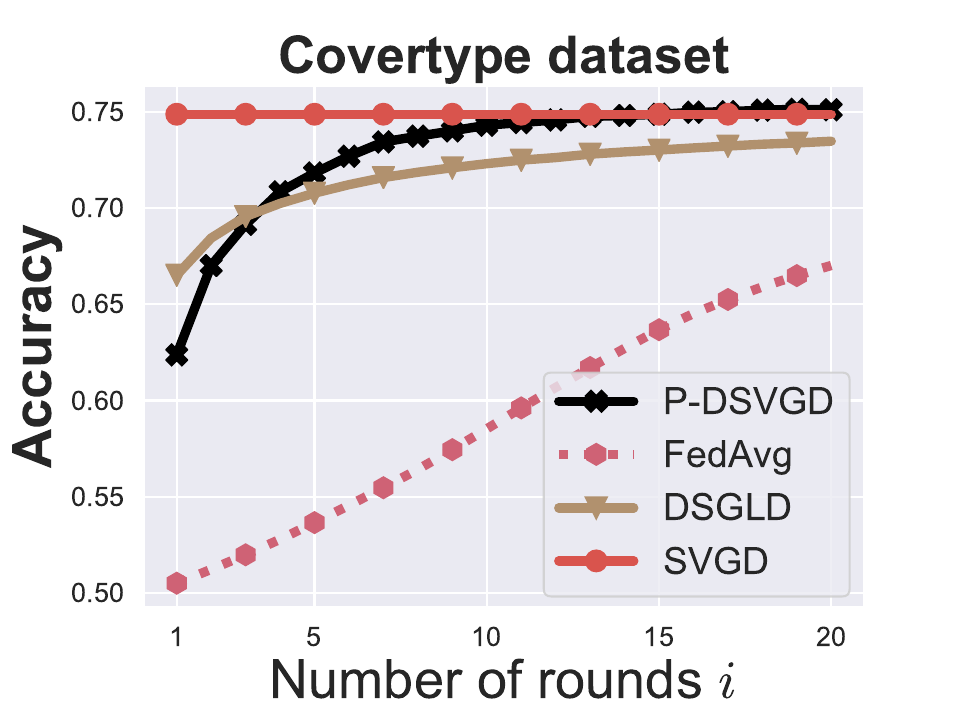}}} 
    \subfigure{\includegraphics[height=0.95 in]{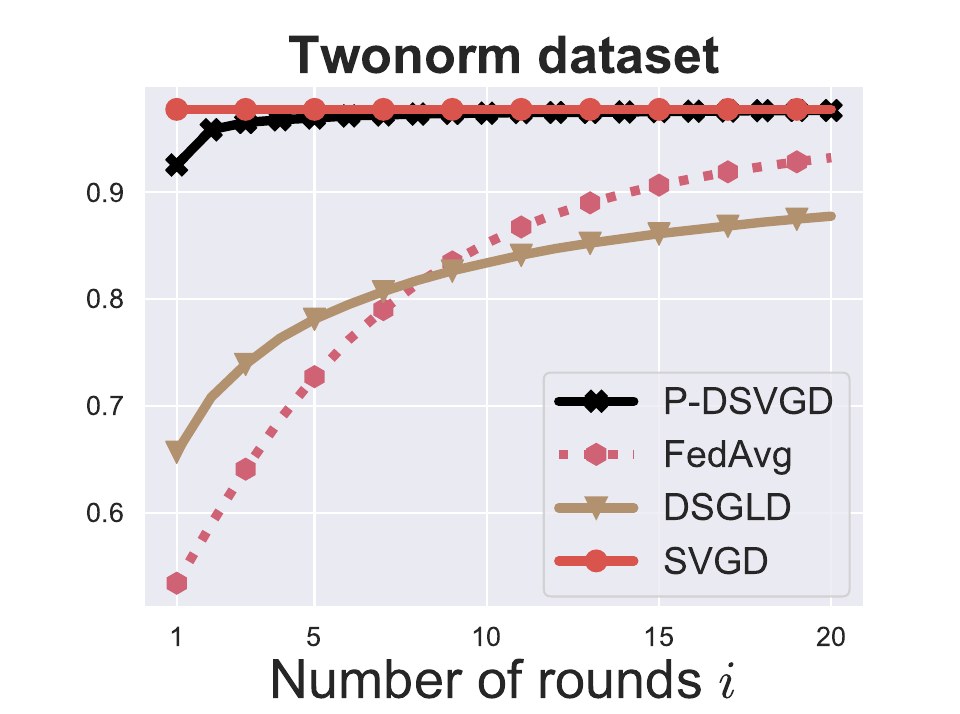}} 
    \subfigure{\includegraphics[height=0.95 in]{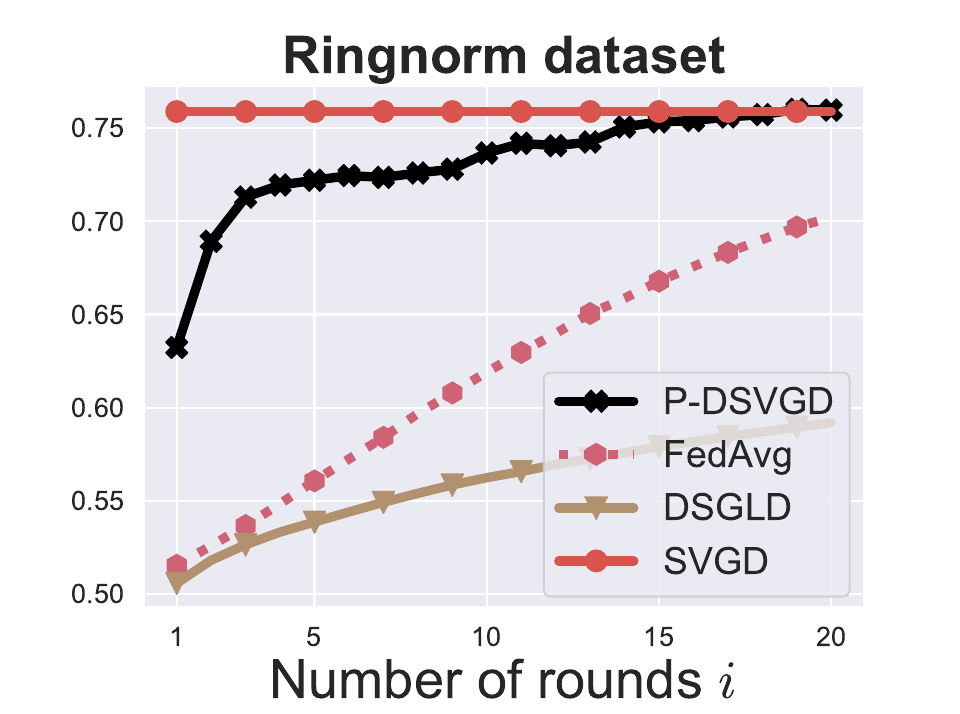}} 
    \subfigure{\includegraphics[height=0.95 in]{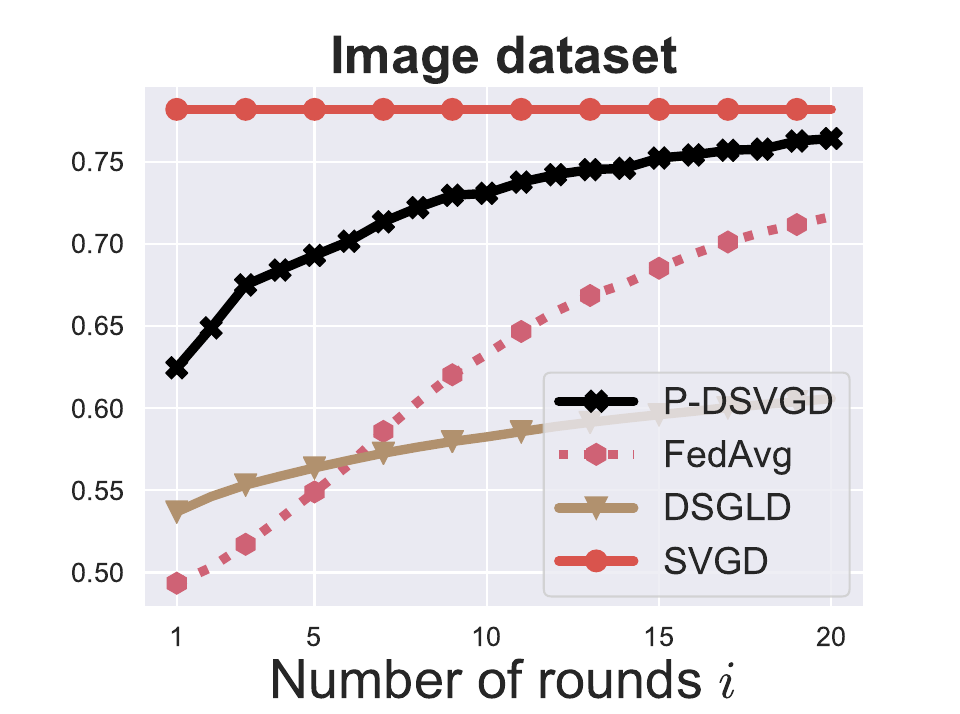}}
    \caption{\textcolor{black}{Accuracy for   Bayesian logistic regression with \ac{P-DSVGD}, \ac{FedAvg}, \ac{DSGLD} and \ac{SVGD} using different datasets with $K=100$ agents and a proportion of $C = 0.2$ randomly scheduled agents.
     \ac{SVGD} was executed for $L=C\times100\times4000$ iterations while we fix $L = L^\prime = L_s=200$ total local iterations for the remaining schemes. We use $N=6$ particles.}}
    \label{fig:BLR_P_I}
\end{figure}

{\color{black}
Finally, Fig. \ref{fig:BLR_P_I} compares the accuracy of P-DSVGD with \ac{FedAvg} and \ac{DSGLD} with $K=100$ agents and a proportion of $0.2$ randomly scheduled agents per communication round. We see that P-DSVGD exhibits similar behaviour and gain over other schemes similarly to DSVGD.}




\subsection{Bayesian Neural Networks for Regression and Classification}
\label{app:distributed_BNN}
\begin{figure}[h]
    \centering
    \subfigure{\includegraphics[height=1 in]{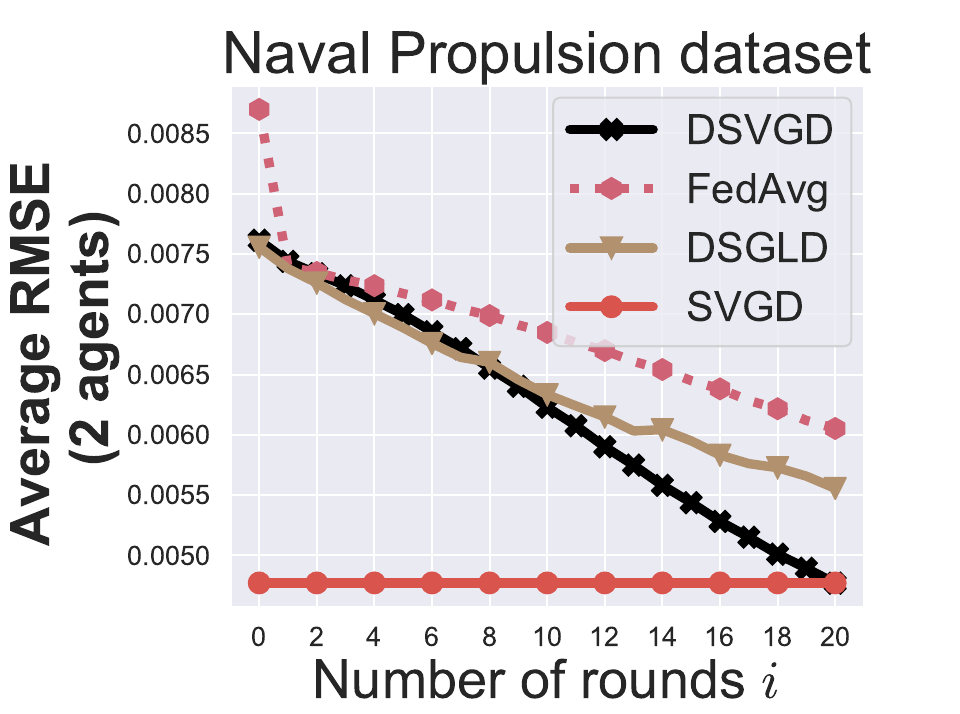}} 
    \subfigure{\includegraphics[height=1 in]{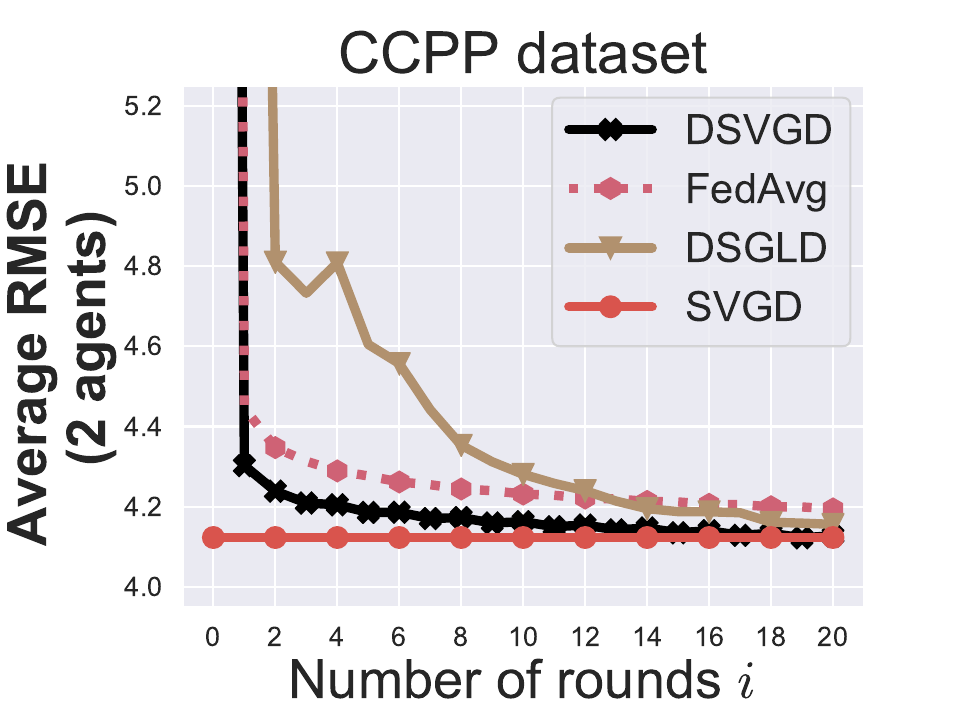}}\\
    \BlankLine
    \vspace{-0.755 cm}
    \BlankLine
    \subfigure{\includegraphics[height=1 in]{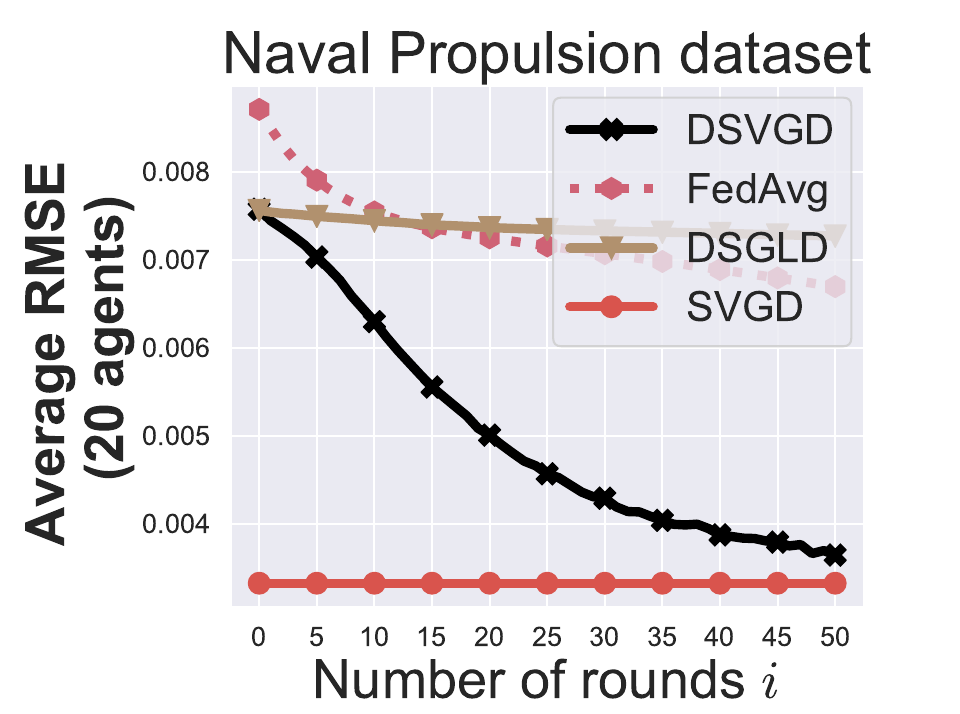}} 
    \subfigure{\includegraphics[height=1 in]{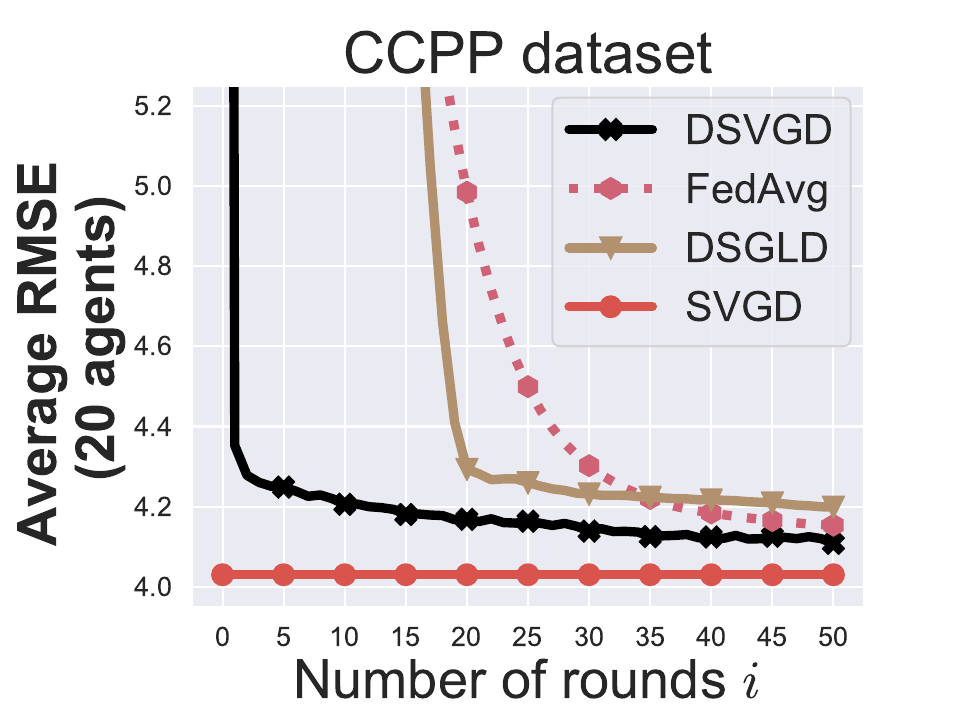}} 
    
    \caption{Average Root Mean Square Error (RMSE) as a function of the number of communication rounds $i$, or number of communication rounds, for regression using Bayesian neural networks with a single hidden layer of ReLUs under the setting of \citet{PBP_lobato}, with $K=2$ (top row) and $K=20$ (bottom row) agents. ($N=20$, $L=L^\prime=200$ and $50$ hidden neurons).}
    \label{fig:BNN_i_regression_app}
\end{figure}

\begin{figure}[h]
    \centering
    \subfigure{{\includegraphics[height=1 in]{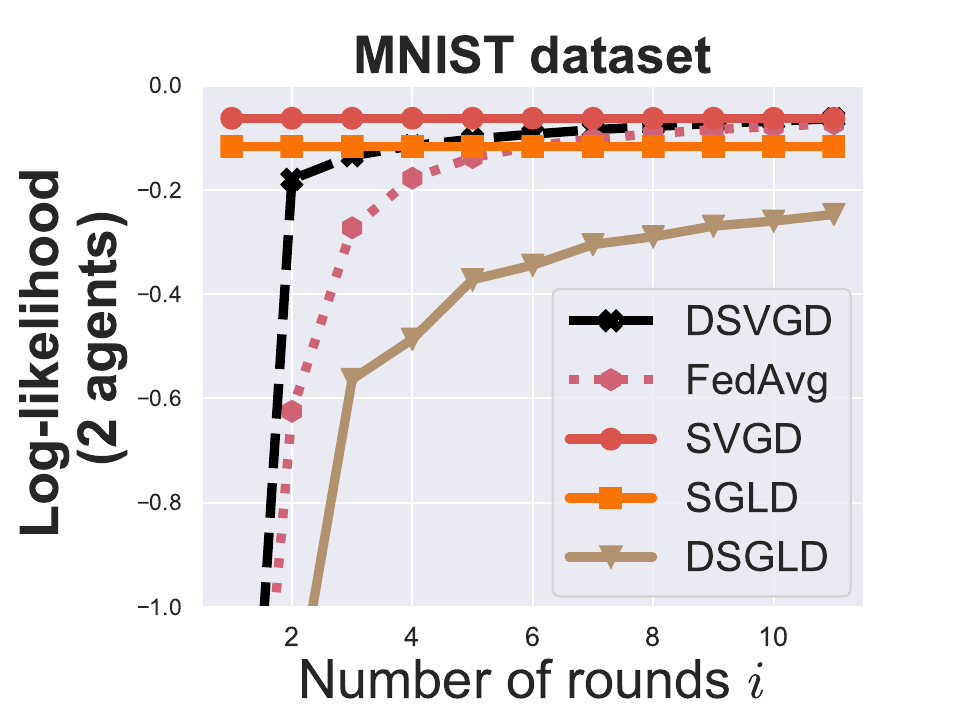}}} 
    \subfigure{\includegraphics[height=1 in]{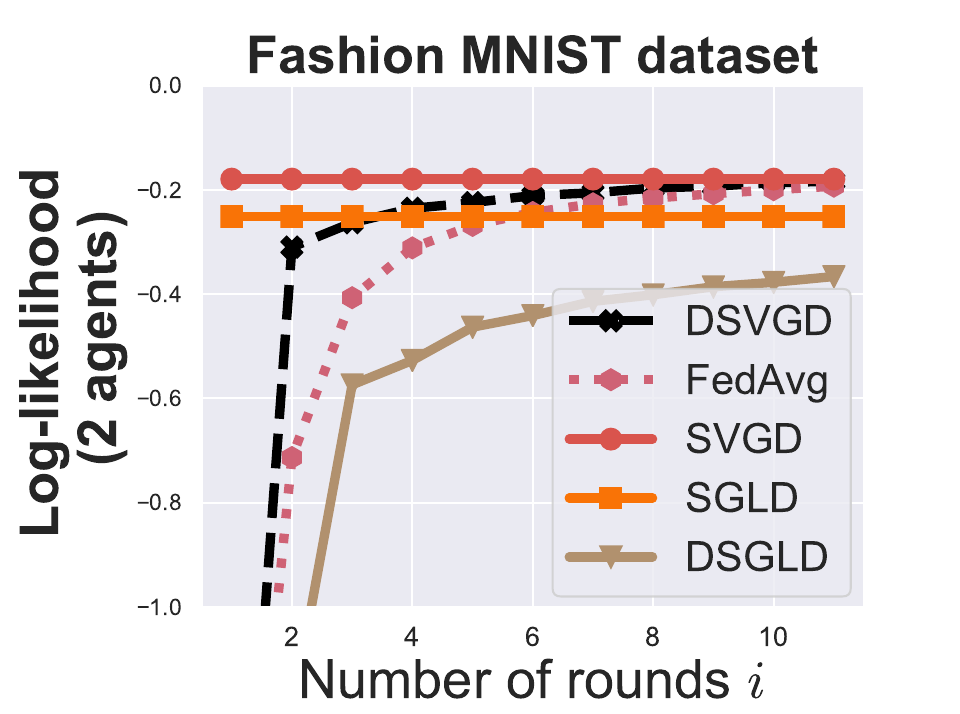}}\\
    \BlankLine
    \vspace{-0.755 cm}
    \BlankLine
    \subfigure{\includegraphics[height=1 in]{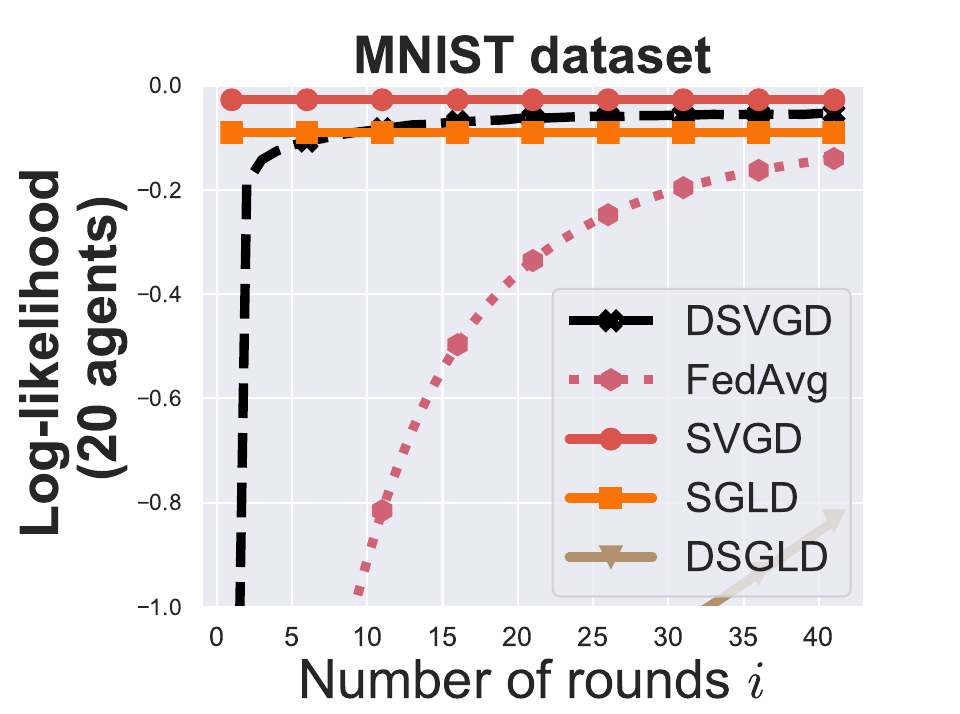}} 
    \subfigure{\includegraphics[height=1 in]{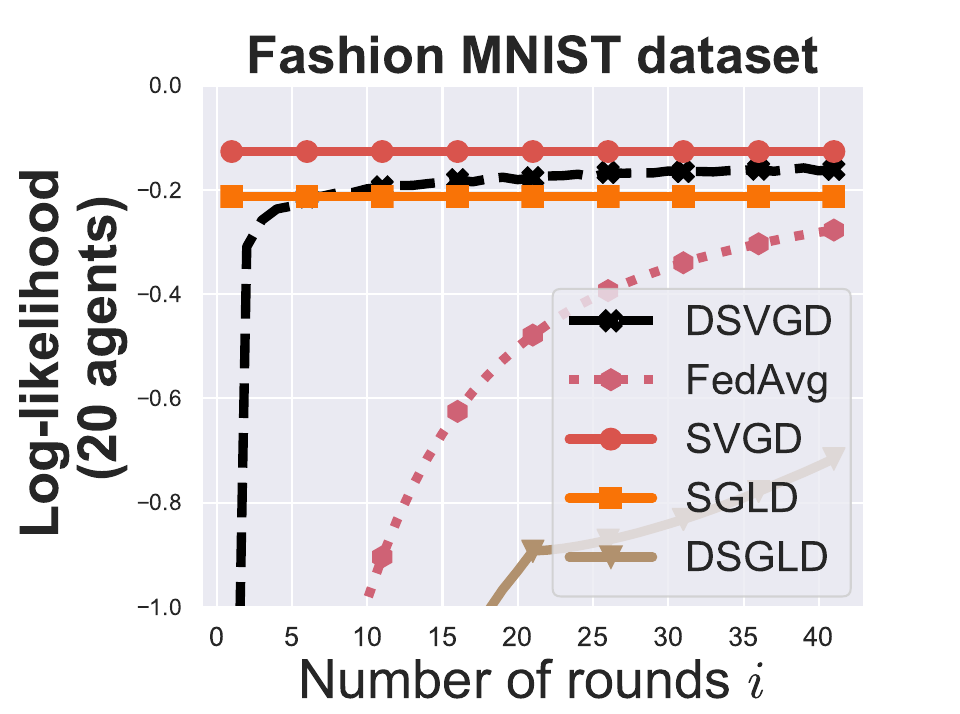}} 
    
    \caption{Log-likelihood for multi-label classification using Bayesian neural networks with a single hidden layer of $100$ neurons as function of the number of communication rounds $i$, or number of communication rounds, using MNIST and Fashion MNIST with $K=2$ (top row) and $K=20$ (bottom row) agents ($N=20$, $L=L^\prime=200$.}
    \label{fig:BNN_i_classification_app}
\end{figure}

This part contains additional results on regression and multilabel classification experiments using Bayesian Neural Networks. Figures \ref{fig:BNN_i_regression_app} and \ref{fig:BNN_i_classification_app} are complementary to Figures \ref{fig:BNN_i_regression} and \ref{fig:BNN_i_classification} in the main text and validate our conclusions using additional datasets for regression and the log-likelihood metric for multi-label classification.
\clearpage
\subsection{Reliability Plots and Maximum Calibration Error}
\label{app:MCE}
\begin{figure}[h]
    \centering
    \subfigure{{\includegraphics[height=0.85 in]{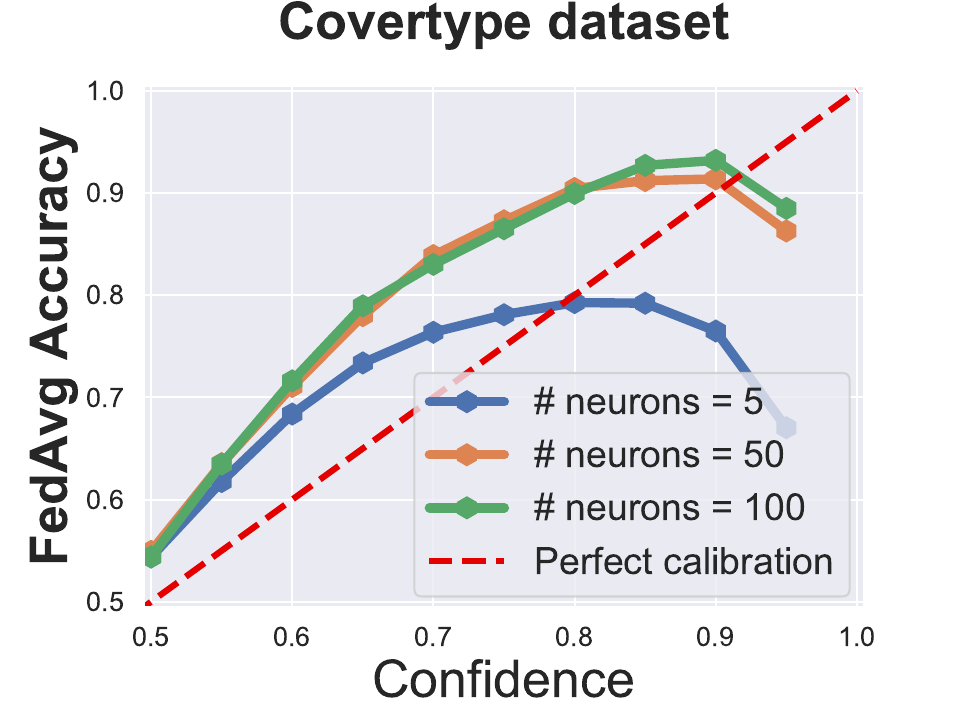}}}\hspace{-0.4cm}
    \subfigure{{\includegraphics[height=0.85 in]{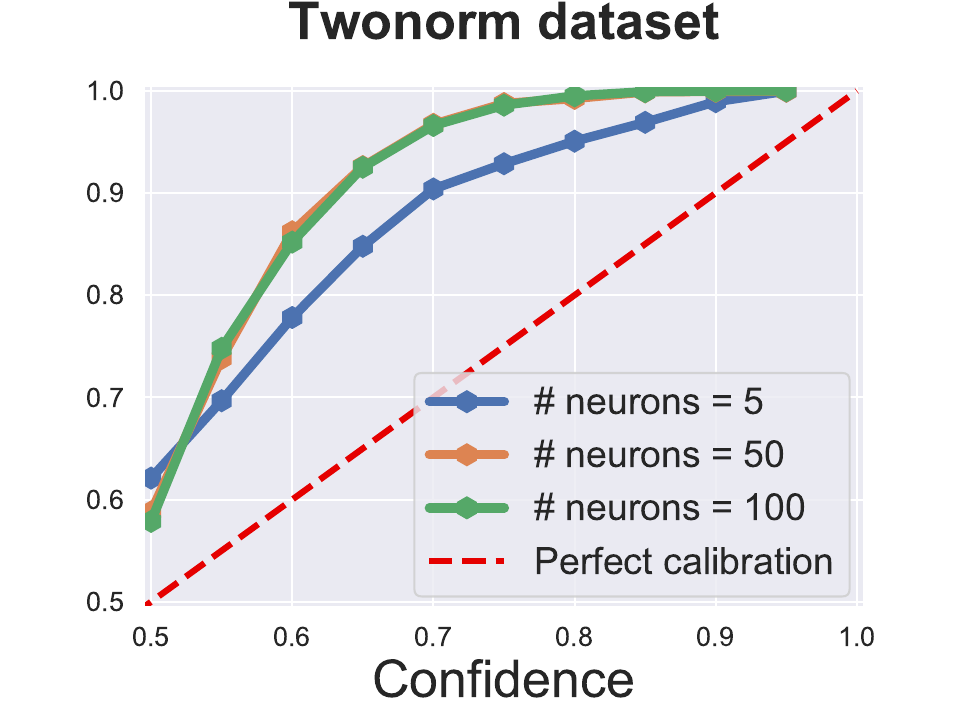}}}\hspace{-0.4cm} 
    \subfigure{{\includegraphics[height=0.85 in]{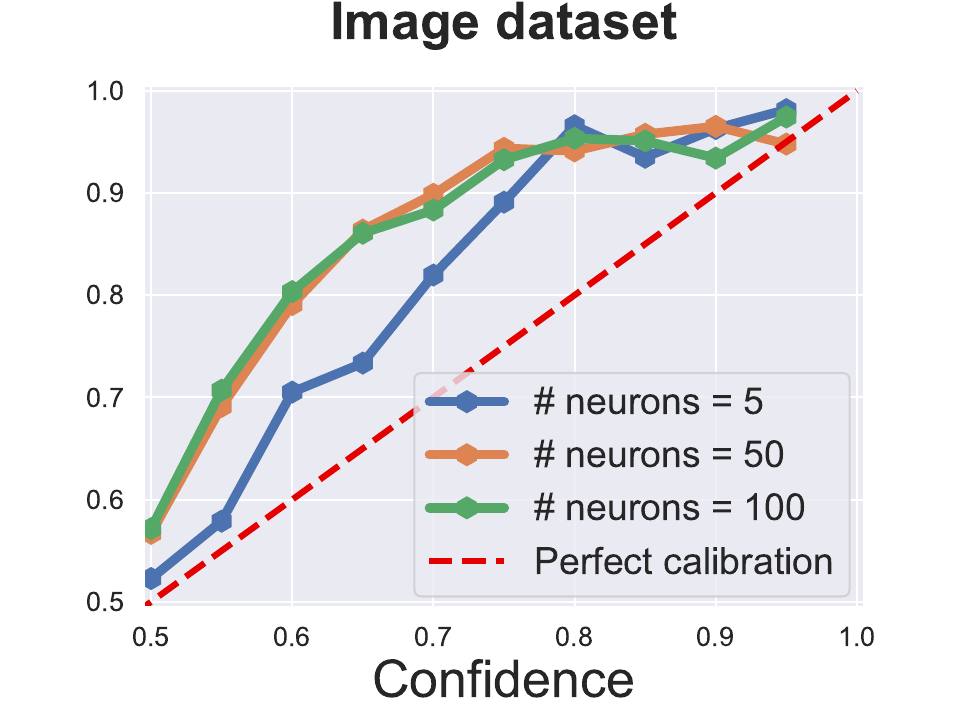}}} \hspace{-0.4cm}
    \subfigure{{\includegraphics[height=0.85 in]{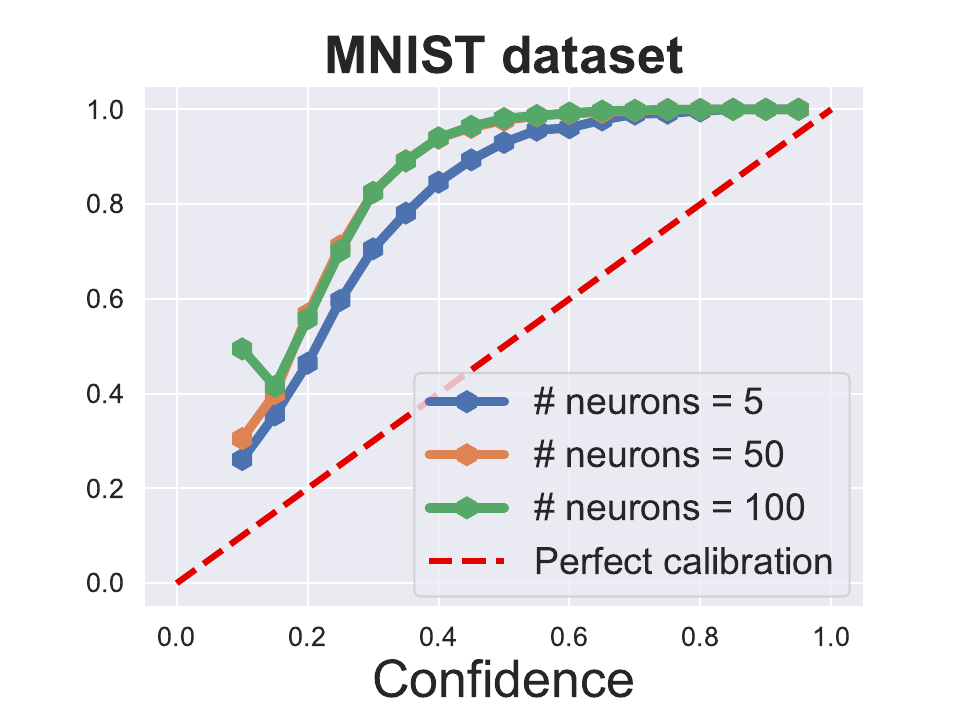}}} \hspace{-0.4cm}
    \subfigure{{\includegraphics[height=0.85 in]{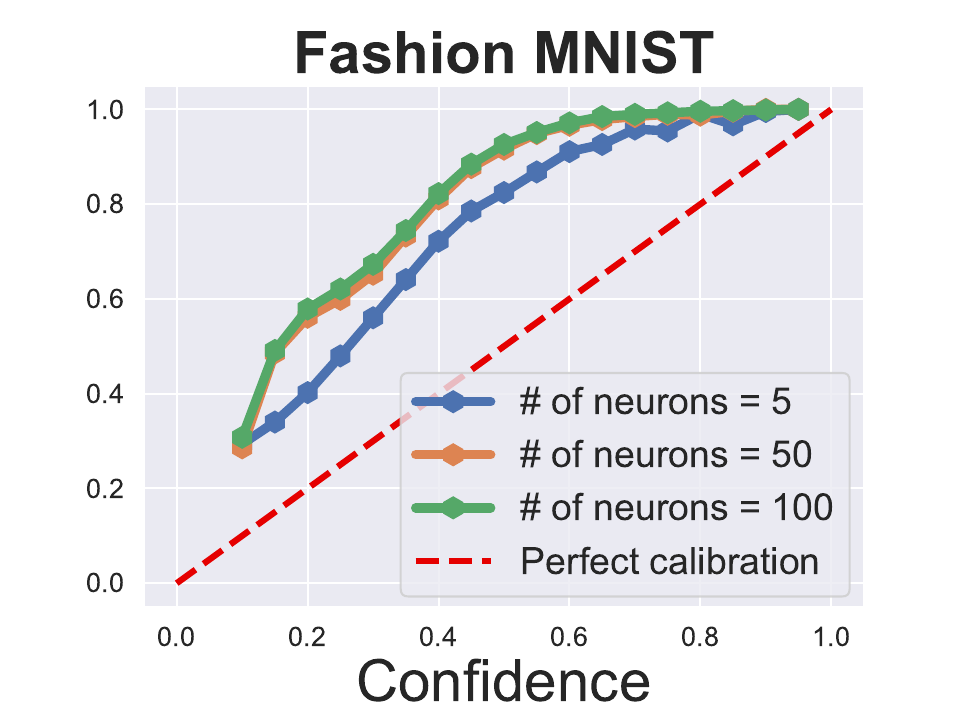}}} \\
    \BlankLine
    \vspace{-0.75 cm}
    \BlankLine
    \subfigure{{\includegraphics[height=0.85 in]{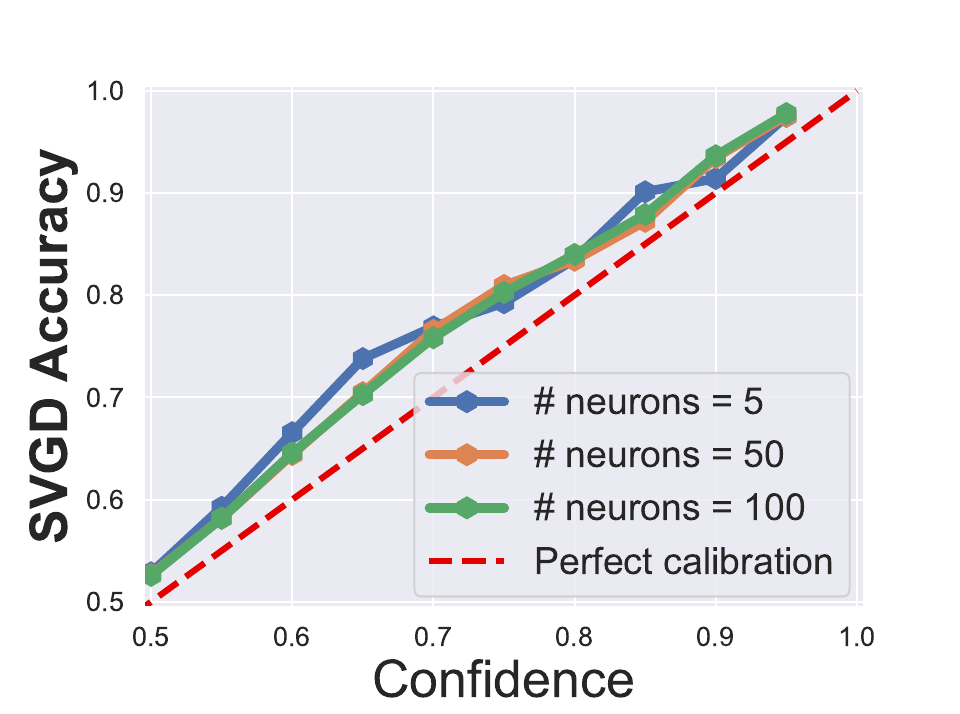}}}\hspace{-0.4cm}
    \subfigure{{\includegraphics[height=0.85 in]{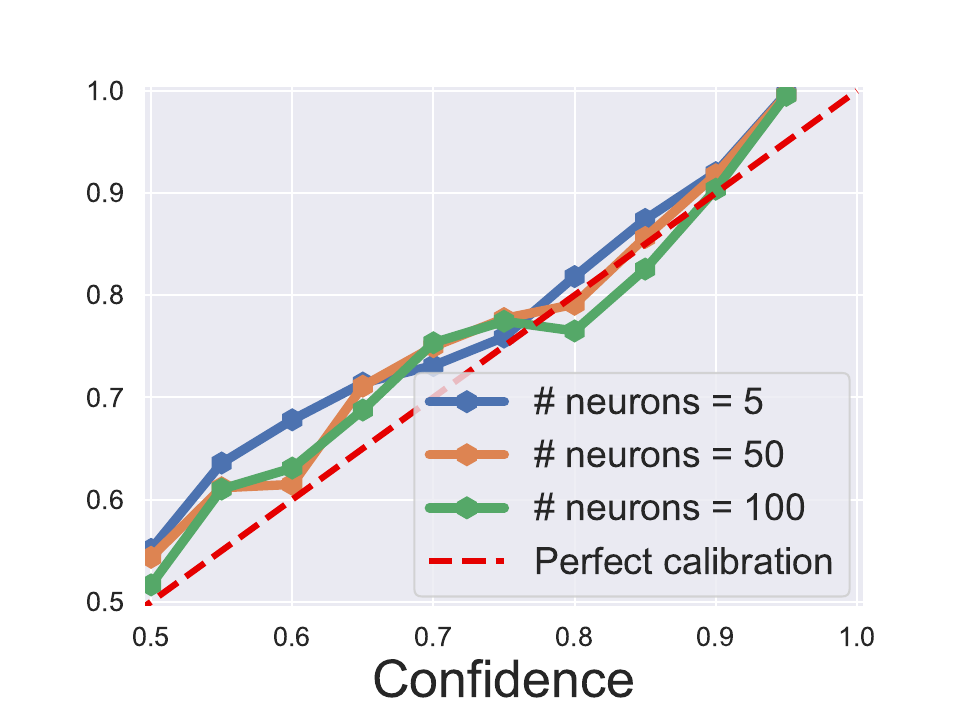}}} \hspace{-0.4cm}
    \subfigure{{\includegraphics[height=0.85 in]{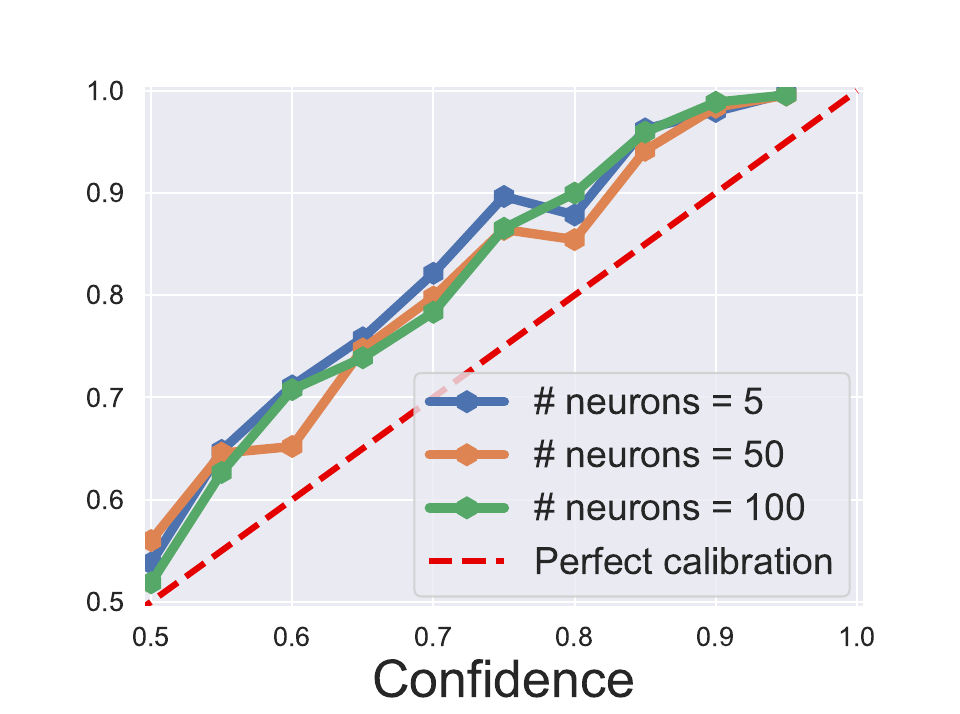}}} \hspace{-0.4cm}
    \subfigure{{\includegraphics[height=0.85 in]{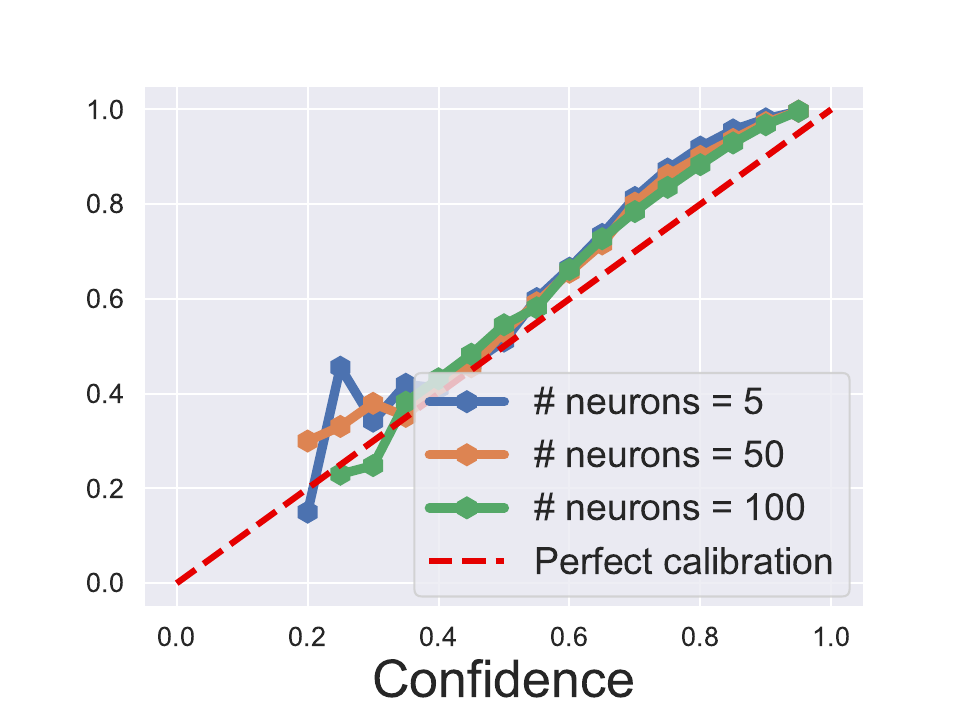}}}\hspace{-0.4cm}
    \subfigure{{\includegraphics[height=0.85 in]{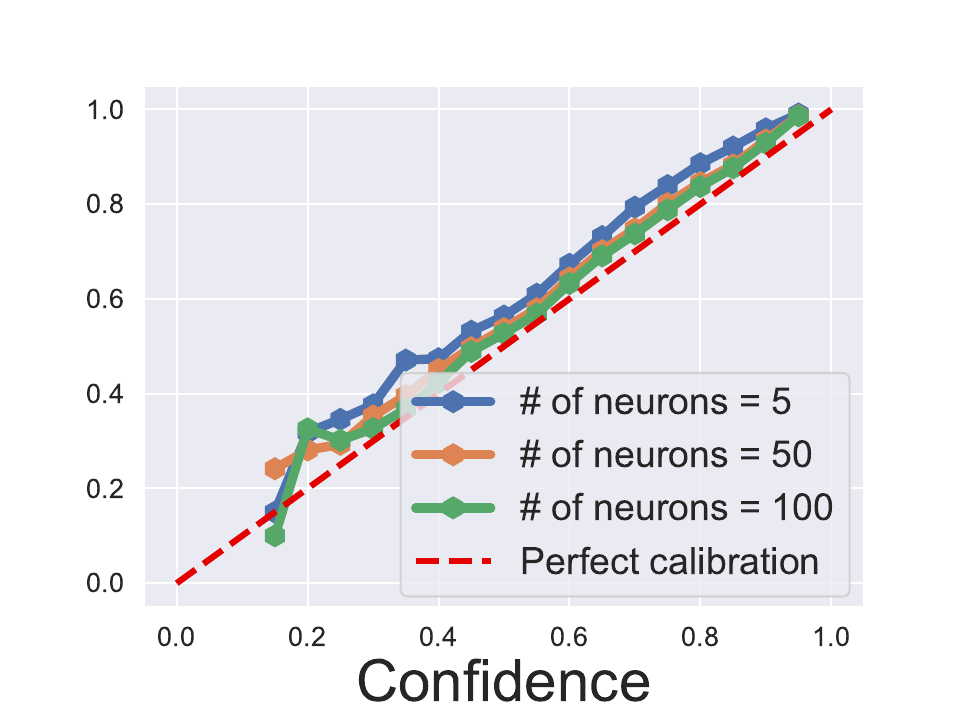}}}\\
    \BlankLine
    \vspace{-0.75 cm}
    \BlankLine
    \subfigure{{\includegraphics[height=0.85 in]{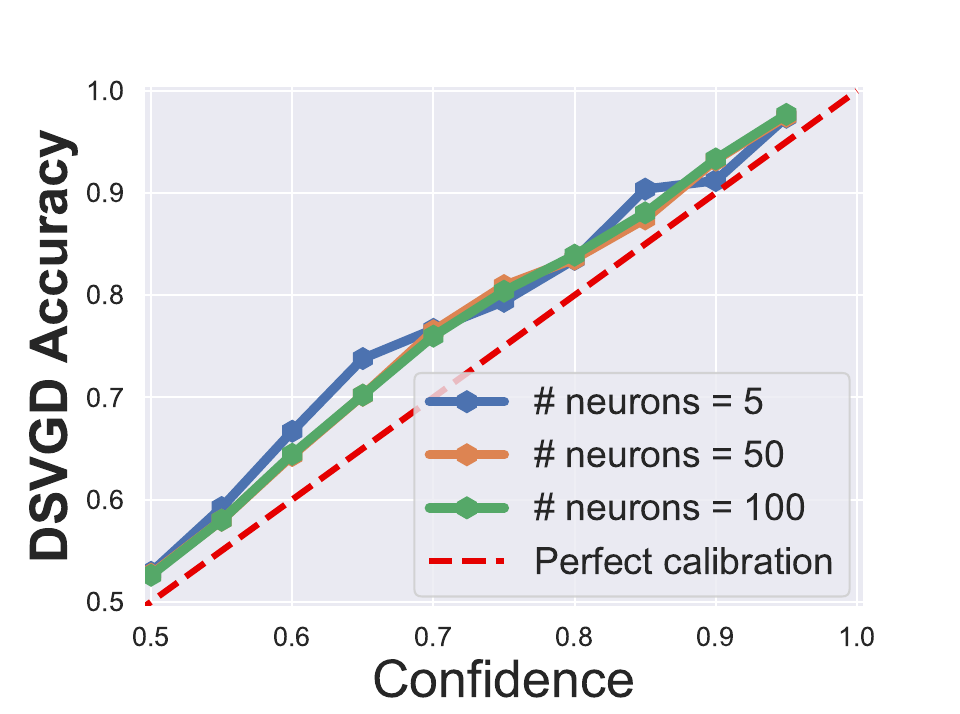}}}\hspace{-0.4cm}
    \subfigure{{\includegraphics[height=0.85 in]{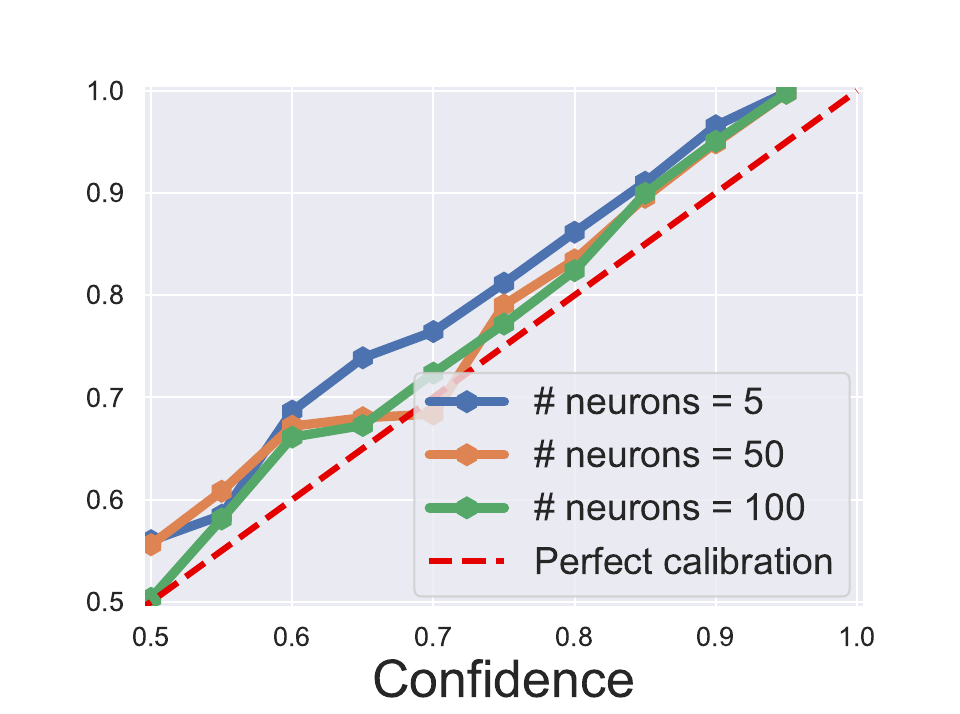}}} \hspace{-0.4cm}
    \subfigure{{\includegraphics[height=0.85 in]{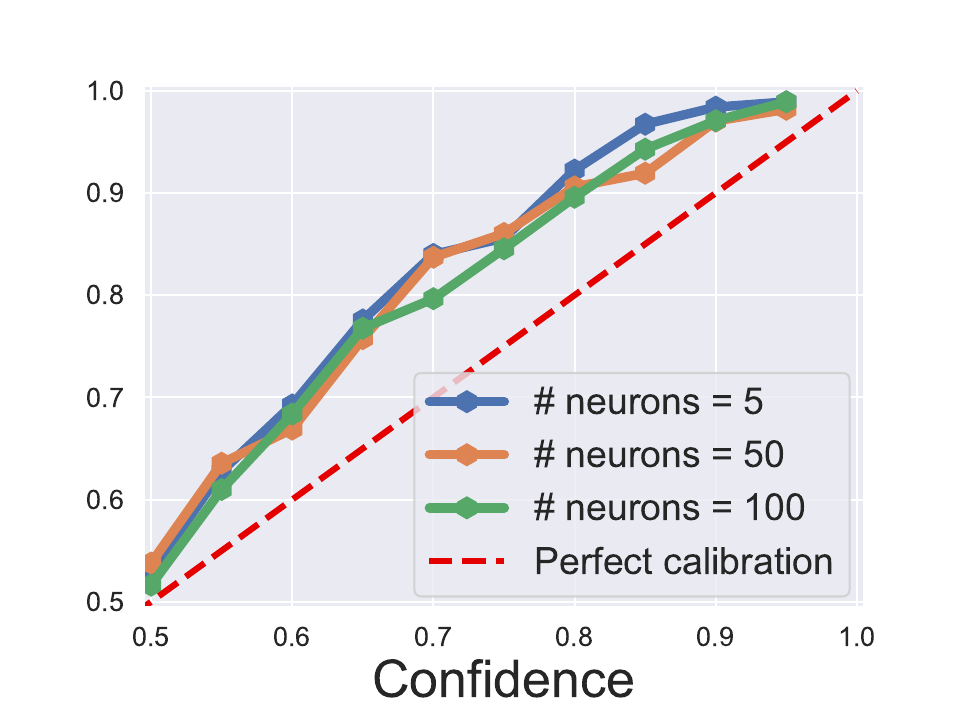}}} \hspace{-0.4cm}
    \subfigure{{\includegraphics[height=0.85 in]{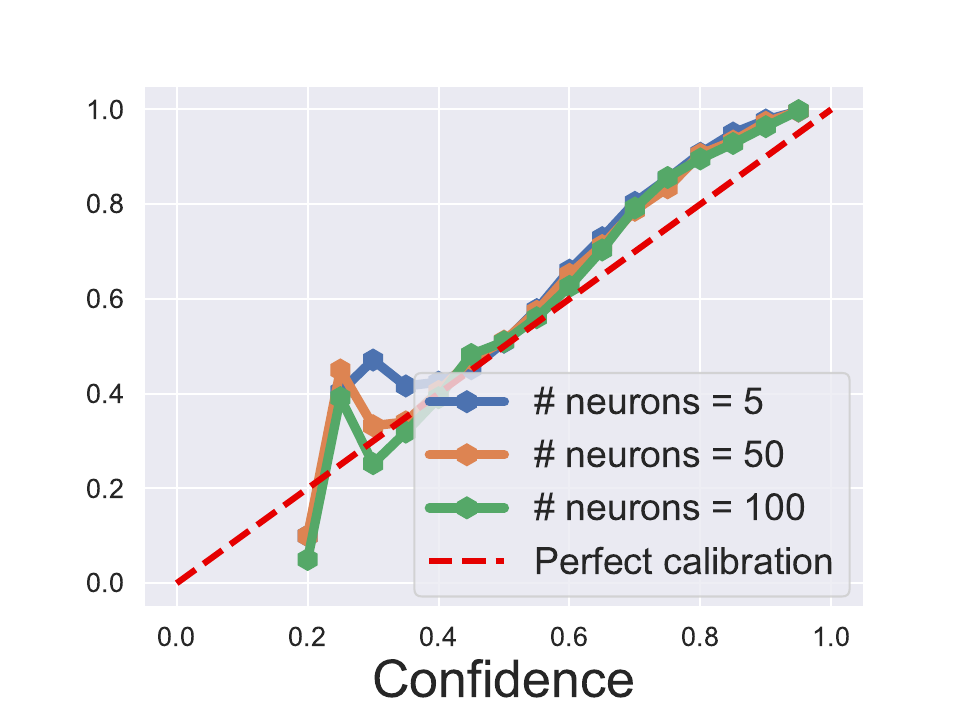}}}\hspace{-0.4cm}
    \subfigure{{\includegraphics[height=0.85 in]{Figures/rel_BNN_Fmnist_dsvgd.pdf}}}
    \caption{Reliability plots for classification using   Bayesian neural networks for a variable number of hidden neurons with \ac{FedAvg} (top row), \ac{SVGD} (middle row) and \ac{DSVGD} (bottom row). We use $N=20$ particles ($I=10$, $L=L^{\prime}=200$ and $K=20$ agents).}
    \label{fig:rel_BNN_app}
\end{figure}
\begin{figure}[h]
    \centering
    \subfigure{{\includegraphics[height=1.15 in]{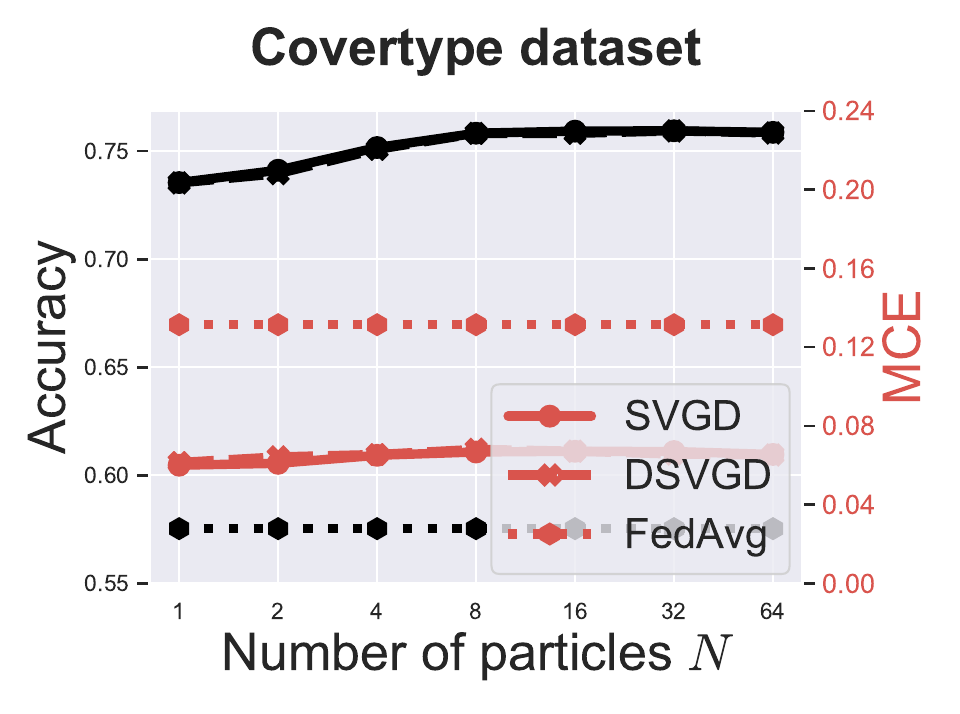}}}
    \subfigure{{\includegraphics[height=1.15 in]{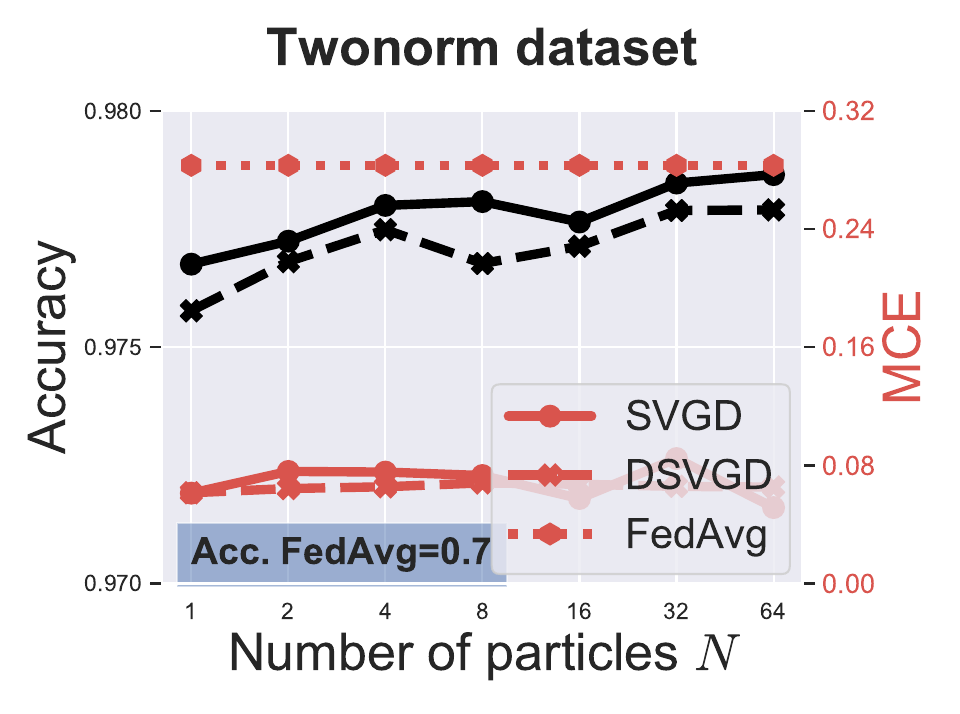}}} 
    \subfigure{{\includegraphics[height=1.15 in]{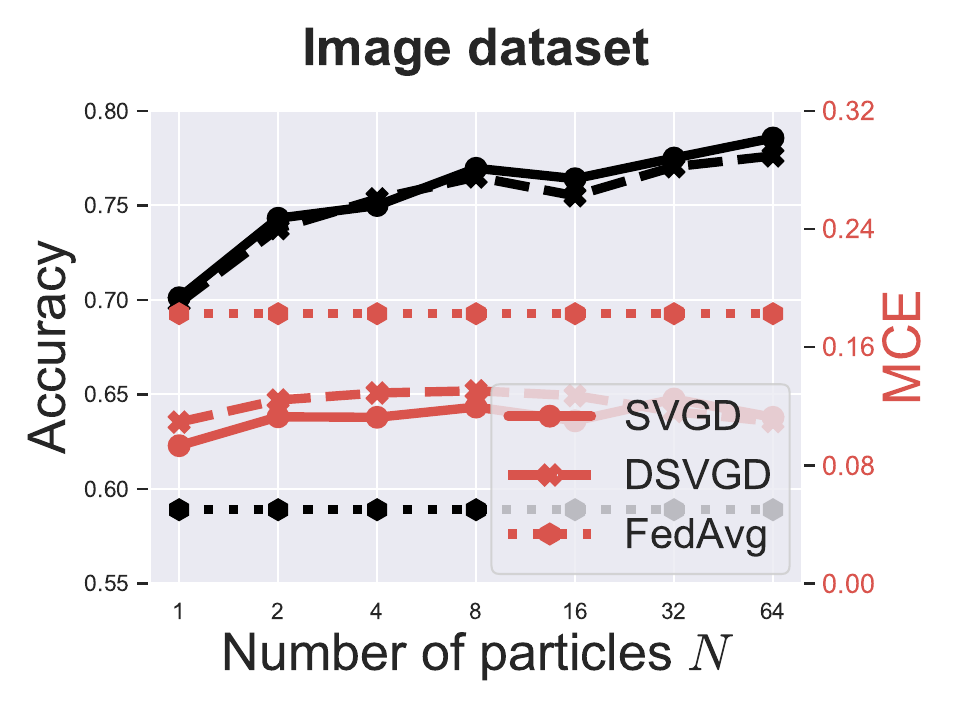}}} 
    \caption{Accuracy and Maximum Calibration Error (MCE) as function of the number of particles $N$ for   Bayesian neural networks. We fix $I=10$, $L=L^{\prime}=200$ and $K=20$ agents in both figures.}
    \label{fig:acc_mce_N}
\end{figure}
This section provides additional results on the calibration experiment conducted in Sec. \ref{sec:experiments} of the main text using additional datasets. In Fig. \ref{fig:rel_BNN_app}, we show the reliability plots for \ac{SVGD}, \ac{DSVGD} and \ac{FedAvg} with $K=20$ agents across various datasets and for different number of neurons in the hidden layer. We first note that \ac{DSVGD} retains the same calibration level as \ac{SVGD} across all datasets. 
Furthermore, while increasing the number of hidden neurons negatively affects \ac{FedAvg} due to overfitting, it does not affect the trustworthiness of the predictions for the Bayesian counterparts. This is a general property for Bayesian methods that contrast with frequentist approaches, for which increasing the number of parameters improves accuracy at the price of miscalibration \citep{NN_calibration}. 
\par
Fig. \ref{fig:acc_mce_N} plots the accuracy and \ac{MCE} as function of the number of particles $N$. While increasing $N$ improves the accuracy (as also shown in Fig. \ref{fig:accuracy_nb_particles_all}) for \ac{SVGD} and \ac{DSVGD}, the \ac{MCE} is unaffected and is lower than the \ac{MCE} value for \ac{FedAvg}.
\clearpage
\section{Implementation Details}
\label{app:implementation_details}
\subsection{Datasets, Benchmarks and Hyperparamters Details}
\label{app:dataset_detail}
\textbf{Datasets.} We summarize in Table \ref{table:datasets} the main parameters used across different datasets that are invariant across all experiments. The covertype dataset\footnote{\url{https://www.csie.ntu.edu.tw/~cjlin/libsvmtools/datasets/binary.html}} and the remaining binary classfication datasets that are selected from the Gunnar Raetsch's Benchmark datasets\footnote{\url{http://theoval.cmp.uea.ac.uk/matlab/default.html}} as compiled by \citet{mika_datasets} are used directly without normalization as in \citet{SVGD} except for the vehicle sensors dataset\footnote{\url{http://www.ecs.umass.edu/~mduarte/Software.html}} which is normalized by removing the mean of each feature and dividing by their standard deviations. Regression datasets\footnote{\url{https://archive.ics.uci.edu/ml/datasets.php}} are normalized by removing the mean of each feature and dividing by their standard deviations, and multi-label classification datasets\footnote{\url{http://yann.lecun.com/exdb/mnist/}}\footnote{\url{https://github.com/zalandoresearch/fashion-mnist}} are normalized by multiplying each pixel value by $0.99/255$ and adding $0.01$ such that every pixel value after normalization belongs to the interval $[0.01,1]$. All performance metrics used are averaged over the number of trials. In each trial, unless specified otherwise, we permute the datasets and randomly split them across different agents.\par
\textbf{Hyperparameters.} The hyperparameters used are summarized in Table \ref{table:hyperparams}. These apply for all schemes except for \ac{DSGLD} and \ac{SGLD}, where the learning rates are annealed and are respectively equal to $a_0 \cdot  (0.5 + i\cdot L + l)^{-0.55}$ and $a_0 \cdot  (0.5 + l)^{-0.55}$ to ensure that they go from the order of $0.01$ to $0.0001$ as advised by \citet{sgld}. $a_0$ is fixed according to the values in Table \ref{table:a_0}.\par 
\textbf{DSGLD implementation.} \ac{DSGLD} is implemented by splitting the $N$ particles among the $K$ agents. More specifically, when scheduled, each agent runs $\ceil{N/K}$ Markov chains. We assumed that the response delay in addition to the trajectory length of the chains \citep{dsgld} to be equal among all workers and unchanged throughout the learning process.\par
\textbf{FedAvg implementation.} \ac{FedAvg} is implemented as in \citet{comm_efficient_learning_dec_data} with the only difference that the server schedules a single agent at a time. Each scheduled agent performs $L$ SGD iterations to minimize its local loss.\par
\textbf{PVI and GVI implementation.} \ac{PVI} and \ac{GVI} are implemented using a Gaussian parametrization for both the posterior and the prior. The natural parameters are updated via the closed form update in \citet[Property $4$]{PVI}.\par
\textbf{Scheduling. }Unless specified otherwise, we use a round robin scheduler to schedule agents. However, any scheduler can be used as long as it schedules one agent per communication round. 
\begin{table}[h]
\caption{Overview of datasets and parameters used in the experiments. Datasets in bold are used in the experiments section of the main text.}
\centering
  \scalebox{0.7}{\begin{tabular}{c  c  c c  c  c}
    \\
    \toprule
    \multirow{1}{*}{Dataset Name}
       & {Size} & {Task} & {batchsize} & {\# trials} & {Train/test split}\\
      \hline
    \textbf{Covertype} & $581,012\times55$ & Binary classification & $100$ & $50$ & $80\%/20\%$\\
    \textbf{Twonorm} & $7,400\times20$ & Binary classification & $10$ & $50$ & $80\%/20\%$\\      
    Ringnorm & $7,400\times20$ & Binary classification & $10$ & $50$ & $80\%/20\%$\\
    Image & $2,086\times18$ & Binary classification & $10$ & $50$ & $80\%/20\%$\\
    Breast Cancer & $263\times9$ & Binary classification & $10$ & $50$ & $80\%/20\%$\\    Diabetis & $768\times8$ & Binary classification
         & $10$ & $50$ & $80\%/20\%$\\    
        German & $1,000\times20$ & Binary classification & $10$ & $50$ & $80\%/20\%$\\    
        Heart & $270\times13$ & Binary classification & $10$ & $50$ & $80\%/20\%$\\    
        Waveform & $5,086\times21$ & Binary classification & $10$ & $50$ & $80\%/20\%$\\
        Vehicle Sensors & $2010\times23$ & Binary classification & $10$ & $50$ & $80\%/20\%$\\
    \textbf{Kin8nm} & $8,192\times8$ & Regression & $100$ & $50$ & $90\%/10\%$\\
    Naval Propulsion & $11,934\times16$ & Regression & $100$ & $50$ & $90\%/10\%$\\
    Combined cycle power plant (CCPP) & $9,568\times4$ & Regression & $100$ & $50$ & $90\%/10\%$\\
    \textbf{Year Prediction} & $515,345\times90$ & Regression & $1000$ & $20$ & $90\%/10\%$\\

    \textbf{MNIST} & $60,000\times785$ & Multi-label classification & $100$ & $20$ & $86\% / 14\%$\\
    \textbf{Fashion MNIST} & $60,000\times785$ & Multi-label classification & $100$ & $20$ & $86\% / 14\%$\\
    \bottomrule
  \end{tabular}}
  \label{table:datasets}
\end{table}

\begin{table}[h]
\caption{Summary of hyperparameters used across various experiments.}
\centering
  \scalebox{0.7}{\begin{tabular}{c | c  c  c}
    \toprule
    \multirow{1}{*}{Hyperparameter} & {Regression}
       & {Binary Classification} & {Multi-label Classification}\\
      \hline
    Ada Learning rate\footnote{All learning rates for non-parametric particle-based benchmark schemes used are scaled by a factor of $1/N$ to match our learning rate and ensure fair comparison.} & $0.001$ & $0.05$ & $0.001$\\
    Ada smoothing term (or fudge factor) & $10^{-6}$ & $10^{-9}$ & $10^{-6}$\\
    Momentum & $0.9$ & $0.9$ & $0.9$\\
    KDE bandwidth & $0.55$ & $0.55$ & $0.55$\\
    \bottomrule
  \end{tabular}}
  \label{table:hyperparams}
\end{table}

\begin{table}[h]
\caption{Learning rate for \ac{DSGLD} and \ac{SGLD} used across various datasets.}
\centering
  \scalebox{0.7}{\begin{tabular}{c | c  c  c c}
    \toprule
    \multirow{1}{*}{Hyperparameter} & {Year}
       & {MNIST} & {F-MNIST} & {Other}\\
      \hline
    \ac{DSGLD} $a_0$ & $0.0005$ & $0.0005$ & $0.0005$ & $0.01$\\
    \ac{SGLD} $a_0$ & - & $0.001$ & $0.001$ & $0.01$\\
    \bottomrule
  \end{tabular}}
  \label{table:a_0}
\end{table}

\subsection{Software Details}
We implement all experiments in PyTorch \citep{pytorch} Version 10.3.1.
Our experiments and code are based on the original \ac{SVGD} experiments and code available at: \url{https://github.com/DartML/Stein-Variational-Gradient-Descent}. More specifically, \ac{DSVGD} can be easily obtained by running \ac{SVGD} twice at each scheduled agent and suitably adjusting its target distribution. Our code is attached with the supplementary materials.

\end{document}

%% file: math_commands.tex
\usepackage{amsmath,amsfonts,bm}

\def\eqref#1{equation~\ref{#1}}

\def\ceil#1{\lceil #1 \rceil}

\def\1{\bm{1}}

\DeclareMathAlphabet{\mathsfit}{\encodingdefault}{\sfdefault}{m}{sl}
\SetMathAlphabet{\mathsfit}{bold}{\encodingdefault}{\sfdefault}{bx}{n}

\DeclareMathOperator*{\argmax}{arg\,max}

%% file: acronymes.tex
\begin{acronym}

\acro{BNN}{Bayesian Neural Networks}
\acro{CDF}{cumulative distribution function}
\acro{DSVGD}{Distributed Stein Variational Gradient Descent}
\acro{DSGLD}{Distributed Stochastic Gradient Langevin Dynamics}
\acro{DVI}{Distributed Variational Inference}

\acro{ELBO}{Evidence Lower Bound}
\acro{FL}{Federated Learning}
\acro{FedAvg}{Federated Averaging}
\acro{FedSGD}{Federated Stochastic Gradient Descent}
\acro{GVI}{Global Variational Inference}

\acro{KL}{Kullback–Leibler}
\acro{KDE}{Kernel Density Estimator}
\acro{MCMC}{Markov Chain Monte Carlo}
\acro{MC}{Monte Carlo}
\acro{MCE}{Maximum Calibration Error}
\acro{NPV}{Non-Parametric Variational Inference}

\acro{PDF}{probability density function}%
\acro{PVI}{Partitioned Variational Inference}
\acro{PMD}{Particle Mirror Descent}
\acro{P-DSVGD}{Parallel-Distributed Stein Variational Gradient Descent}

\acro{RKHS}{Reproducing Kernel Hilbert Space}
\acro{SPSA}{simultaneous perturbation stochastic approximation}
\acro{RBF}{Radial Basis Function}
\acro{RMSE}{Root Mean Square Error}
\acro{SVGD}{Stein Variational Gradient Descent}
\acro{SGLD}{Stochastic Gradient Langevin Dynamics}
\acro{U-DSVGD}{Unconstrained-DSVGD}
\acro{VI}{Variational Inference}

\end{acronym}

%% file: my_theorems.tex
\subsection{Proofs}
\label{app:theoretical_analysis}
{\color{black}
In this section, we prove Theorem \ref{th:stationary_points} and \ref{th:global_energy_decrease}.
\stationarypoints*
\begin{proof}
Consider the general implementation of \ac{DVI}, were a set $\mathcal{K}$ of agents are scheduled in parallel. \ac{DVI} is equivalent to the following functional mapping
\begin{align}
    \begin{bmatrix}
          \prod_{i \not\in \mathcal{K}} t_{i}(\theta) \\ \\
            \{ t_k(\theta)\}_{k \in \mathcal{K}}
         \end{bmatrix} 
        &\xrightarrow{}
        \begin{bmatrix}
          \prod_{i \not\in \mathcal{K}} t_{i}(\theta) \\ \\
            \Big\{ t_{k}^{\prime}(\theta) = \frac{1}{Z} \exp\Big(-\frac{1}{\alpha} L_k (\theta)\Big)\Big\}_{k \in \mathcal{K}}
         \end{bmatrix} \nonumber\\
             \Big(q(\theta) = p_0(\theta) \prod_{i=1}^{K} t_i(\theta)\Big)&\ \ \ \ \ \ \ \Big(q^\prime(\theta) = p_0(\theta)
    \prod_{i \not\in \mathcal{K}} t_{i}(\theta)
    \prod_{k \in \mathcal{K}} t_{k}^{\prime}(\theta)\Big) 
  \nonumber\end{align}
where $Z = \int p_0(\theta) 
    \prod_{i \not\in \mathcal{K}} t_{i}(\theta)
    \prod_{k \in \mathcal{K}} t_{k}^{\prime}(\theta) d\theta$.\\
Therefore, assuming that all devices $k$ are periodically scheduled, $q(\theta)$ is a fixed point of \ac{DVI} if and only if the following equality holds
\begin{align}
    t_k(\theta) = t_{k}^{\prime} (\theta)\ \text{for}\ k=1,\ldots, K. \nonumber
\end{align}
This condition is satisfied by $q(\theta) = q_{opt}(\theta)$ and by no other distribution. 
This concludes the proof.
\end{proof}}
We move now to Theorem \ref{th:global_energy_decrease} for \ac{U-DSVGD}. We leave the analysis of the impact of the additional distillation step used by \ac{DSVGD} for future work. The analysis builds on the following result from \citet{korba_svgd_nonasymptotic}, which is restated here using our notation.\par
Denote by $||\cdot||_{\mathcal{H}}$ the norm in the \ac{RKHS} $\mathcal{H}$ defined by the positive definite kernel $\mathrm{k}(\theta, \theta^{\prime})$. We assume that the kernel satisfies the following technical condition: there exist a constant $B>0$ such that
\begin{equation}
    ||\mathrm{k}(\theta, \cdot) ||_{\mathcal{H}} \leq B\ \mathrm{and}\ \sum_{j=1}^{d} \Big|\Big|\frac{\partial\mathrm{k}(\theta, \cdot)}{\partial \theta_j} \Big|\Big|_{\mathcal{H}}^2 \leq B^2. \label{th:kernel_cond}
\end{equation}
This condition is for instance satisfied by the \ac{RBF} kernel with $B=1$ \citep{derivative_RKHS}. Furthermore, we define the kernelized Stein discrepancy \citep{kernelized_stein_discrepancy} between two distributions $p$ and $q$ as $S(p, q)$, and the total variation distance as $||q-p||_{TV} = \frac{1}{2}\int |q(\theta)-p(\theta)|d\theta$.
\par

\textbf{Lemma 1.} \textit{(Guaranteed per-iteration decrease of the local free energy.)} \citep{korba_svgd_nonasymptotic} For a kernel satisfying (\ref{th:kernel_cond}), assume that, at a given communication round $i$ and local iteration $l$, with agent $k$ scheduled, we have:
\begin{itemize}
    \item the maximum absolute eigenvalue of the Hessian $-\nabla^2\log \tilde{p}_{k}^{(i)}(\theta)$ is upper bounded by a constant $M > 0$; and
    \item the inequality $S(q^{[l]}(\theta), \tilde{p}_{k}^{(i)}) < C$ holds for some $C>0$.
\end{itemize}
For learning rate $\epsilon \leq (\beta-1)/(\beta BC^{\frac{1}{2}})$ with any $\beta > 1$, the decrease in the local KL divergence from local iteration $l$ to $l+1$ satisfies the inequality
\begin{equation}
    F(q^{[l+1]}(\theta)) -  F(q^{[l]}(\theta)) \leq - \alpha\epsilon S(q^{[l]}, \tilde{p}_{k}^{(i)})(1- \epsilon  \gamma ), \label{eq:kl_dissipation}
\end{equation}
where $\gamma=((\beta^2+M)B^2)/2$. \par
Lemma 1 shows that by choosing a learning rate $\epsilon \leq \min(\gamma^{-1}, (\beta-1)/(\beta BC^{\frac{1}{2}})$, one can guarantee a per-iteration decrease in the local-free energy, i.e., in the KL divergence between the particles' distribution and the target tilted distribution $\tilde{p}_{k}^{(i)}(\theta)$ that depends on the kernelized Stein discrepancy $S(q^{[l]}, \tilde{p}_{k}^{(i)})$ at the iteration before the update.
\par
\textbf{Lemma 2.} \textit{(Relationship between global and local free energy.)} The global free energy $F(q(\theta))$ in (\ref{eq:main_minimization}) is related to the local free energy $F_{k}^{(i)}(q(\theta))$ in (\ref{eq:min_local_energy_cavity}) of the $k$-th scheduled agent as
\begin{equation}  
   F(q(\theta)) =  F_{k}^{(i)}(q(\theta))  + \alpha \sum_{m \neq k}\mathbb{E}_{q(\theta)} \log \bigg( \frac{t_{m}^{(i-1)}(\theta)}{\exp(-\frac{1}{\alpha} L_m (\theta))}\bigg). \label{th:lemma2}
\end{equation}
\\
\textit{Proof.} The global free energy (\ref{eq:main_minimization}) can be written as
\begin{equation}
\begin{aligned}
    F(q(\theta)) &= \alpha \mathbb{E}_{q(\theta)}\log\bigg( \frac{q(\theta)}{p_{0}(\theta) \exp(-\frac{1}{\alpha}\sum_{m=1}^{K} L_m (\theta))} \bigg)\\
    &= \alpha \mathbb{E}_{q(\theta)}\log\bigg( 
    \frac{q(\theta)}{\Tilde{p}_{k}^{(i)}(\theta)} \cdot \frac{\frac{q^{(i-1)}(\theta)}{ t_{k}^{(i-1)}(\theta)}}{p_0 (\theta) \exp(-\frac{1}{\alpha}\sum_{m \neq k}L_m (\theta))}\bigg) \\
    & = \alpha \mathbb{E}_{q(\theta)} \log \bigg( \frac{q(\theta)}{\Tilde{p}_{k}^{(i)}(\theta)} \bigg)  + \alpha \mathbb{E}_{q(\theta)} \log \bigg( \frac{p_0(\theta) \prod_{m \neq k} t_{m}^{(i-1)}(\theta)}{p_0 (\theta) \exp (-\frac{1}{\alpha}\sum_{m \neq k} L_m (\theta))}\bigg) \\
    & = F_{k}^{(i)}(q(\theta)) + \alpha \sum_{m \neq k}\mathbb{E}_{q(\theta)} \log \bigg( \frac{t_{m}^{(i-1)}(\theta)}{\exp(-\frac{1}{\alpha} L_m (\theta))}\bigg),  \label{eq:F_q}
    \end{aligned}
\end{equation}
where in the second equality we have used (\ref{eq:tilted_udsvgd}); and in the third equality we have used the equality $q^{(i-1)}(\theta) = p_{0}(\theta)\prod_{m=1}^{K}t_{m}^{(i-1)} (\theta)$, which is guaranteed by the \ac{U-DSVGD} update (\ref{eq:t_part_udsvgd}) and (\ref{eq:tilted_udsvgd}) (see \citet[Property 2]{PVI}).\qed
\par
\globalenergydecrease*

We know from Lemma 1 that a learning rate $\epsilon \leq \min(\gamma^{-1}, (\beta-1)/(\beta BC^{\frac{1}{2}})$ is sufficient to ensure a per-iteration decrease in the \textit{local} free energy. Given that the KL divergence in the second term in (\ref{eq:bound_energy}) generally increases with $\epsilon$, \ref{th:global_energy_decrease} demonstrates that, in order to guarantee a reduction of the \textit{global} free energy, a smaller learning rate may be required. We also note that the KL divergence term $\mathbb{D}(q^{[l+1]}||q^{[l]})$ may be explicitly related to the learning rate by following \citet[Sec. 8]{SVGP}, but we do not further pursue this aspect here. We finally remark that, in the presence of $K=1$ agent, the upper bound (\ref{eq:kl_dissipation}) in \citep{korba_svgd_nonasymptotic} is recovered. This is because, in the presence of one agent, the global free energy reduces to the local free energy (see (\ref{th:lemma2})) and accordingly \ac{U-DSVGD} reduces to \ac{SVGD}. 
\par
\textit{Proof. } We wish to obtain an upper bound on the decrease of the global free energy $F(q^{[l+1]}(\theta)) - F(q^{[l]}(\theta))$ across each local \ac{SVGD} iteration during communication round $i$. Using (\ref{th:lemma2}), the decrease in the global free energy can be written as 
\begin{equation}
\begin{aligned}
    F(q^{[l+1]}(\theta)) - & F(q^{[l]}(\theta))= \underset{(a)}{\underbrace{F_{k}^{(i)}(q^{[l+1]}(\theta)) - F_{k}^{(i)}(q^{[l]}(\theta))}} \\ +  & \alpha \sum_{m \neq k}\bigg[\underset{(b)}{\underbrace{\mathbb{E}_{q^{[l+1]}(\theta)} \log \bigg( \frac{t_{m}^{(i-1)}(\theta)}{\exp(-\frac{1}{\alpha} L_m (\theta))}\bigg) - \mathbb{E}_{q^{[l]}(\theta)} \log \bigg( \frac{t_{m}^{(i-1)}(\theta)}{\exp(-\frac{1}{\alpha} L_m (\theta))}\bigg)}}\bigg]. \label{eq:Fab}
\end{aligned}
\end{equation}
We now derive upper bounds for $(a)$ and $(b)$. Using Lemma 1 and the definition of the local free energy in (\ref{eq:min_local_energy_cavity}), we have the following upper bound on $(a)$ 
\begin{equation}
    (a)=F_{k}^{(i)}(q^{[l+1]}(\theta)) - F_{k}^{(i)}(q^{[l]}(\theta)) \leq  - \alpha \epsilon S(q^{[l]}(\theta), \tilde{p}_{k}^{(i)})(1- \epsilon  \gamma ),
\end{equation}
while $(b)$ can be rewritten and upper bounded by using the properties of the total variation distance as
\begin{equation}
(b) = \int (q^{[l+1]}(\theta) - q^{[l]}(\theta)) \log \bigg( \frac{t_{m}^{(i-1)}(\theta)}{\exp(-\frac{1}{\alpha} L_m (\theta))}\bigg) d\theta \leq 2 l_{\mathrm{max}}^{(i)} ||q^{[l+1]}-q^{[l]} ||_{TV}.
\end{equation}
Using Pinsker's inequality \citep{pinsker}, the term $(b)$ can be further upper bounded as 
\begin{equation}
    (b) \leq 2 l_{\mathrm{max}}^{(i)} \sqrt{2 \mathbb{D}(q^{[l+1]}||q^{[l]})}.
\end{equation}
Accordingly, the global energy dissipation in (\ref{eq:Fab}) can be upper bounded as in (\ref{eq:bound_energy}). \qed

%% file: dsvgd.bbl
\begin{thebibliography}{56}
\providecommand{\natexlab}[1]{#1}
\providecommand{\url}[1]{\texttt{#1}}
\expandafter\ifx\csname urlstyle\endcsname\relax
  \providecommand{\doi}[1]{doi: #1}\else
  \providecommand{\doi}{doi: \begingroup \urlstyle{rm}\Url}\fi

\bibitem[Ahn et~al.(2014)Ahn, Shahbaba, and Welling]{dsgld}
Sungjin Ahn, Babak Shahbaba, and Max Welling.
\newblock {Distributed Stochastic Gradient MCMC}.
\newblock In Eric~P. Xing and Tony Jebara (eds.), \emph{Proceedings of the 31st
  International Conference on Machine Learning}, volume~32 of \emph{Proceedings
  of Machine Learning Research}, pp.\  1044--1052, Bejing, China, 22--24 Jun
  2014. PMLR.
\newblock URL \url{http://proceedings.mlr.press/v32/ahn14.html}.

\bibitem[Alquier et~al.(2016)Alquier, Ridgway, and Chopin]{properties_gibbs}
Pierre Alquier, James Ridgway, and Nicolas Chopin.
\newblock {On the properties of variational approximations of Gibbs
  posteriors}.
\newblock \emph{Journal of Machine Learning Research}, 17\penalty0
  (236):\penalty0 1--41, 2016.
\newblock URL \url{http://jmlr.org/papers/v17/15-290.html}.

\bibitem[Amari(1998)]{natural_gradient_amari}
Shun-Ichi Amari.
\newblock {Natural gradient works efficiently in learning}.
\newblock \emph{Neural computation}, 10\penalty0 (2):\penalty0 251--276, 1998.

\bibitem[Angelino et~al.(2016)Angelino, Johnson, and
  Adams]{angelino_patterns_scalable_BI}
Elaine Angelino, Matthew~James Johnson, and Ryan~P Adams.
\newblock {Patterns of scalable Bayesian inference}.
\newblock \emph{arXiv preprint arXiv:1602.05221}, 2016.

\bibitem[Bishop(2006)]{bishop_book_pattern_recognition}
Christopher~M. Bishop.
\newblock \emph{{Pattern Recognition and Machine Learning (Information Science
  and Statistics)}}.
\newblock Springer-Verlag, Berlin, Heidelberg, 2006.
\newblock ISBN 0387310738.

\bibitem[Broderick et~al.(2013)Broderick, Boyd, Wibisono, Wilson, and
  Jordan]{streaming_variational_bayes_tamara}
Tamara Broderick, Nicholas Boyd, Andre Wibisono, Ashia~C Wilson, and Michael~I
  Jordan.
\newblock {Streaming Variational Bayes}.
\newblock In C.~J.~C. Burges, L.~Bottou, M.~Welling, Z.~Ghahramani, and K.~Q.
  Weinberger (eds.), \emph{Advances in Neural Information Processing Systems
  26}, pp.\  1727--1735. Curran Associates, Inc., 2013.
\newblock URL
  \url{http://papers.nips.cc/paper/4980-streaming-variational-bayes.pdf}.

\bibitem[Bui et~al.(2018)Bui, Nguyen, Swaroop, and Turner]{PVI}
Thang~D Bui, Cuong~V Nguyen, Siddharth Swaroop, and Richard~E Turner.
\newblock {Partitioned Variational Inference: A unified framework encompassing
  federated and continual learning}.
\newblock \emph{arXiv preprint arXiv:1811.11206}, 2018.

\bibitem[Chen \& Chao(2021)Chen and Chao]{feddistill}
Hong-You Chen and Wei-Lun Chao.
\newblock Fed{\{}be{\}}: Making bayesian model ensemble applicable to federated
  learning.
\newblock In \emph{International Conference on Learning Representations}, 2021.
\newblock URL \url{https://openreview.net/forum?id=dgtpE6gKjHn}.

\bibitem[Claici et~al.(2020)Claici, Yurochkin, Ghosh, and
  Solomon]{model_fusion_KL}
Sebastian Claici, Mikhail Yurochkin, Soumya Ghosh, and Justin Solomon.
\newblock {Model Fusion with Kullback--Leibler Divergence}.
\newblock \emph{arXiv preprint arXiv:2007.06168}, 2020.

\bibitem[Corinzia \& Buhmann(2019{\natexlab{a}})Corinzia and
  Buhmann]{variational_federated_MTL}
Luca Corinzia and Joachim~M Buhmann.
\newblock {Variational federated multi-task learning}.
\newblock \emph{arXiv preprint arXiv:1906.06268}, 2019{\natexlab{a}}.

\bibitem[Corinzia \& Buhmann(2019{\natexlab{b}})Corinzia and
  Buhmann]{variational_multi_task}
Luca Corinzia and Joachim~M Buhmann.
\newblock {Variational federated multi-task learning}.
\newblock \emph{arXiv preprint arXiv:1906.06268}, 2019{\natexlab{b}}.

\bibitem[Dai et~al.(2016)Dai, He, Dai, and Song]{PMD_bo_dai}
Bo~Dai, Niao He, Hanjun Dai, and Le~Song.
\newblock {Provable Bayesian Inference via Particle Mirror Descent}.
\newblock In Arthur Gretton and Christian~C. Robert (eds.), \emph{Proceedings
  of the 19th International Conference on Artificial Intelligence and
  Statistics}, volume~51 of \emph{Proceedings of Machine Learning Research},
  pp.\  985--994, Cadiz, Spain, 09--11 May 2016. PMLR.
\newblock URL \url{http://proceedings.mlr.press/v51/dai16.html}.

\bibitem[DeGroot \& Fienberg(1983)DeGroot and Fienberg]{rel_diagram_degroot}
Morris~H. DeGroot and Stephen~E. Fienberg.
\newblock {The Comparison and Evaluation of Forecasters}.
\newblock \emph{Journal of the Royal Statistical Society. Series D (The
  Statistician)}, 32\penalty0 (1/2):\penalty0 12--22, 1983.
\newblock ISSN 00390526, 14679884.
\newblock URL \url{http://www.jstor.org/stable/2987588}.

\bibitem[Gershman et~al.(2012)Gershman, Hoffman, and Blei]{NPV}
Samuel~J. Gershman, Matthew~D. Hoffman, and David~M. Blei.
\newblock {Nonparametric Variational Inference}.
\newblock In \emph{Proceedings of the 29th International Coference on
  International Conference on Machine Learning}, ICML’12, pp.\  235–242,
  Madison, WI, USA, 2012. Omnipress.
\newblock ISBN 9781450312851.

\bibitem[Guo et~al.(2017)Guo, Pleiss, Sun, and Weinberger]{NN_calibration}
Chuan Guo, Geoff Pleiss, Yu~Sun, and Kilian~Q. Weinberger.
\newblock {On Calibration of Modern Neural Networks}.
\newblock In \emph{Proceedings of the 34th International Conference on Machine
  Learning - Volume 70}, ICML’17, pp.\  1321–1330. JMLR.org, 2017.

\bibitem[Hern\'{a}ndez-Lobato \& Adams(2015)Hern\'{a}ndez-Lobato and
  Adams]{PBP_lobato}
Jos\'{e}~Miguel Hern\'{a}ndez-Lobato and Ryan~P. Adams.
\newblock Probabilistic backpropagation for scalable learning of bayesian
  neural networks.
\newblock In \emph{Proceedings of the 32nd International Conference on
  International Conference on Machine Learning - Volume 37}, ICML'15, pp.\
  1861–1869. JMLR.org, 2015.

\bibitem[Hinton et~al.(2015)Hinton, Vinyals, and
  Dean]{hinton_distilling_knowledge_NN}
Geoffrey Hinton, Oriol Vinyals, and Jeff Dean.
\newblock {Distilling the Knowledge in a Neural Network}.
\newblock \emph{arXiv preprint arXiv:1503.02531}, 2015.

\bibitem[Jordan et~al.(2019)Jordan, Lee, and
  Yang]{comm_efficient_statistical_inference}
Michael~I. Jordan, Jason~D. Lee, and Yun Yang.
\newblock {Communication-Efficient Distributed Statistical Inference}.
\newblock \emph{Journal of the American Statistical Association}, 114\penalty0
  (526):\penalty0 668--681, 2019.
\newblock \doi{10.1080/01621459.2018.1429274}.
\newblock URL \url{https://doi.org/10.1080/01621459.2018.1429274}.

\bibitem[Jospin et~al.(2020)Jospin, Buntine, Boussaid, Laga, and
  Bennamoun]{hands_on_BNN_tutorial}
Laurent~Valentin Jospin, Wray Buntine, Farid Boussaid, Hamid Laga, and Mohammed
  Bennamoun.
\newblock {Hands-on Bayesian Neural Networks--a Tutorial for Deep Learning
  Users}.
\newblock \emph{arXiv preprint arXiv:2007.06823}, 2020.

\bibitem[Kairouz et~al.(2019)Kairouz, McMahan, Avent, Bellet, Bennis, Bhagoji,
  Bonawitz, Charles, Cormode, Cummings,
  et~al.]{advances_openproblems_FL_kairouz}
Peter Kairouz, H~Brendan McMahan, Brendan Avent, Aur{\'e}lien Bellet, Mehdi
  Bennis, Arjun~Nitin Bhagoji, Keith Bonawitz, Zachary Charles, Graham Cormode,
  Rachel Cummings, et~al.
\newblock {Advances and open problems in federated learning}.
\newblock \emph{arXiv preprint arXiv:1912.04977}, 2019.

\bibitem[Knoblauch et~al.(2019)Knoblauch, Jewson, and Damoulas]{generalized_VI}
Jeremias Knoblauch, Jack Jewson, and Theodoros Damoulas.
\newblock {Generalized variational inference}.
\newblock \emph{stat}, 1050:\penalty0 21, 2019.

\bibitem[Korba et~al.(2020)Korba, Salim, Arbel, Luise, and
  Gretton]{korba_svgd_nonasymptotic}
Anna Korba, Adil Salim, Michael Arbel, Giulia Luise, and Arthur Gretton.
\newblock {A Non-Asymptotic Analysis for Stein Variational Gradient Descent}.
\newblock \emph{arXiv preprint arXiv:2006.09797}, 2020.

\bibitem[Li et~al.(2018)Li, Sahu, Zaheer, Sanjabi, Talwalkar, and
  Smith]{fedprox}
Tian Li, Anit~Kumar Sahu, Manzil Zaheer, Maziar Sanjabi, Ameet Talwalkar, and
  Virginia Smith.
\newblock {Federated Optimization in Heterogeneous Networks}.
\newblock \emph{arXiv preprint arXiv:1812.06127}, 2018.

\bibitem[Lin et~al.(2020)Lin, Brinton, and Michelusi]{FL_comm_delay}
Frank Po-Chen Lin, Christopher~G Brinton, and Nicol{\`o} Michelusi.
\newblock {Federated Learning with Communication Delay in Edge Networks}.
\newblock \emph{arXiv preprint arXiv:2008.09323}, 2020.

\bibitem[Liu(2017{\natexlab{a}})]{SVGD_proof}
Qiang Liu.
\newblock {Stein Variational Gradient Descent as Gradient Flow}.
\newblock In I.~Guyon, U.~V. Luxburg, S.~Bengio, H.~Wallach, R.~Fergus,
  S.~Vishwanathan, and R.~Garnett (eds.), \emph{Advances in Neural Information
  Processing Systems}, pp.\  3115--3123. Curran Associates, Inc.,
  2017{\natexlab{a}}.

\bibitem[Liu(2017{\natexlab{b}})]{svgd_gradient_flow}
Qiang Liu.
\newblock {Stein Variational Gradient Descent as Gradient Flow}.
\newblock In \emph{Advances in Neural Information Processing Systems}, pp.\
  3115--3123. 2017{\natexlab{b}}.

\bibitem[Liu \& Ihler(2014)Liu and Ihler]{distributed_estimation_expo}
Qiang Liu and Alexander Ihler.
\newblock {Distributed Estimation, Information Loss and Exponential Families}.
\newblock In \emph{Proceedings of the 27th International Conference on Neural
  Information Processing Systems - Volume 1}, NIPS'14, pp.\  1098–1106,
  Cambridge, MA, USA, 2014. MIT Press.

\bibitem[Liu \& Wang(2016)Liu and Wang]{SVGD}
Qiang Liu and Dilin Wang.
\newblock {Stein variational gradient descent: A general purpose bayesian
  inference algorithm}.
\newblock In \emph{Advances in neural information processing systems}, pp.\
  2378--2386, 2016.

\bibitem[Liu et~al.(2016)Liu, Lee, and Jordan]{kernelized_stein_discrepancy}
Qiang Liu, Jason Lee, and Michael Jordan.
\newblock {A kernelized Stein discrepancy for goodness-of-fit tests}.
\newblock In \emph{International conference on machine learning}, pp.\
  276--284, 2016.

\bibitem[MacKay(2002)]{david_mackay_book}
David J.~C. MacKay.
\newblock \emph{{Information Theory, Inference \& Learning Algorithms}}.
\newblock Cambridge University Press, USA, 2002.
\newblock ISBN 0521642981.

\bibitem[McMahan et~al.(2017)McMahan, Moore, Ramage, Hampson, and
  y~Arcas]{comm_efficient_learning_dec_data}
Brendan McMahan, Eider Moore, Daniel Ramage, Seth Hampson, and
  Blaise~Ag{\"{u}}era y~Arcas.
\newblock {Communication-Efficient Learning of Deep Networks from Decentralized
  Data}.
\newblock In \emph{Proceedings of the 20th International Conference on
  Artificial Intelligence and Statistics, {AISTATS} 2017, 20-22 April 2017,
  Fort Lauderdale, FL, {USA}}, volume~54 of \emph{Proceedings of Machine
  Learning Research}, pp.\  1273--1282. {PMLR}, 2017.
\newblock URL \url{http://proceedings.mlr.press/v54/mcmahan17a.html}.

\bibitem[Mesquita et~al.(2020)Mesquita, Blomstedt, and
  Kaski]{embarassingly_parallel_MCMC}
Diego Mesquita, Paul Blomstedt, and Samuel Kaski.
\newblock {Embarrassingly Parallel MCMC using Deep Invertible Transformations}.
\newblock volume 115 of \emph{Proceedings of Machine Learning Research}, pp.\
  1244--1252, Tel Aviv, Israel, 22--25 Jul 2020. PMLR.
\newblock URL \url{http://proceedings.mlr.press/v115/mesquita20a.html}.

\bibitem[{Mika} et~al.(1999){Mika}, {Ratsch}, {Weston}, {Scholkopf}, and
  {Mullers}]{mika_datasets}
S.~{Mika}, G.~{Ratsch}, J.~{Weston}, B.~{Scholkopf}, and K.~R. {Mullers}.
\newblock {Fisher discriminant analysis with kernels}.
\newblock In \emph{Neural Networks for Signal Processing IX: Proceedings of the
  1999 IEEE Signal Processing Society Workshop (Cat. No.98TH8468)}, pp.\
  41--48, 1999.

\bibitem[Mitros \& Mac~Namee(2019)Mitros and
  Mac~Namee]{uncertainty_validity_BNN}
John Mitros and Brian Mac~Namee.
\newblock {On the Validity of Bayesian Neural Networks for Uncertainty
  Estimation}.
\newblock \emph{arXiv preprint arXiv:1912.01530}, 2019.

\bibitem[Neal(2012)]{neal_bayesian_nn}
Radford~M Neal.
\newblock \emph{{Bayesian learning for neural networks}}, volume 118.
\newblock Springer Science \& Business Media, 2012.

\bibitem[Neiswanger et~al.(2015)Neiswanger, Wang, and Xing]{EPVI}
Willie Neiswanger, Chong Wang, and Eric Xing.
\newblock {Embarrassingly parallel variational inference in nonconjugate
  models}.
\newblock \emph{arXiv preprint arXiv:1510.04163}, 2015.

\bibitem[Nguyen et~al.(2020)Nguyen, Sehwag, Hosseinalipour, Brinton, Chiang,
  and Poor]{fast_convergent_fl}
Hung~T Nguyen, Vikash Sehwag, Seyyedali Hosseinalipour, Christopher~G Brinton,
  Mung Chiang, and H~Vincent Poor.
\newblock {Fast-Convergent Federated Learning}.
\newblock \emph{arXiv preprint arXiv:2007.13137}, 2020.

\bibitem[Niculescu-Mizil \& Caruana(2005)Niculescu-Mizil and
  Caruana]{rel_diagrams_niculescu}
Alexandru Niculescu-Mizil and Rich Caruana.
\newblock {Predicting Good Probabilities with Supervised Learning}.
\newblock In \emph{Proceedings of the 22nd International Conference on Machine
  Learning}, ICML ’05, pp.\  625–632, New York, NY, USA, 2005. Association
  for Computing Machinery.
\newblock ISBN 1595931805.
\newblock \doi{10.1145/1102351.1102430}.
\newblock URL \url{https://doi.org/10.1145/1102351.1102430}.

\bibitem[Paszke et~al.(2019)Paszke, Gross, Massa, Lerer, Bradbury, Chanan,
  Killeen, Lin, Gimelshein, Antiga, Desmaison, Kopf, Yang, DeVito, Raison,
  Tejani, Chilamkurthy, Steiner, Fang, Bai, and Chintala]{pytorch}
Adam Paszke, Sam Gross, Francisco Massa, Adam Lerer, James Bradbury, Gregory
  Chanan, Trevor Killeen, Zeming Lin, Natalia Gimelshein, Luca Antiga, Alban
  Desmaison, Andreas Kopf, Edward Yang, Zachary DeVito, Martin Raison, Alykhan
  Tejani, Sasank Chilamkurthy, Benoit Steiner, Lu~Fang, Junjie Bai, and Soumith
  Chintala.
\newblock {PyTorch: An Imperative Style, High-Performance Deep Learning
  Library}.
\newblock In H.~Wallach, H.~Larochelle, A.~Beygelzimer, F.~d~Alch\'{e}-Buc,
  E.~Fox, and R.~Garnett (eds.), \emph{Advances in Neural Information
  Processing Systems 32}, pp.\  8026--8037. Curran Associates, Inc., 2019.

\bibitem[Pathak \& Wainwright(2020)Pathak and Wainwright]{fedsplit}
Reese Pathak and Martin~J Wainwright.
\newblock {FedSplit: An algorithmic framework for fast federated optimization}.
\newblock \emph{arXiv preprint arXiv:2005.05238}, 2020.

\bibitem[Pinder et~al.(2020)Pinder, Nemeth, and Leslier]{SVGP}
Thomas Pinder, Christopher Nemeth, and David Leslier.
\newblock {Stein Variational Gaussian Processes}.
\newblock \emph{arXiv preprint arXiv:2009.12141}, 2020.

\bibitem[Pinsker(1964)]{pinsker}
Mark~S Pinsker.
\newblock \emph{{Information and information stability of random variables and
  processes}}.
\newblock Holden-Day, 1964.

\bibitem[Rendell et~al.(2018)Rendell, Johansen, Lee, and
  Whiteley]{global_consensus_MC}
Lewis~J Rendell, Adam~M Johansen, Anthony Lee, and Nick Whiteley.
\newblock {Global consensus Monte Carlo}.
\newblock \emph{arXiv preprint arXiv:1807.09288}, 2018.

\bibitem[{Sato}(2001)]{GVI_sato}
M.~{Sato}.
\newblock {Online Model Selection Based on the Variational Bayes}.
\newblock \emph{Neural Computation}, 13\penalty0 (7):\penalty0 1649--1681,
  2001.

\bibitem[Scott et~al.(2016)Scott, Blocker, Bonassi, Chipman, George, and
  McCulloch]{bayes_and_big_data}
Steven~L. Scott, Alexander~W. Blocker, Fernando~V. Bonassi, Hugh~A. Chipman,
  Edward~I. George, and Robert~E. McCulloch.
\newblock {Bayes and Big Data: The Consensus Monte Carlo Algorithm}.
\newblock \emph{International Journal of Management Science and Engineering
  Management}, 11:\penalty0 78--88, 2016.
\newblock URL
  \url{http://www.tandfonline.com/doi/full/10.1080/17509653.2016.1142191}.

\bibitem[Vehtari et~al.(2020)Vehtari, Gelman, Sivula, Jylanki, Tran, Sahai,
  Blomstedt, Cunningham, Schiminovich, and
  Robert]{expectation_propagation_way_life_paratitioned_data}
Aki Vehtari, Andrew Gelman, Tuomas Sivula, Pasi Jylanki, Dustin Tran, Swupnil
  Sahai, Paul Blomstedt, John~P. Cunningham, David Schiminovich, and
  Christian~P. Robert.
\newblock {Expectation Propagation as a Way of Life: A Framework for Bayesian
  Inference on Partitioned Data}.
\newblock \emph{Journal of Machine Learning Research}, 21\penalty0
  (17):\penalty0 1--53, 2020.
\newblock URL \url{http://jmlr.org/papers/v21/18-817.html}.

\bibitem[Wang et~al.(2020)Wang, Liu, Liang, Joshi, and Poor]{wang2020tackling}
Jianyu Wang, Qinghua Liu, Hao Liang, Gauri Joshi, and H~Vincent Poor.
\newblock {Tackling the Objective Inconsistency Problem in Heterogeneous
  Federated Optimization}.
\newblock \emph{arXiv preprint arXiv:2007.07481}, 2020.

\bibitem[Wang \& Dunson(2013)Wang and Dunson]{parallel_MCMC_weierstrass}
Xiangyu Wang and David~B. Dunson.
\newblock {Parallel MCMC via Weierstrass Sampler}.
\newblock \emph{ArXiv}, abs/1312.4605, 2013.

\bibitem[Wei \& Conlon(2019)Wei and Conlon]{parallel_MCMC_hierarchical}
Zheng Wei and Erin~M Conlon.
\newblock {Parallel Markov chain Monte Carlo for Bayesian hierarchical models
  with big data, in two stages}.
\newblock \emph{Journal of Applied Statistics}, 46\penalty0 (11):\penalty0
  1917--1936, 2019.

\bibitem[Welling \& Teh(2011)Welling and Teh]{sgld}
Max Welling and Yee~Whye Teh.
\newblock {Bayesian Learning via Stochastic Gradient Langevin Dynamics}.
\newblock In \emph{Proceedings of the 28th International Conference on
  International Conference on Machine Learning}, ICML’11, pp.\  681–688,
  Madison, WI, USA, 2011. Omnipress.
\newblock ISBN 9781450306195.

\bibitem[Yoon et~al.(2018)Yoon, Kim, Dia, Kim, Bengio, and
  Ahn]{Bayesian_agnostic_Meta_learning}
Jaesik Yoon, Taesup Kim, Ousmane Dia, Sungwoong Kim, Yoshua Bengio, and Sungjin
  Ahn.
\newblock {Bayesian Model-Agnostic Meta-Learning}.
\newblock In \emph{Advances in Neural Information Processing Systems 31}, pp.\
  7332--7342. Curran Associates, Inc., 2018.

\bibitem[Yurochkin et~al.(2019)Yurochkin, Agarwal, Ghosh, Greenewald, Hoang,
  and Khazaeni]{Bayesian_NP_FL_NN}
Mikhail Yurochkin, Mayank Agarwal, Soumya Ghosh, Kristjan Greenewald, Nghia
  Hoang, and Yasaman Khazaeni.
\newblock {Bayesian Nonparametric Federated Learning of Neural Networks}.
\newblock In \emph{Proceedings of the 36th International Conference on Machine
  Learning}, volume~97 of \emph{Proceedings of Machine Learning Research}, pp.\
   7252--7261, Long Beach, California, USA, 09--15 Jun 2019. PMLR.

\bibitem[Zhang(2006)]{IT_bounds_statistics_estimation}
Tong Zhang.
\newblock {Information-theoretic upper and lower bounds for statistical
  estimation}.
\newblock \emph{IEEE Transactions on Information Theory}, 52\penalty0
  (4):\penalty0 1307--1321, 2006.

\bibitem[Zhang et~al.(2020)Zhang, Hong, Dhople, Yin, and Liu]{fedpd}
Xinwei Zhang, Mingyi Hong, Sairaj Dhople, Wotao Yin, and Yang Liu.
\newblock {FedPD: A Federated Learning Framework with Optimal Rates and
  Adaptivity to Non-IID Data}.
\newblock \emph{arXiv preprint arXiv:2005.11418}, 2020.

\bibitem[Zhou(2008)]{derivative_RKHS}
Ding-Xuan Zhou.
\newblock {Derivative reproducing properties for kernel methods in learning
  theory}.
\newblock \emph{Journal of computational and Applied Mathematics}, 220\penalty0
  (1-2):\penalty0 456--463, 2008.

\bibitem[Zhuo et~al.(2018)Zhuo, Liu, Shi, Zhu, Chen, and
  Zhang]{message_passing_SVGD}
Jingwei Zhuo, Chang Liu, Jiaxin Shi, Jun Zhu, Ning Chen, and Bo~Zhang.
\newblock {Message Passing Stein Variational Gradient Descent}.
\newblock In \emph{International Conference on Machine Learning}, pp.\
  6018--6027. PMLR, 2018.

\end{thebibliography}
